%% file: CoT-MLLMs_survey.tex
\documentclass{article} 
\usepackage{iclr2025_conference,times}

\input{math_commands.tex}

\usepackage{hyperref}
\usepackage{url}
\usepackage{tikz}
\usepackage{graphicx}
\iclrfinalcopy

\usepackage{tabularx}
\usepackage{array}
\usepackage{booktabs}  
\usepackage{float}

\title{From Perception to Reasoning: Deep Thinking Empowers Multimodal Large Language Models}

\author{
\small{\textbf{Wenxin Zhu}\normalfont{\textsuperscript{1}}, \textbf{Andong Chen}\textsuperscript{1}, 
\textbf{Yuchen Song}\textsuperscript{1},
\textbf{Kehai Chen}\textsuperscript{2},
\textbf{Conghui Zhu}\textsuperscript{1},
\textbf{Ziyan Chen}\textsuperscript{3},
\textbf{Tiejun Zhao}\textsuperscript{1}
}
\\
\small{\textsuperscript{1}Harbin Institute of Technology \quad \textsuperscript{2}Harbin Institute of Technology, Shenzhen} \\
\small{\textsuperscript{3}Global Tone Communication Technology} \\
\small{\texttt{spacetravel.xin@gmail.com, ands691119@gmail.com, songyuchn@126.com,}} \\
\small{\texttt{\{chenkehai, conghui\}@hit.edu.cn, chenziyan@gtcom.com.cn, tjzhao@hit.edu.cn}}
}

\begin{document}

\maketitle

\begin{abstract}
With the remarkable success of Multimodal Large Language Models (MLLMs) in perception tasks, enhancing their complex reasoning capabilities has emerged as a critical research focus. Existing models still suffer from challenges such as opaque reasoning paths and insufficient generalization ability. Chain-of-Thought (CoT) reasoning, which has demonstrated significant efficacy in language models by enhancing reasoning transparency and output interpretability, holds promise for improving model reasoning capabilities when extended to the multimodal domain. This paper provides a systematic review centered on "Multimodal Chain-of-Thought" (MCoT). First, it analyzes the background and theoretical motivations for its inception from the perspectives of technical evolution and task demands. Then, it introduces mainstream MCoT methods from three aspects: CoT paradigms, the post-training stage, and the inference stage, while also analyzing their underlying mechanisms. Furthermore, the paper summarizes existing evaluation benchmarks and metrics, and discusses the application scenarios of MCoT. Finally, it analyzes the challenges currently facing MCoT and provides an outlook on its future research directions.
\end{abstract}

\section{Introduction}
With the support of vast data and powerful computational resources, Multimodal Large Language Models (MLLMs) have made significant advances in the understanding and generation of cross-modal content (e.g., text, images, videos) in recent years\citep{kim2021vilt, li2021align, li2022blip, yu2022coca, wang2023image, chen2023x, li2023blip, wang2023caption, bai2025qwen2, zhang2023video, liu2023visual, lu2024deepseek, dong2023dreamllm, lin2024video, yin2023survey, caffagni2024revolution, zhang2024mm}. They have been widely applied to tasks such as Image Captioning, Visual Question Answering (VQA)\citep{antol2015vqa}, and Video Captioning\citep{venugopalan2015sequence}. Despite their impressive performance in perception and generation, MLLMs still exhibit notable shortcomings when confronted with tasks involving complex reasoning\citep{ghaffari2024exploring, malkinski2024reasoning, shiri2024empirical, imam2025can}. Specifically, current MLLMs primarily rely on implicit reasoning, which entails making predictions based on statistical patterns in training data\citep{bai2024survey, wang2025multimodal}, and they lack explicit, interpretable intermediate reasoning steps. Consequently, they are limited in their capabilities for multi-step logical inference, causal reasoning, and compositional generalization\citep{lu2022learn, li2025perception}.

To address similar issues in Large Language Models (LLMs), researchers have introduced the Chain-of-Thought (CoT) reasoning mechanism\citep{wei2022chain, kojima2022large, wang2022self}, aiming to enhance their logical reasoning abilities. The core idea of Chain-of-Thought is to decompose complex problems into a series of explicit intermediate reasoning steps, thereby simulating the human process of constructing a logical chain step by step. This approach has demonstrated notable advantages in tasks involving arithmetic, commonsense, and logical reasoning, while also improving the interpretability and transparency of model decision-making\citep{huang2022towards, chu2024navigate, xia2025beyond}. In the LLM domain, representative models such as OpenAI o1 and DeepSeek-R1\citep{guo2025deepseek}have integrated Chain-of-Thought strategies to achieve significant breakthroughs in mathematical and logical reasoning tasks, effectively advancing the reasoning capabilities of LLMs.

Inspired by this successful paradigm, researchers have recently begun exploring the extension of Chain-of-Thought reasoning to MLLMs, leading to the development of Chain-of-Thought-based Multimodal Large Language Models (CoT-MLLMs)\citep{zhang2023multimodal, chen2024m3cot}. This emerging paradigm aims to embed structured reasoning steps into the multimodal modeling framework, enabling models to perform higher-level cross-modal logical reasoning when processing various modalities such as vision and language. By doing so, CoT-MLLMs can capture deep semantic associations across modalities, thereby improving overall reasoning performance and interpretability\citep{bi2025reasoning, lin2025mind, chen2025towards}. Compared to LLMs that deal with a single text modality, enhancing the reasoning capabilities of MLLMs poses additional challenges, as they must simultaneously handle multimodal information. Thus, establishing effective Chains of Thought across modalities becomes a critical challenge. This involves not only aligning information across modalities but also constructing hierarchical reasoning structures to support deep cross-modal inference.

To synthesize the current state of development in the field, several surveys\citep{wang2025multimodal, bi2025reasoning, li2025perception} have emerged, offering researchers a comprehensive overview. Building upon prior work, this survey aims to provide a differentiated perspective with a greater emphasis on in-depth theoretical analysis. In contrast to existing works that focus on summarizing technical approaches, the core contribution of this survey lies in further analyzing and discussing the underlying mechanism by which MCoT enhances the reasoning capabilities of models, thereby answering the key question of "why it is effective." Furthermore, in the sections concerning the evaluation of, as well as the challenges and future directions, this survey provides a more systematic classification and summary. In summary, this review systematically organizes the latest research progress in the field of CoT-MLLMs, covering its core methods, evaluation benchmarks and metrics, typical application domains, as well as existing challenges and future directions. We hope that the insights and synthesis provided in this work can offer a structured reference and theoretical foundation for this emerging research direction of CoT-MLLMs, thereby promoting the sustained development of the field.

\subsection{Survey Organization}
\begin{figure}[h]
\begin{center}
\includegraphics[width=1.0\textwidth]{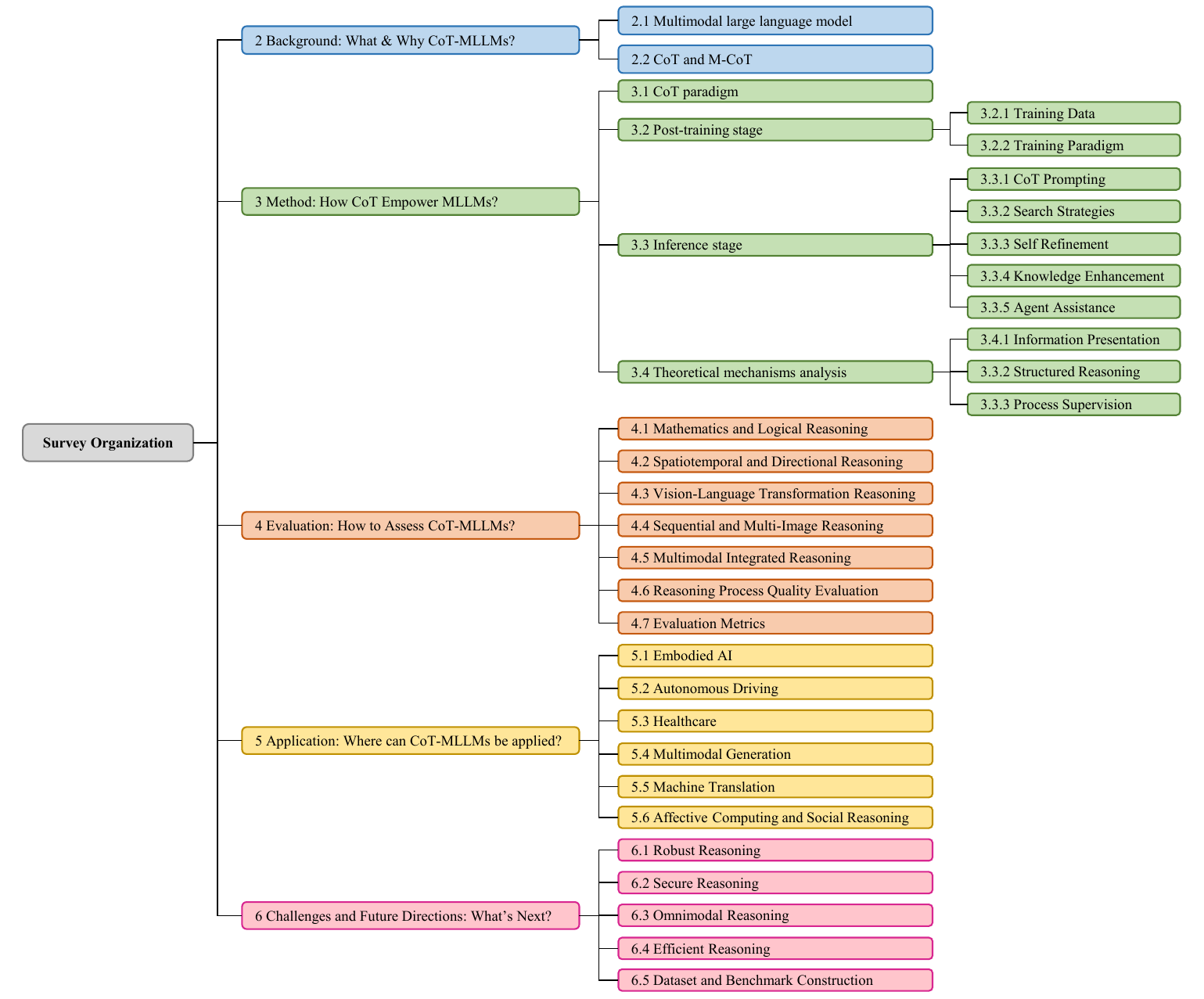}
\end{center}
\caption{The Organization of our survey.}
\label{fig_organization}
\end{figure}

As shown in Figure~\ref{fig_organization}, the organizational structure of the main body of this survey is as follows: We begin in section~\ref{background} by introducing the fundamental concepts and background knowledge related to CoT-MLLMs. Then, in section~\ref{method}, we summarize the Chain-of-Thought paradigms adopted in existing CoT-MLLMs, along with mainstream methods used during training and inference, and analyzed their underlying mechanisms. Next, in section~\ref{evaluation}, we provide an overview of the current evaluation benchmarks and metrics. Subsequently, in section~\ref{application}, we present several representative application domains of multimodal Chain-of-Thought reasoning. Finally, in section~\ref{challenge}, we conduct an in-depth discussion of the challenges faced in this field and outline potential future research directions from multiple perspectives.

\section{Background:What and Why CoT-MLLMs?}
\label{background}
Against the backdrop of MLLMs continuously pushing the boundaries of traditional artificial intelligence, researchers have gradually come to recognize their limitations in handling complex reasoning tasks. Although existing MLLMs have achieved significant progress in cross-modal understanding and generation, their reasoning capabilities remain constrained by the end-to-end learning paradigm, making it difficult to handle tasks that require logical deduction, causal inference, or multi-step reasoning. As a result, the introduction of Chain-of-Thought reasoning has emerged as a potential solution. The successful application of CoT in large-scale language models demonstrates that incorporating explicit reasoning chains can effectively enhance the model’s capacity for complex reasoning and improve its interpretability. Building on this foundation, researchers have begun to explore how to integrate the Chain-of-Thought approach into multimodal models, giving rise to CoT-MLLMs—Multimodal Large Language Models equipped with Chain-of-Thought reasoning capabilities. This research direction not only involves constructing explicit reasoning chains within MLLMs but also requires addressing challenges such as cross-modal information fusion, knowledge alignment, and training optimization. As research deepens, CoT-MLLMs are gradually revealing their potential in complex multimodal reasoning tasks, driving progress in areas such as embodied intelligence, healthcare, and machine translation.

\subsection{Multimodal Large Language Model}
In recent years, MLLMs have achieved remarkable progress in cross-modal understanding and generation tasks, giving rise to a number of representative models. For example, Flamingo\citep{alayrac2022flamingo} demonstrates excellent performance in zero-shot and few-shot tasks through a vision-language bridging module; BLIP-2\citep{li2023blip} employs a two-stage training strategy to achieve efficient vision-language alignment; LLaVA\citep{liu2023visual} enhances multimodal reasoning capabilities by effectively adapting visual encoders to language models. In addition, GPT-4V, as the multimodal version of GPT-4, exhibits strong integrated reasoning abilities in image understanding tasks.

The fundamental architecture of MLLMs typically comprises three core components: modality encoders, a modality fusion module, and a large language model. The modality fusion module is a key component of MLLMs, responsible for semantic alignment and information integration across different modalities to facilitate cross-modal understanding and generation. Common approaches include the Q-Former mechanism, adopted by models such as BLIP-2\citep{li2023blip}, Video-LLaMA \citep{zhang2023video}, and InstructBLIP\citep{dai2023instructblip}, as well as linear projection layers or multi-layer perceptrons (MLPs) used in models like MiniGPT-4\citep{zhu2023minigpt}, LLaVA\citep{liu2023visual}, and PMC-VQA\citep{zhang2023pmc}. Through the coordinated operation of these three modules, MLLMs are capable of inter-modal information interaction, enabling cross-modal reasoning and content generation.

Currently, most MLLMs perform well on basic perception tasks—such as object recognition and image captioning—but still exhibit notable deficiencies in complex reasoning tasks. For instance, Yang et al.\citep{yang2025r1} observed that while Qwen2.5-VL\citep{bai2025qwen2} possesses certain multimodal processing capabilities, it tends to make reasoning errors involving spatial relationships and logical judgments in region coloring tasks, ultimately leading to incorrect answers. Similarly, Toh et al.\citep{toh2025jumping} found that the OpenAI o1 model underperforms humans on the PUZZLEVQA dataset, which involves simple visual abstract reasoning, and that GPT-4o shows significant weaknesses in perceptual and inductive reasoning. The primary reason lies in the fact that most current MLLMs rely heavily on statistical pattern prediction, lacking explicit reasoning mechanisms. This makes them prone to jumping steps, misattribution, and other logical errors, rendering it difficult to replicate the step-by-step deductive reasoning process characteristic of human cognition.

On the other hand, the visual perception capabilities of current MLLMs also remain incomplete. In particular, for reasoning-intensive tasks, there is often a disconnect between perception and understanding\citep{wang2024mr, jiang2024chatrex, chen2025bring, yan2025visuriddles}. Multimodal reasoning requires the model to efficiently extract, integrate, and organize information from multiple input sources such as images and text. However, existing models still face evident bottlenecks in this process. In experiments, Ghosal et al.\citep{ghosal2024language} found that when models are provided with more accurate and structured perceptual information, their performance on visual reasoning tasks improves significantly—further highlighting the inadequacies in current MLLMs' mechanisms for coordinating perception and reasoning.

\begin{figure}[h]
\begin{center}
\includegraphics[width=0.8\textwidth]{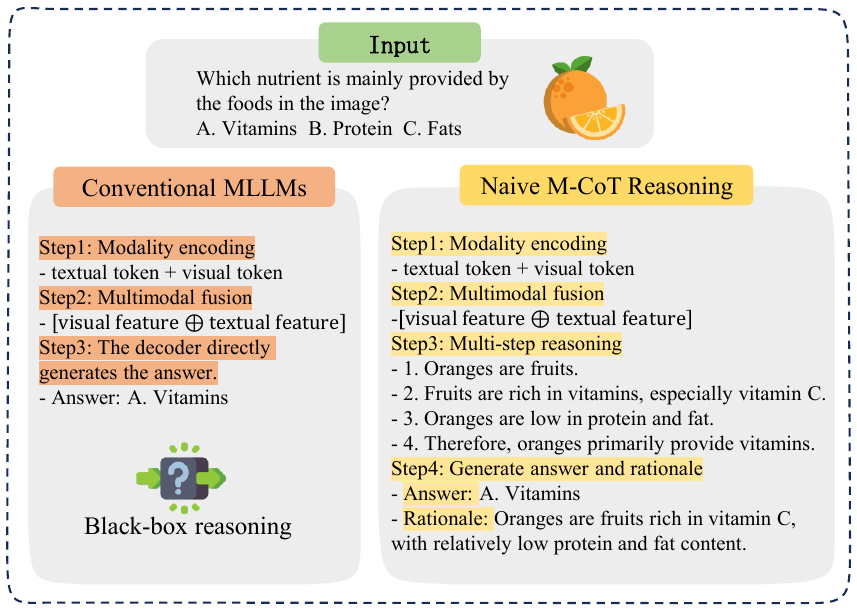}
\end{center}
\caption{Comparison between conventional MLLMs and naive M-CoT reasoning.}
\label{fig_comparison}
\end{figure}

\subsection{CoT and M-CoT}

To address the two major challenges faced by current MLLMs in complex reasoning tasks—namely, the lack of explicit and structured reasoning mechanisms, and the disconnection between perception and reasoning—researchers have introduced the CoT reasoning mechanism as an effective solution. By incorporating explicit intermediate reasoning steps, CoT enables models to simulate the step-by-step cognitive reasoning process of humans, and has been proven to significantly improve the performance of large models on complex cognitive tasks\citep{wei2022chain}.

Formally, given a textual input \(x\), traditional large language models directly model \( P(y \mid x) \) when answering questions, i.e., generating answers in a probability-based autoregressive manner. In contrast, with the introduction of Chain-of-Thought, the reasoning process can be restructured as:

\begin{equation}
P(y \mid x) = \sum_{r_1, \dots, r_n} P(y \mid r_n) \prod_{t=1}^{n} P(r_t \mid r_{t-1}, x)
\end{equation}

where \(r_t\) denotes the textual reasoning process at step \(t\). This method alleviates issues such as step-skipping and vague reasoning by incrementally constructing a reasoning path, and improves the traceability of the reasoning chain.
Based on this, researchers have further developed various extended forms. For example, Long et al.\citep{long2023large} and Yao et al.\citep{yao2023tree} proposed Tree-of-Thought (ToT), which enhances exploration and evaluation capabilities by constructing branched reasoning paths; Chen et al. proposed Chain-of-Symbolic, which transforms natural language reasoning steps into formal symbolic operations, significantly improving model performance in mathematical proof tasks.

To further address the problem of "disconnection between perception and reasoning", researchers have extended the CoT paradigm to the multimodal domain, proposing Multimodal Chain-of-Thought (M-CoT), enabling collaborative reasoning across heterogeneous modalities\citep{zhang2023multimodal}. Figure~\ref{fig_comparison} shows a comparative example between the black-box reasoning of conventional multimodal large models and a naive M-CoT reasoning method. It can be seen that the main advantage of M-CoT lies in the introduction of an explicit reasoning process. Some more advanced M-CoT methods further integrate modality perception into the multi-step reasoning process to achieve better collaborative reasoning across multiple modalities. M-CoT not only inherits the advantages of traditional CoT in modeling intermediate states and structured reasoning, but also introduces a cross-modal interaction mechanism, enabling joint perception and collaborative reasoning across modalities such as vision and language. Its reasoning process can be divided into three stages: the intra-modal semantic evolution stage, which depends not only on its own historical states and original input, but also on the regulation and guidance from the previous multimodal fusion state; the construction stage of cross-modal CoT fusion representation, where the current fusion state is based on all modalities’ current semantic states and the previous fusion state, further achieving semantic alignment; and the task-oriented output stage, which generates the final output based on the final state of the fusion reasoning chain, ensuring the coherence of the reasoning path and the interpretability of the answer.
The above process can be formally expressed as:

\begin{equation}
P_{modal} = \prod_{t=1}^T \prod_{m\in \mathcal{M}} P(r_t^m \mid r_{t-1}^m, r_{t-1}, x^m)
\end{equation}

\begin{equation}
P_{fusion} = \prod_{t=1}^T P(r_t \mid r_{t-1}, \{r_t^m\}_{m\in \mathcal{M}})
\end{equation}

\begin{equation}
P(\mathcal{Y} \mid \mathcal{X}) = \sum_{r_1, \dots, r_T} P(\mathcal{Y} \mid r_T) \cdot P_{fusion} \cdot P_{modal}
\end{equation}

where \(\mathcal{M}\) denotes the set of all modalities, such as visual modality \(v\) and textual modality \(l\);  \(\mathcal{X}=\{x^m\}_{m \in \mathcal{M}}\) represents the multimodal input, and \(x^m\) is the original input of modality \(m\); \(r_t^m\) denotes the intermediate reasoning process of modality \(m\) at time step \(t\), while \(r_t\) represents the joint reasoning process after fusion of all modalities; \(P_{modal}\) models the intra-modal state evolution, \(P_{fusion}\) describes semantic interaction and reasoning integration across modalities, and \(P(\mathcal{Y} \mid \mathcal{X})\) is the final task output, which can be in multiple modalities.

In summary, M-CoT enhances the modeling capability of traditional multimodal large models for complex semantic relationships through structured reasoning paths and effectively alleviates the interpretability issues of "black-box" prediction. As a novel multimodal reasoning paradigm, M-CoT bridges the gap between modality perception and logical deduction, providing strong support for improving the performance and interpretability of multimodal tasks.

\section{Method: How CoT Empowers CoT-MLLMs?}
\label{method}

\begin{table}[t]
\caption{Summary of the methods and models of MCoT.}
\label{table_method}
\centering 
\resizebox{\textwidth}{!}{%
\begin{tabular}{ccccccc}
\toprule 
\textbf{Method/Model} & \textbf{Time} & \textbf{Organization} & \textbf{Modality} & \textbf{Post-training} & \textbf{TTS} & \textbf{Domain} \\
\midrule 
HoT\citep{yao2023thinking} & 2023-8 & UCAS & Image,Text & FT & — & Science QA \\
MaTCR\citep{liu2023matcr} & 2023-10 & ICT CAS & Image,Text & FT & — & Multimodal Generation \\
DDCoT\citep{zheng2023ddcot} & 2023-10 & ShanghaiTech & Image,Text & — & CoT prompting & Visual QA \\
Chain-of-Look\citep{xi2023chain} & 2023-10 & SUNY Buffalo & Video,Text & — & CoT prompting & Medical Image Analysis \\
GCoT\citep{ma2024dolphins} & 2023-12 & UW-Madison & Video,Text & — & — & Autonomous Driving \\
CoCoT\citep{zhang2024cocot} & 2024-1 & UR & Image,Text & — & CoT prompting & Multi-image Understanding \\
KAM-CoT\citep{mondal2024kam} & 2024-1 & Samsung & Image,Text & — & Knowledge Augmentation & Science QA \\
VCoT\citep{rose2023visual} & 2024-1 & UCSB & Image,Text & — & — & Multi-domain \\
BBA\citep{zhao2024bba} & 2024-2 & Tencent & Image,Text & — & — & Multi-domain \\
MM-Retrieval\citep{liu2023retrieval} & 2024-3 & XMU & Image,Text & — & Knowledge Augmentation & Math, Science QA \\
CoS\citep{liu2024chain} & 2024-3 & THU & Image,Text & — & — & Image QA \\
CCoT\citep{mitra2024compositional} & 2024-4 & UC Berkeley & Image,Text & — & CoT prompting & Multi-domain \\
MI-CoT\citep{anand2024mm} & 2024-4 & IIIT-Delhi & Image,Text & — & CoT prompting & Physics Reasoning \\
Cantor\citep{gao2024cantor} & 2024-4 & XMU & Image,Text & — & — & Math, Science QA \\
VoT\citep{fei2024video} & 2024-5 & NUS & Video,Text & — & Self-refinement & Visual Understanding \\
Multimodal-CoT\citep{zhang2023multimodal} & 2024-5 & SJTU & Image,Text & — & — & Science QA \\
IoT\citep{zhou2024image} & 2024-5 & Westlake & Image,Text & — & CoT prompting & Visual Understanding \\
CoTDiffusion\citep{ni2024generate} & 2024-6 & TJU & Image,Text & — & — & Embodied AI \\
MC-CoT\citep{tan2024boosting} & 2024-7 & Westlake & Image,Text & — & Self-consistency & Visual QA \\
DCoT\citep{jia2025dcot} & 2024-9 & NEU & Image,Text & — & CoT prompting & Visual QA \\
CoT-ST\citep{du2024cot} & 2024-9 & HIT & Speech,Text & FT & — & Speech Translation \\
ARES\citep{byun2024ares} & 2024-10 & OSU & Image,Text & FT + RL & — & Visual QA \\
R-CoT\citep{deng2024r} & 2024-10 & HUST & Image,Text & — & — & Geometric Reasoning \\
SKETCHPAD\citep{hu2024visual} & 2024-11 & UW & Image,Text & — & — & Math Reasoning \\
LLaVA-o1\citep{xu2024llava} & 2024-11 & THU & Image,Text & FT & Beam Search & Multimodal Reasoning \\
VIC\citep{zheng2024thinking} & 2024-11 & UPenn & Image,Text & — & — & Multimodal Reasoning \\
Insight-V\citep{dong2025insight} & 2024-11 & NTU & Image,Text & FT + RL & — & Multimodal Reasoning \\
VisVm\citep{wang2024scaling} & 2024-12 & Microsoft & Image,Text & — & Search Method & Multimodal Reasoning \\
AtomThink\citep{xiang2024atomthink} & 2024-12 & SYSU & Image,Text & FT & Search Method & Math Reasoning \\
EMMA-X\citep{sun2024emma} & 2024-12 & SUTD & Image,Text & — & — & Embodied AI \\
AR-MCTS\citep{dong2024progressive} & 2024-12 & RUC & Image,Text & — & Tree Search & Multi-domain \\
Mulberry\citep{yao2024mulberry} & 2024-12 & NTU & Image,Text & FT & Tree Search & Multi-task \\
LlamaV-o1\citep{thawakar2025llamav} & 2025-1 & MBZUAI & Image,Text & FT & Beam Search & Multi-task \\
MVoT\citep{li2025imagine} & 2025-1 & Microsoft & Image,Text & FT & — & Spatial Reasoning \\
CoR\citep{yu2025chain} & 2025-1 & THU & Image,Text & — & — & Math Reasoning \\
DeepSeek-R1\citep{guo2025deepseek} & 2025-1 & DeepSeek & Image,Text & FT + RL & — & Multimodal Reasoning \\
SpatialCoT\citep{liu2025spatialcot} & 2025-1 & Huawei & Image,Text & FT & — & Spatial Reasoning \\
Virgo\citep{du2025virgo} & 2025-2 & RUC & Image,Text & FT & — & Multimodal Reasoning \\
AStar\citep{wu2025boosting} & 2025-2 & THU & Image,Text & — & Tree Search & Multimodal Reasoning \\
LLaVA-CoT\citep{xu2024llava} & 2025-2 & THU & Image,Text & FT & Beam Search & Multimodal Reasoning \\
MedVLM-R1\citep{pan2025medvlm} & 2025-2 & ICL & Image,Text & RL & — & Medical Image Analysis \\
Visual-RFT\citep{liu2025visual} & 2025-3 & SJTU & Image,Text & RL & — & Visual Perception \\
Audio-Reasoner\citep{xie2025audio} & 2025-3 & NTU & Audio,Text & FT & — & Audio Reasoning \\
R1-Omni\citep{zhao2025r1} & 2025-3 & Alibaba & Video,Audio,Text & RL & — & Emotion Recognition \\
Vision-R1\citep{huang2025vision} & 2025-3 & ECNU & Image,Text & FT + RL & — & Math Reasoning \\
DriveLMM-o1\citep{ishaq2025drivelmm} & 2025-3 & MBZUAI & Image,Text & FT & — & Autonomous Driving \\
EmbodiedVSR\citep{zhang2025embodiedvsr} & 2025-3 & Beijing HRC & Image,Text & — & — & Embodied AI \\
ICoT\citep{gao2025interleaved} & 2025-3 & Soochow & Image,Text & — & CoT prompting & Visual Reasoning \\
R1-Onevision\citep{yang2025r1} & 2025-3 & ZJU & Image,Text & FT + RL & — & Multimodal Reasoning \\
Embodied-Reasoner\citep{zhang2025embodied} & 2025-3 & ZJU & Image,Text & — & — & Embodied AI \\
VisuoThink\citep{wang2025visuothink} & 2025-4 & FDU & Image,Text & — & Tree Search & Geometric, Spatial Reasoning \\
\bottomrule 
\end{tabular}%
}
\end{table}

Current MLLMs still exhibit significant limitations in reasoning capabilities. To enhance their capacity for complex reasoning, researchers have explored the integration of CoT mechanisms to enable clear and interpretable multi-step reasoning. This chapter focuses on the question of how CoT empowers MLLMs with reasoning abilities. It first introduces the core characteristics of different CoT paradigms, then discusses key implementation strategies during the post-training and inference stages, and finally summarizes the mainstream Multimodal CoT methods and representative models in Table~\ref{table_method}.

\subsection{CoT Paradigm}

\begin{figure}[h]
\begin{center}
\includegraphics[width=1.0\textwidth]{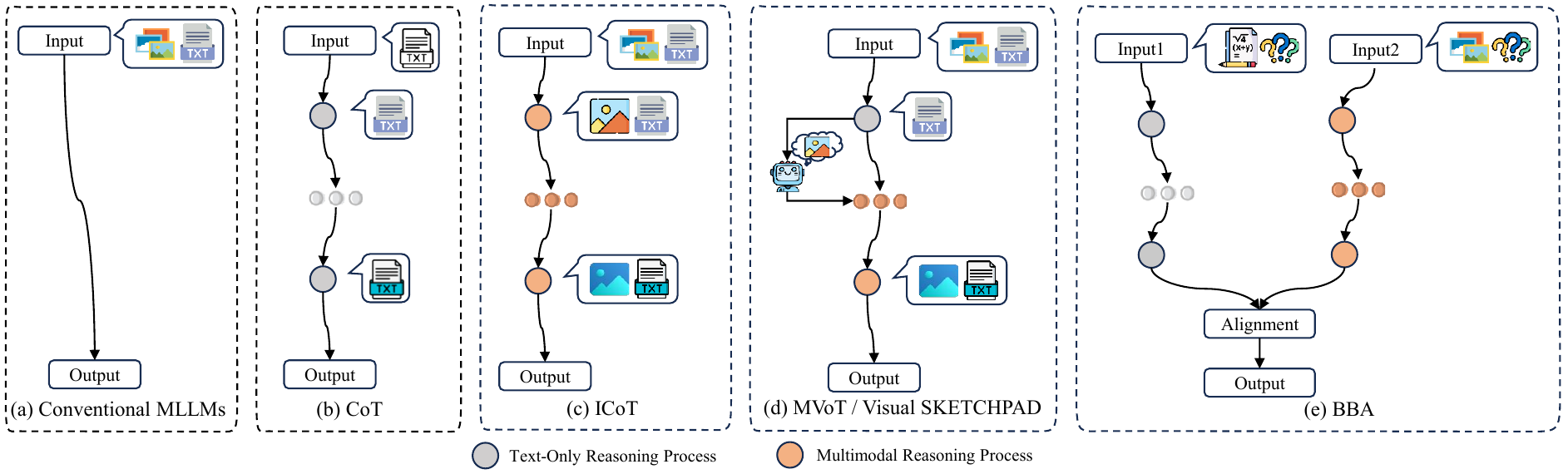}
\end{center}
\caption{Comparison of reasoning paradigms among conventional MLLMs, text-only CoT, and M-CoT:
(a) End-to-end “black-box” reasoning in conventional MLLMs;
(b) Text-only CoT reasoning;
(c) ICoT incorporates visual inputs as part of the reasoning content;
(d) MVoT and Visual SKETCHPAD assist the reasoning process by generating images;
(e) The BBA approach constructs separate reasoning chains for textual and visual modalities and performs alignment and integration across them.}
\label{fig_CoTparadigm}
\end{figure}

Most current MLLMs adopt an end-to-end reasoning approach based on the “modality encoder encoding + cross-modal fusion + decoder direct generation” paradigm, which represents a typical black-box reasoning process. This structure is efficient and fast in response but mainly relies on pattern matching and implicit association modeling. It lacks explicit reasoning paths and is prone to shallow semantic alignment, making it more suitable for perception-driven tasks with relatively low complexity. This approach is analogous to "System 1" thinking in cognitive science—fast and intuitive, but susceptible to errors when faced with complex problems.

To activate the model's deeper reasoning capabilities and simulate the logically rigorous "System 2" thinking of humans, researchers have proposed the CoT mechanism and further extended the initial chain structure into various topological variants of reasoning paths\citep{chu2023navigate}, including typical forms such as chain structures\citep{wei2022chain}, tree structures\citep{long2023large, yao2023tree}, and graph structures\citep{besta2024graph, lei2023boosting}. These structures organize reasoning steps in different ways to improve the model’s capacity for multi-path thinking and reasoning flexibility. The core mechanism lies in decomposing a complex, one-step multimodal reasoning task into a series of simpler, more manageable intermediate steps, thereby reducing the risk of error accumulation in end-to-end generation.

As shown in Figure~\ref{fig_CoTparadigm}, we compare the black-box reasoning approach of conventional MLLMs with representative textual and M-CoT paradigms. The chain-style CoT is the most fundamental reasoning structure, which generates intermediate steps in a linear sequence, where each reasoning result directly influences the next step. On this basis, Wang et al.\citep{wang2022self} proposed CoT-SC, which introduces multiple independent reasoning chains. Due to the inherent uncertainty in large model reasoning, the same initial input may result in different outputs across reasoning chains. A predefined scoring function is then used to select the highest-scoring answer. The effectiveness of this strategy stems from its use of multi-path sampling and consistency checking, which effectively reduces "compounding errors" caused by incidental mistakes in a single autoregressive path, thereby significantly enhancing the robustness of the results.

In multimodal scenarios, most works still adopt a single chain of thought\citep{yang2025r1, liu2024chain, gao2025interleaved}, its primary value is in making the implicit cross-modal alignment process explicit. By generating intermediate language descriptions, it builds a bridge connecting visual perception and linguistic logic, thus effectively mitigating the representation gap between modalities. Some works have incorporated multimodal elements to further strengthen this bridge. For example, CoS\citep{liu2024chain} method emphasizes the perception of key regions in an image to enhance visual reasoning, which essentially guides the model to anchor linguistic symbols to key visual evidence at each reasoning step. ICoT\citep{gao2025interleaved} innovatively incorporates images as part of the reasoning content within intermediate steps. This represents a more thorough attempt to bridge the modality gap by creating a hybrid text-image semantic space where reasoning occurs directly within the multimodal information stream. Meanwhile, MVoT\citep{li2025imagine} and Visual SKETCHPAD\citep{hu2024visual} assist the model's reasoning by generating visualizations of the process, which not only provides a visual scratchpad for the model but also offloads the cognitive load of purely linguistic symbolic reasoning through externalized visual representations.

Other variants such as CoCoT\citep{zhang2024cocot} enable MLLMs to compare similarities and differences across multiple image inputs and guide the model to answer detailed questions about them based on these findings. BBA\citep{zhao2024bba} adopts a strategy similar to CoT-SC, constructing independent reasoning chains for the visual modality and domain-specific language (DSL) representations, and subsequently aligning and integrating them. Additionally, VISUALCODER \citep{le2024visualcoder} combines CoT reasoning with visual control flow graphs for code reasoning tasks. To alleviate overthinking and structure the output, Xiang et al. proposed SCoT \citep{xiang2025can}, which adaptively generates various atomic steps while maintaining reasoning efficiency.

For more complex modalities such as video and speech, VoT\citep{fei2024video} was the first to successfully apply Chain-of-Thought techniques to achieve human-level video reasoning capabilities, demonstrating great potential across a wide range of video understanding scenarios. CoT-ST\citep{du2024cot} significantly improved performance in speech translation tasks by introducing MCoT reasoning. The success of these methods on temporal modalities further highlights the advantage of task decomposition inherent in CoT. It breaks down continuous, high-dimensional video/speech signals into a series of manageable, text-based logical checkpoints, thereby making reasoning about complex dynamic scenes feasible.

However, the topological structure of chain-of-thought is rather monotonous. The model can only follow a linear reasoning process; once an error occurs in any given step, it will propagate and be amplified along the chain. The inability to backtrack to previous steps for error checking exposes the high sensitivity of the linear reasoning paradigm to compounding errors, limiting the model's freedom and depth of exploration.

To systematically address this issue, the Tree of Thoughts (ToT) paradigm\citep{long2023large, yao2023tree} was proposed, which mimics the human tree-like thought process for problem-solving and endows the model with the ability to backtrack. In a Tree of Thoughts, a single node represents a partial solution. For a given node, the model generates multiple child nodes (i.e., alternative options) and evaluates each one. Finally, search algorithms such as Breadth-First Search (BFS) or Depth-First Search (DFS) are used to determine the expansion of the tree. The essence of ToT is to expand a single, fragile reasoning path into a robust and explorable solution space. Its evaluation and search mechanisms are equivalent to introducing active pruning and error correction capabilities, thereby systematically overcoming the compounding error problem of the linear chain. Additionally, the BoT framework\citep{chen2024boosting} iteratively explores and evaluates the tree of thoughts to gain experience from trial and error, thus arriving at a precise solution. The ToT paradigm allows the chain of thought to extensively explore the problem's solution space and achieve local or even global optimal solutions.

The Graph of Thoughts (GoT)\citep{besta2024graph, lei2023boosting} introduces a more complex topological structure, modeling a problem's concepts as graph nodes and using edges to represent the relationships between them. Unlike a tree of thoughts, a graph of thoughts allows nodes to have multiple parent nodes, introducing cycles and N-to-1 connections. This enables the model not only to explore different reasoning paths but also to aggregate and synergize multiple parallel reasoning results, thus forming a more comprehensive solution. In a graph of thoughts, each subgraph represents the solving process of a subproblem, and the combination of all subgraphs constitutes the final solution. If ToT explores in depth and breadth, then GoT adds the capability to merge, constructing a more flexible and powerful non-linear reasoning structure. In multimodal scenarios, BDoG\citep{zheng2024picture} constrains the debate process within a graph it refers to as a "blueprint," storing viewpoints and evidence in graph branches to reduce interference from frequent yet irrelevant concepts. Furthermore, Hypergraph-of-Thought (HoT)\citep{yao2023thinking} models higher-order relationships by introducing hyperedges, further enhancing the ability to align and integrate complex cross-modal semantic relationships.

\subsection{Post-Training Phase}
In MLLMs, the pre-training phase primarily serves to acquire world knowledge and align different modalities, thereby enabling basic cross-modal understanding. However, most CoT-MLLMs do not undergo pre-training from scratch; instead, they incorporate explicit reasoning mechanisms on top of existing models. As a result, the post-training phase becomes particularly critical in the modeling of M-CoT, with its core objective being to guide the model in learning CoT-style reasoning patterns and output formats, and further optimize the multi-step reasoning process. Common approaches in this phase include Supervised Fine-tuning (SFT) and Reinforcement Learning (RL), both of which offer strong task adaptability and high data efficiency. Fundamentally, the goal of the post-training stage is to transform the model from a "black box" that primarily relies on pattern matching into a "white box" capable of performing explicit logical inference.

\begin{table}[t]
\caption{Summary of the datasets used for training CoT-MLLMs. "MC" and "Open" refer to multiple-choice and open-ended answer formats, while "T", "I", "V", "A" and "PC" represent Text, Image, Video, Audio and Point Cloud, respectively.}
\label{table_dataset}
\centering
\resizebox{\textwidth}{!}{%
\begin{tabular}{ccccccc}
\toprule
\textbf{Dataset} & \textbf{Year} & \textbf{Task} & \textbf{Domain} & \textbf{Modality} & \textbf{Format} & \textbf{Size} \\
\midrule 
COCO-MMR\cite{wei2024enhancing} & 2023 & VQA & Multi-domain & T, I & Open & 62,351\\
MM-PhyQA\cite{anand2024mm} & 2024 & Physics QA & Physics & T, I & Open & 4,500\\
HaloQuest\cite{wang2024haloquest} & 2024 & VQA & Hallucination & T, I & Open & 7.7K\\
MathV360K\cite{shi2024math} & 2024 & Math QA & Math & T, I & Open & 360K\\
CMM-Math\cite{liu2024cmm} & 2024 & Math QA & Math & T, I & MC, Completion, Open & 28K\\
Visual CoT\cite{shao2024visual} & 2024 & VQA & Multi-domain & T, I & Open & 438K\\
GPSM4K\cite{anand2024improving} & 2024 & Math QA & Geometry & T, I & Open & 4,440\\
AMATH-SFT\cite{xiang2024atomthink} & 2024 & Math QA & Math & T, I & MC, Open & 157K\\
LLaVA-CoT-100k\cite{xu2024llava} & 2024 & VQA & Commonsense, Science & T, I & MC, Open & 834K\\
MLRQA\cite{xiao2024mlrqa} & 2025 & VQA & Multi-domain & T, I & MC & 4,356\\
MM-Verify\cite{sun2025mm} & 2025 & Math QA & Math & T, I & MC, Open & 59,772\\
MMathCoT1M\cite{luo2025ursa} & 2025 & Math QA & Math & T, I & MC & 1M\\
CoTA\cite{xie2025audio} & 2025 & Audio QA & Multi-domain & T, A & Open & 1.2M\\
Vision-R1-cold\cite{huang2025vision} & 2025 & VQA & Multi-domain & T, I & MC, Open & 200K\\
VisualPRM400K\cite{wang2025visualprm} & 2025 & Science QA & Math, Science & T, I & MC, Open & 400K\\
R1-Onevision\cite{yang2025r1} & 2025 & Multi-domain & Multi-domain & T, I & MC, Open & 155K\\
LongPerceptual Thoughts\cite{liao2025longperceptualthoughts} & 2025 & VQA & Visual Perception & T, I & MC & 30K \\
DriveLMM-o1\cite{ishaq2025drivelmm} & 2025 & VQA & Autonomous Driving & T, I, PC & MC, Open & 18K \\
Emma-X\cite{sun2024emma} & 2024 & Robot Manipulation & Embodied AI & T, V & Robot Action & 60K\\
Embodied-Reasoner\cite{zhang2025embodied} & 2025 & Robot Manipulation & Embodied AI & T, I & MC, Open & 9,390 \\
\bottomrule 
\end{tabular}%
}
\end{table}

\subsubsection{Training Data}
Fine-tuning CoT-MLLMs requires specialized datasets. The core value of these datasets lies in providing direct supervisory signals for the model's thought process itself, and not just for the outcome of that thinking (i.e., the final answer). Currently, such datasets are still scarce. To address this critical issue, several studies have constructed fine-tuning datasets that include reasoning processes within specific domains, thereby improving the reasoning performance of CoT-MLLMs on targeted tasks. A summary of relevant datasets is provided in Table~\ref{table_dataset}.

The COCO-MMR dataset\cite{wei2024enhancing} extracts a large number of open-ended questions, reasoning chains, and answers from the COCO dataset\cite{lin2014microsoft}, and pioneers the use of open-ended questions in the context of multimodal Chain-of-Thought reasoning to introduce new challenges for training and evaluating reasoning capabilities. MM-PhyQA\cite{anand2024mm} comprises well-structured high-school-level multimodal physics problems and enhances model performance on multi-step physical reasoning tasks. HaloQuest\cite{wang2024haloquest} focuses on capturing various aspects of multimodal hallucinations, contributing over 7.7K examples to help mitigate hallucination issues in multimodal reasoning models. Datasets such as MathV360K\cite{shi2024math}, AMATH-SFT\cite{xiang2024atomthink}, and MMathCoT1M\cite{luo2025ursa} provide high-quality data for multimodal mathematical reasoning. Additionally, CMM-Math\cite{liu2024cmm} offers a Chinese-language multimodal math reasoning dataset, while GPSM4K\cite{anand2024improving} delivers detailed step-by-step solutions in a unified format for multimodal geometric reasoning tasks.

Datasets including Visual CoT\cite{shao2024visual}, LLaVA-CoT-100K\cite{xu2024llava}, Vision-R1-cold\cite{huang2025vision}, and R1-Onevision\cite{yang2025r1} support visual reasoning tasks across various domains and contain detailed multi-step reasoning annotations. MLRQA\cite{xiao2024mlrqa}, composed by domain experts, includes logic reasoning data designed to activate the reasoning potential of MLLMs using multimodal Chain-of-Thoughts. Sun et al.\cite{sun2025mm} propose two effective methods for synthesizing multimodal CoT datasets, offering practical solutions for generating large-scale multimodal reasoning data. For audio data, CoTA\cite{xie2025audio} provides multi-task audio datasets with structured Chain-of-Thought reasoning processes.

A common characteristic of the aforementioned datasets is that they explicitly encode human problem-solving rationales and logical steps into text. When a model is fine-tuned on these data, it is not merely learning an input-to-output mapping, but rather a generalizable capability for interpretable, structured problem decomposition and step-by-step solution. Furthermore, by compelling the model to generate intermediate text that connects visual evidence to the final conclusion, these datasets significantly promote the alignment of visual and language modalities at a deeper semantic level, providing a critical training signal for mitigating the representation gap.

Given that most current CoT datasets focus on mathematically rigorous and code-based reasoning tasks, some researchers have begun exploring CoT data for other application scenarios. This expansion is of crucial importance because it demonstrates that CoT is not merely a specialized technique applicable only to mathematical logic, but rather a general framework with wide applicability that simulates the complex cognitive processes of humans. LongPerceptualThoughts\cite{liao2025longperceptualthoughts} aims to fill the gap in reasoning datasets for visual perception tasks, providing a dataset of 30K perceptual reasoning samples with long-form CoT traces. Meanwhile, Emma-X\cite{sun2024emma} and Embodied-Reasoner\cite{zhang2025embodied} supply training data for embodied intelligence models. Emma-X offers a hierarchical embodied dataset containing 60K robot operation trajectories, while Embodied-Reasoner provides 9.3K coherent observation–thought–action trajectories. DriveLMM-o1 \cite{ishaq2025drivelmm} contributes over 18K visual question-answering instances for step-by-step visual reasoning in autonomous driving, covering tasks such as perception, prediction, and planning.

\subsubsection{Training Paradigm}

\begin{figure}[h]
\begin{center}
\includegraphics[width=1.0\textwidth]{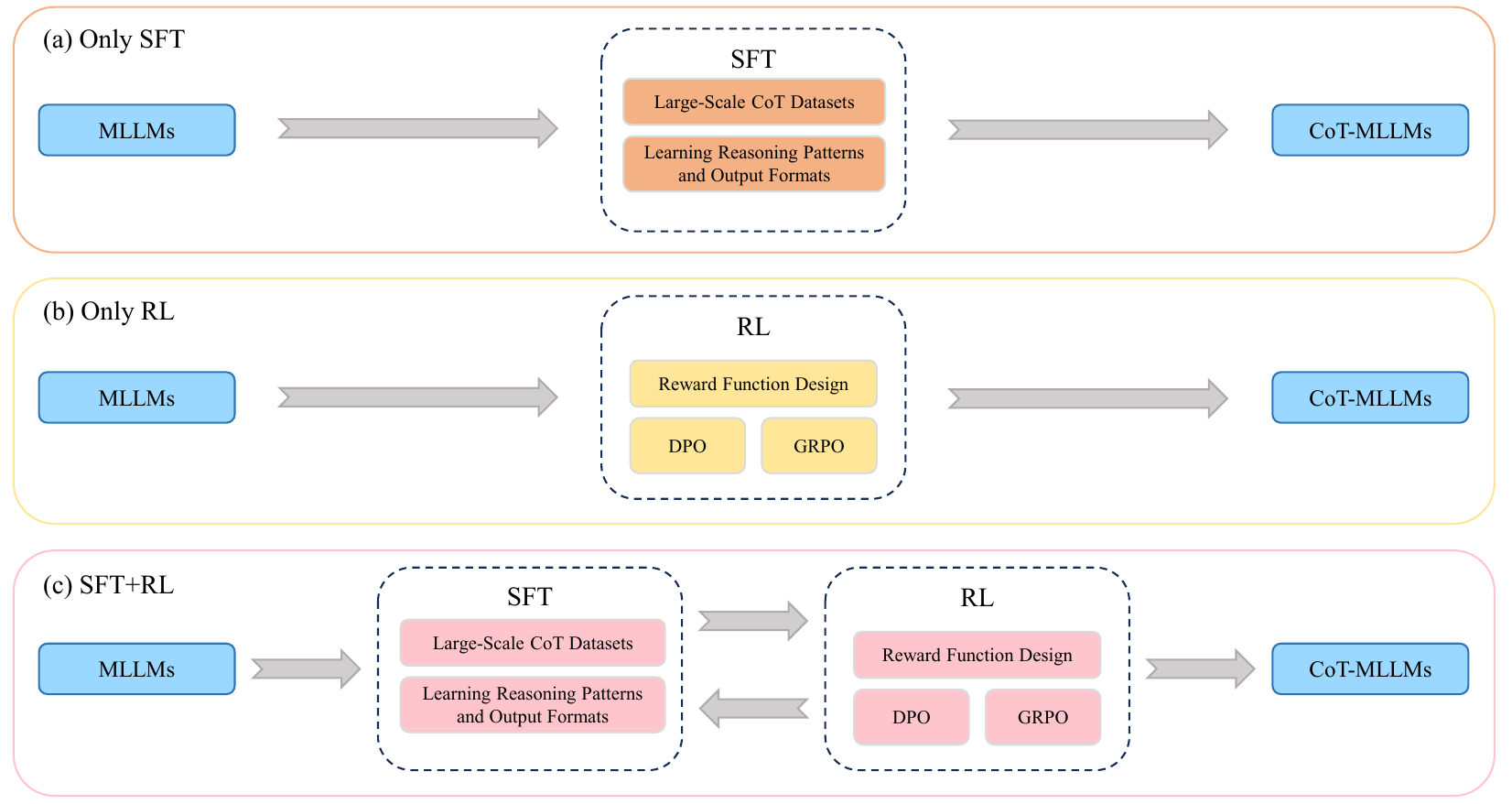}
\end{center}
\caption{Three training paradigms of CoT-MLLMs: (a) only supervised fine-tuning, (b) only reinforcement learning, (c) supervised fine-tuning + reinforcement learning}
\label{fig_trainingParadigm}
\end{figure}

In the post-training stage, MLLMs can acquire CoT reasoning capabilities through three mainstream training paradigms: supervised fine-tuning only, reinforcement learning only, and a combination of supervised fine-tuning and reinforcement learning. As illustrated in Figure~\ref{fig_trainingParadigm}, subfigures (a), (b), and (c) correspond to these three training workflows, respectively. We introduce each paradigm in detail below and compare them with the training processes and objectives of conventional MLLMs.

\textbf{Supervised Fine-tuning Only:}
As shown in subfigure (a) of Figure~\ref{fig_trainingParadigm}, this paradigm trains the model using large-scale CoT-annotated datasets mentioned earlier, enabling it to generate intermediate reasoning steps and final answers conforming to the structure of CoT. From a mechanistic perspective, this is a behavioral cloning of the reasoning process of human experts. During the training process, the model learns the reasoning pattern from the input modal data to the step-by-step reasoning process and finally to the final output.

Unlike conventional MLLMs that use supervised fine-tuning to enhance generalization, supervised fine-tuning in CoT-MLLMs focuses more on the model’s ability to produce logically structured, explicit reasoning processes. Hence, it heavily relies on the quality and diversity of the CoT structures in the training data. This method is stable and efficient to adapt, but it may suffer from limited generalization when data is insufficient or the task type is highly complex. Because it mainly learns "how to do", rather than necessarily deeply understanding "why to do it this way".

To address this limitation and enhance learning efficiency, the researchers incorporated the scaffolding theory from cognitive science and guided the model to gradually acquire the reasoning ability. For instance, LlamaV-o1\cite{thawakar2025llamav} adopts a multi-step curriculum learning strategy where the model first learns basic subtasks before progressing to multi-step reasoning training. Similarly, CoR\citep{yu2025chain} and LLaVA-CoT\citep{xu2024llava} employ a Progressive Paradigm Training (PPT) approach, introducing different types of reasoning data in stages to gradually expand the model’s reasoning capabilities.

\textbf{Reinforcement Learning Only:}
As illustrated in subfigure (b) of Figure~\ref{fig_trainingParadigm}, this paradigm skips supervised fine-tuning and directly performs reinforcement fine-tuning based on a pretrained model. Its theoretical foundation is the shift from imitation learning to goal-driven exploratory learning. The model no longer passively replicates fixed paths from the data, but rather acts as an agent to actively explore various possible reasoning paths, uses reward signals to judge the quality of those paths, and thereby discovers superior solutions that were previously unseen in the data.

Unlike conventional MLLMs that use Reinforcement Learning from Human Feedback (RLHF) to align with human preferences and reduce hallucinations, reinforcement learning in CoT-MLLMs focuses more on improving the logicality, completeness, and problem-solving effectiveness of reasoning chains. This training paradigm typically employs reinforcement fine-tuning (RFT) methods based on Direct Preference Optimization (DPO)\citep{rafailov2023direct} or Group Relative Policy Optimization (GRPO)\citep{shao2024deepseekmath}. 

For example, Seg-Zero\citep{liu2025seg} uses five manually designed reward functions to enable the model to acquire a reasonable level of reasoning ability without relying on any supervised reasoning data. Inspired by the cold-start strategy of DeepSeek-R1\citep{guo2025deepseek}, Vision-R1\citep{huang2025vision} proposes a Progressive Thinking Suppression Training (PTST) strategy to alleviate the optimization challenge of excessive reasoning in early-stage reinforcement learning. This method restricts reasoning length initially and gradually relaxes this constraint during training, enabling the model to autonomously learn to use longer reasoning chains to solve complex problems, thereby enhancing its reasoning capabilities. In addition, Visual-RFT\citep{liu2025visual} demonstrates that reinforcement fine-tuning can be effectively applied beyond mathematics and code to visual perception tasks, and it is the first to adopt Reinforcement Learning with Verifiable Rewards (RLVR) in image-based multimodal models. R1-Omni\citep{zhao2025r1} further extends RLVR to video-based multimodal models for the first time.

\textbf{Supervised Fine-tuning + Reinforcement Learning:}
As shown in subfigure (c) of Figure~\ref{fig_trainingParadigm}, this paradigm combines supervised fine-tuning with reinforcement learning in a multi-stage training process. It first uses supervised fine-tuning to establish basic reasoning patterns in the model, and then applies reinforcement learning to optimize reasoning paths, thereby enhancing the model’s performance in terms of rationality and task alignment.

ARES\citep{byun2024ares} proposes a two-stage algorithm that alternates between reinforcement learning and supervised fine-tuning. First, a teacher model scores the outputs to guide reinforcement learning; then, the teacher corrects erroneous reasoning steps, followed by supervised fine-tuning to stabilize post-RL performance. Moreover, Insight-V\citep{dong2025insight} adopts an iterative DPO strategy in the reinforcement learning phase. By conducting multiple rounds of DPO training and sampling, the model better simulates the online DPO process, effectively addressing the issue of preference data drifting from the model distribution in offline generation. To further enhance the stability of multimodal reasoning models, Wang et al.\citep{wang2024enhancing} introduce a Mixed Preference Optimization (MPO) approach, whose loss function consists of three components: preference loss, quality loss, and generation loss.

\subsection{Inference Phase}
\begin{figure}[h]
\begin{center}
\includegraphics[width=1.0\textwidth]{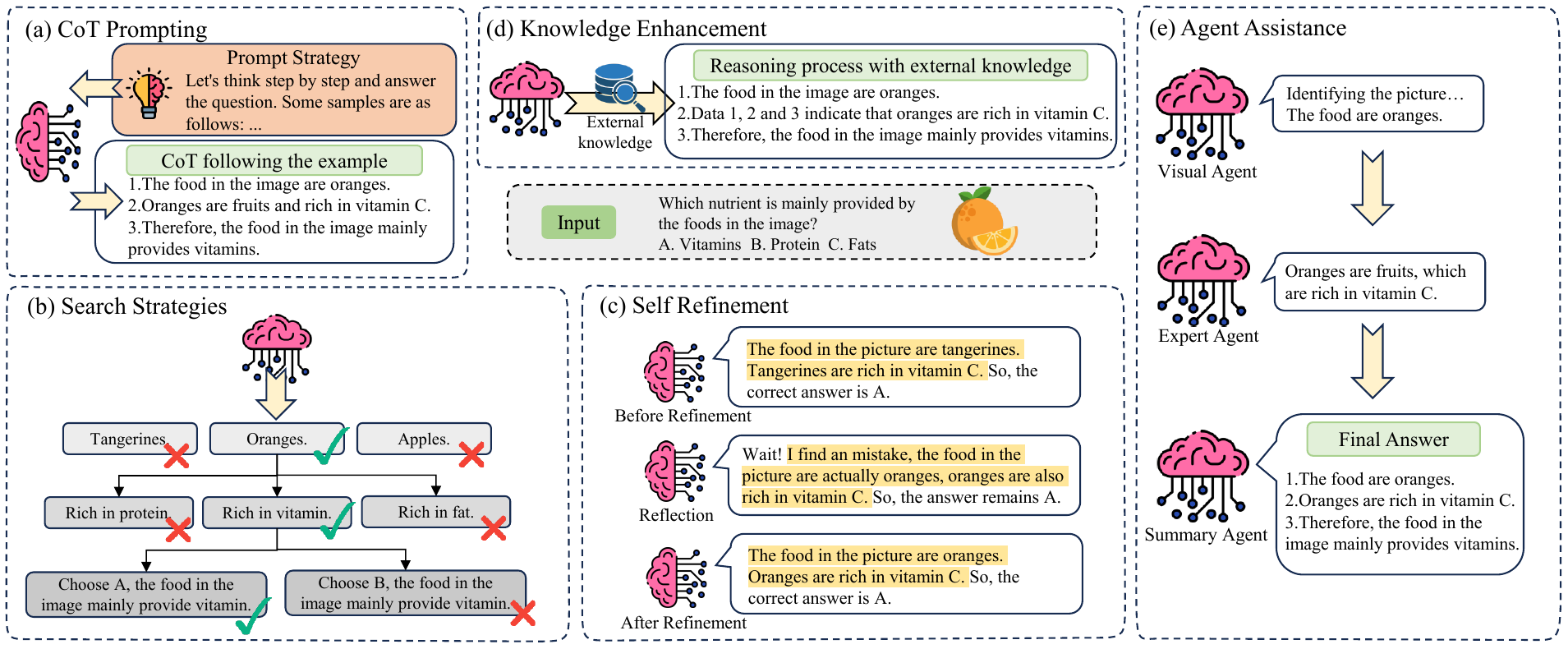}
\end{center}
\caption{Summary of the inference-time scaling methods for CoT-MLLMs: (a) CoT prompting, (b) search strategies, (c) self refinement, (d) knowledge enhancement, (e) agent assistance}
\label{fig_tts}
\end{figure}

During the model training phase, methods such as supervised fine-tuning and reinforcement learning adjust the model’s parameters to endow it with a certain level of multimodal Chain-of-Thought reasoning capability. In contrast, during the inference phase, there also exist a range of strategies that do not require updating model parameters but can still enhance the model's CoT reasoning performance. The theoretical cornerstone of these methods lies in the fact that large models have already implicitly learned rich knowledge and diverse capabilities during the pre-training process. Consequently, inference-phase strategies do not aim to teach the model new knowledge, but rather to activate its potential through ingenious guidance, directing its generation process to follow optimal computational paths. These approaches, collectively referred to as Inference-Time Scaling methods, include Chain-of-Thought prompting, search strategies, self-optimization, knowledge augmentation, and agent-assisted techniques, which are summarized in Figure~\ref{fig_tts}.

\subsubsection{CoT Prompting}
Chain-of-Thought Prompting is a low-cost method with strong generalization capability. As illustrated in subfigure (a) of Figure~\ref{fig_tts}, it explicitly provides reasoning steps through prompting strategies, guiding the model to think step-by-step, thereby enhancing its logical rigor and its ability to solve complex problems. Mechanistically, CoT prompting leverages In-Context Learning (ICL) to establish a temporary reasoning paradigm for the model, guiding it to decompose a complex single-step prediction task into a series of simple, self-consistent generation steps.

Compared to directly generating an answer, CoT prompting, through few-shot example learning or zero-shot prompts, induces the model to spontaneously form a reasoning chain, improving its performance on multi-step reasoning tasks. Wei et al.\citep{wei2022chain} first introduced CoT prompting method in LLMs, improving model performance on arithmetic, commonsense, and logical reasoning tasks. Kojima et al.\citep{kojima2022large} further proposed the Zero-shot CoT prompting method, which only requires adding the cognitive trigger instruction "Let's think step by step" to the prompt to activate the model's internal serialization and logical inference pathways, successfully guiding its reasoning pattern.

In multimodal scenarios, the core challenge for CoT prompting lies in providing a guiding framework to synergistically process and transform information from different modalities to bridge the representation gap. The BBA method\citep{zhao2024bba} guides a vision-language model to create separate reasoning chains for visual representations and Domain-Specific Language (DSL) representations, and then integrates the information from different modalities by aligning inconsistencies; this decoupling-alignment strategy reduces cognitive load and problem complexity. CCoT\citep{mitra2024compositional} first utilizes a Multimodal Large Language Model to generate a scene graph, which is then embedded into the prompt to generate an answer. The IoT prompting method by Zhou et al.\citep{zhou2024image} guides the model to automatically design visual information extraction operations based on the image and question, acquiring visual auxiliary information for each reasoning step. This transforms the model from a passive receiver of information into an active agent that acquires information, making the reasoning results more credible. Furthermore, prompting methods such as DDCoT\citep{zheng2023ddcot}, MI-CoT\citep{anand2024mm}, and ICoT\citep{gao2025interleaved} incorporate visual cues in the prompts, compelling the model to trace back and verify visual evidence at each step of language generation. This achieves a continuous grounding of linguistic symbols with visual information, effectively suppressing model hallucinations.s.

\subsubsection{Search Strategies}
Autoregressive generation models, during inference, follow a greedy strategy, meaning they select the locally optimal next token at each step. This decision-making mechanism is highly susceptible to compounding errors, where minor early deviations are continuously amplified in subsequent steps, ultimately deviating from the correct reasoning trajectory. Search strategies in CoT reasoning aim to systematically overcome this limitation. The core idea is to expand the single, fragile reasoning path into an explorable solution space at the cost of increased computation, thereby finding a more optimal reasoning path globally and improving the correctness and stability of the output. As shown in subfigure (b) of Figure~\ref{fig_tts}, compared to single-path reasoning, search strategies enable the model to more comprehensively explore plausible reasoning chains by evaluating and selecting from multiple branches, enhancing its robustness and generalization capabilities in multi-step reasoning tasks. The search process for MCoT is more complex than for its textual counterpart, as it requires joint decision-making across dimensions such as modal attention, cross-modal alignment, and perceptual understanding, resulting in a higher-dimensional search space. Common search strategies include Beam Search, Monte Carlo Tree Search (MCTS), and Best-of-N (BoN) search\citep{amini2024variational, wang2024improved}.

\textbf{Beam Search:} As a widely used heuristic search algorithm, its core idea is to retain a fixed number (known as the "beam width") of the most probable candidate sequences at each step of the generation process and to expand upon them in subsequent steps. Theoretically, it is a practical compromise between greedy search and exhaustive search, mitigating the fragility of a single path through limited parallel exploration. However, its heuristic assumption may lead to missing the global optimum due to short-sightedness. In the multimodal context, to balance the granularity and complexity of the search, Xu et al.\citep{xu2024llava} proposed a stage-level beam search method, which executes the search by leveraging the phased output characteristic of LLaVA-CoT, with its performance improving as the beam width increases.

\textbf{Monte Carlo Tree Search:} MCTS is a decision-making algorithm based on random sampling and tree-based search, comprising four steps: selection, expansion, simulation, and backpropagation. Compared to the fixed-width exploration of beam search, the core advantage of MCTS lies in its asymmetric tree growth strategy, which can intelligently allocate the computational budget to more deeply explore promising reasoning branches, embodying the exploration-exploitation trade-off.

To address the reliability issue of reasoning path expansion in MCoT, AR-MCTS\citep{dong2024progressive} integrates MCTS with an active retrieval mechanism to achieve automatic generation of step-by-step annotations. Its innovation lies in combining internal contemplation (MCTS) with external knowledge seeking (active retrieval), ensuring the search process is both logically coherent and factually grounded, thus enhancing the reliability of the exploration. Similarly, to tackle the search effectiveness and efficiency problems in multimodal search methods, Yao et al.\citep{yao2024mulberry} proposed CoMCTS, which introduces collective learning into the tree search. It utilizes the collective knowledge of multiple models to collaboratively hypothesize, search, and identify effective reasoning paths through iterative operations, thereby smoothing out the cognitive biases of a single model and improving search coverage and the probability of finding the optimal solution.

To balance the performance and efficiency of AR-MCTS and CoMCTS, Wu et al.\citep{wu2025boosting} proposed AStar, a multimodal automatic structured thought paradigm based on MCTS. Its contribution is to elevate the search object from low-dimensional token sequences to the level of high-dimensional, pre-defined cognitive actions, allowing for the derivation of high-level cognitive reasoning patterns with greater efficiency. Furthermore, Sun et al.\citep{sun2025mm} proposed a simulation-based tree search algorithm. Considering that MLLMs are affected by hallucinations and struggle to produce reliable rewards, this method employs a simulation-based reward mechanism to replace the traditional one in MCTS, providing a more reliable signal for the value assessment (backpropagation step) of MCTS and enhancing the robustness of the search process.

\textbf{Best-of-N Search:} The basic idea of Best-of-N is that the model generates N possible outputs, and then the optimal solution is selected from them based on a preset evaluation criterion. This method increases the likelihood of obtaining a high-quality output by increasing the number of candidate solutions, making it suitable for tasks with high requirements for the quality of the generated results. The SPECULATIVE REJECTION method\citep{sun2024fast} improves upon Best-of-N by halting the generation of low-scoring candidate answers, thereby significantly reducing the model's demand for computational resources. The TreeBoN method\citep{qiu2024treebon} merges a speculative tree search strategy with the Best-of-N method, achieving a balance between high output quality and efficient inference by iteratively expanding branches and pruning low-quality answers. Lin et al.\citep{lin2025investigating} systematically studied inference-time scaling methods like Best-of-N, demonstrating that the performance of MCoT reasoning consistently surpasses that of text-only reasoning.

\textbf{Other Search Methods:} Some works employ a combination of multiple search strategies to enhance search stability and accuracy or propose new Process Reward Models (PRMs) to guide the tree search. 

For challenging multimodal mathematical reasoning tasks, the AtomThink framework\citep{xiang2024atomthink} combines four search strategies and a policy reward model. It divides the search into path search (using majority voting and Best-of-N to aggregate scores and find the optimal solution) and step-wise search (gradually expanding the atomic action sampling space from an initial path, using beam search and greedy strategies to prune low-quality paths). To generate model answers with better visual understanding, Wang et al.\citep{wang2024scaling} proposed VisVM to serve as a signal to guide the model's search during inference. The innovation of VisVM lies in introducing a more forward-looking guidance signal, providing long-term value to avoid local optima. 

Additionally, to address the problems of granularity mismatch and interference from irrelevant information in in-context learning, Zhang et al.\citep{zhang2025booststep} proposed BoostStep, which improves reasoning accuracy through step-level alignment in in-context learning and can be seamlessly integrated with CoT and tree search algorithms. To mimic the interleaved nature of the human visual-language reasoning process, Wang et al.\citep{wang2025visuothink} proposed the VisuoThink reasoning framework, which dynamically integrates visual and textual information. Its core is a predictive rolling search mechanism that fuses visual-textual thought, enabling timely corrections by predicting intermediate state results, thereby enhancing the model's reasoning performance.

\subsubsection{Self Refinement}
In recent years, a significant body of work\citep{dhuliawala2023chain, gou2023critic, paul2023refiner, madaan2023self, li2023making, shinn2023reflexion} has demonstrated that large models possess self-correction capabilities; that is, they can improve the quality of their answers through reflection. This paradigm transforms the model's generation process from a linear "think-output" pattern to an iterative "think-output-review-refine" loop, optimizing the reasoning path through an internal feedback mechanism without updating the model's parameters. As illustrated in subfigure (c) of Figure~\ref{fig_tts}, after generating an initial response, the model reflects upon the original input and its initial answer to generate feedback. Finally, the model refines the initial response based on this feedback. Generally, self-optimization can rely on the model's internal prompting and external information.

To address the issue that MLLMs struggle to learn effectively from erroneous reasoning, R\textsuperscript{3}V\citep{cheng2024vision} introduces self-correction and self-selection, enabling the model to amend flawed reasoning processes and arrive at the correct answer by comparing candidates. EVLM\citep{khalid2024evlm}, on the other hand, introduces a multi-step reflective reasoning framework that includes initial reasoning, intermediate output generation, and reflective refinement, allowing the model to internalize and generalize transformation principles through iterative reasoning. The essence of such methods lies in transforming a one-shot generation process, which may get stuck in a local optimum, into an iterative optimization process where the model progressively approaches a reasonable answer through self-critique. Furthermore, to solve the lack of visual memory and multi-step visual logic modeling capabilities in GUI agent reasoning, InfiGUIAgent\citep{liu2025infiguiagent} utilizes a cyclical process of Expectation-Reflection reasoning to enhance the GUI agent's behavioral consistency and self-correction abilities.

\subsubsection{Knowledge Enhancement}
As shown in subfigure (d) of Figure~\ref{fig_tts}, some recent work enhances the reasoning capabilities of LLMs by introducing external knowledge and integrating it with the input. Such methods aim to overcome the static nature (inability to be updated in a timely manner) and unreliability (the hallucination problem) of the model's parameterized knowledge, enabling the model to utilize more accurate and timely external facts to substantiate its reasoning process. RAGAR\citep{khaliq2024ragar} and RMR\citep{tan2024retrieval} utilize the Retrieval-Augmented Generation (RAG) method to incorporate domain-specific knowledge during the reasoning process to enhance the model's reasoning capabilities. The MR-MKG method\citep{lee2024multimodal} significantly enhances the reasoning capabilities of MLLMs by introducing Multimodal Knowledge Graphs (KGs) to learn rich semantic knowledge across modalities.

In the context of MCoT reasoning, the KAM-CoT framework\citep{mondal2024kam} integrates CoT reasoning, knowledge graphs, and multimodal information. By introducing external knowledge from knowledge graphs during the reasoning process, it achieves a deeper contextual understanding, thereby reducing model hallucinations and improving the quality of the answers. Furthermore, Liu et al.\citep{liu2023retrieval} proposed a retrieval-augmented CoT reasoning method that utilizes cross-modal and intra-modal similarities to dynamically select demonstration examples, which are then structured into the CoT paradigm and included as part of the input.

\subsubsection{Agent Assistance}
Recently, researchers have explored the effectiveness of multi-agent collaboration for model reasoning. This paradigm decomposes complex cognitive tasks into multiple specialized sub-roles, such as a planner, an executor, a verifier, or an evaluator, which are undertaken by different agents. Through structured interaction among them, this approach simulates human collaborative intelligence. As illustrated in subfigure (e) of Figure~\ref{fig_tts}, each large model (agent) is assigned a specific role and iteratively refines its output through structured interactions with other agents.

In multimodal scenarios, Elhenawy et al.\citep{elhenawy2024visual} proposed a multi-agent method to solve the Traveling Salesperson Problem (TSP) and the multi-TSP (mTSP). Their proposed "Multi-Agent 1" includes an initializer, a critic, and a scorer agent, while "Multi-Agent 2" comprises an initializer and a critic agent; both multi-agent frameworks significantly improved the quality of problem-solving. Insight-V\citep{dong2025insight} employs a multi-agent system that breaks down the solution process into two steps: reasoning (generating a detailed reasoning process) and summarizing (identifying key information and selectively answering). This division of labor reduces the cognitive load on a single agent, allowing it to focus on its core task, thereby enhancing overall efficiency and output quality. Additionally, the works of MultiMedRes\citep{gu2024inquire}, InfiGUIAgent\citep{liu2025infiguiagent}, and Zhai et al.\citep{zhai2024fine} utilize a single agent to enhance the model's visual reasoning capabilities.

\subsection{Theoretical Mechanisms Analysis}
In the preceding sections, this paper has systematically reviewed the implementation paths of MCoT methods—that is, the "how"—and correspondingly explained their underlying mechanisms. To more systematically and comprehensively answer the core question of "why MCoT is effective," this section will delve into an in-depth analysis of three key aspects of its operation. First is how the model receives and comprehends the input multimodal information to mitigate the representation gap. Second is how the model utilizes this information to generate a step-by-step reasoning process, thereby reducing compounding errors. Finally, the third aspect is how the model optimizes its intermediate reasoning steps in the form of process supervision during both training and inference. Through a discussion of each of these three aspects, this section aims to analyze the intrinsic mechanism by which CoT enhances the reasoning performance of MLLMs.

\subsubsection{Information Representation}
When fusing heterogeneous modal information, MLLMs face the fundamental challenge of the modality representation gap—that is, how to align high-dimensional, continuous visual signals with discrete, symbolic linguistic logic. This directly impacts the model's subsequent reasoning outcomes.

To address this problem, CoT makes the visual perception process explicit by compelling the model to generate intermediate text, which translates key objects, relationships, and other information from the image into linguistic descriptions. This process not only creates a shared semantic space for subsequent logical operations but, more importantly, it provides anchor points for each step of the reasoning to trace back and verify visual evidence. This achieves a continuous grounding of linguistic symbols to visual facts, thereby reducing the representation gap. This step-by-step verification mechanism establishes a soft constraint within the model's feature space, promoting the creation of a precise mapping between low-level perceptual features and high-level semantic representations. Specifically, methods such as ICoT\citep{gao2025interleaved} innovatively incorporate the image itself as part of a reasoning step, while CoS\citep{liu2024chain} guides the model to focus on key regions within the image, and the IoT method\citep{zhou2024image} even enables the model to actively plan the extraction of visual information. The successful application of these methods validates the effectiveness of using explicit intermediate steps to enhance cross-modal understanding and alignment, which in turn significantly improves the factual consistency of the model's answers and effectively suppresses the generation of model hallucinations.

\subsubsection{Structured Reasoning}
CoT reshapes the model's reasoning paradigm. Its primary mechanism is task decomposition, which breaks down a complex, one-step reasoning task into a series of simpler, more controllable logical sub-steps. This approach significantly reduces the model's cognitive load at any single decision point and effectively suppresses the compounding errors that readily occur in autoregressive generation. In practice, this decomposition process prompts the model to shift from the fast, intuitive "System 1" mode of thinking to a slower, more deliberate, and logically rigorous "System 2" mode. As demonstrated by Kojima et al..\citep{kojima2022large}, even a simple meta-instruction can effectively trigger this mode switch, thereby activating the latent reasoning capabilities the model has already acquired during pre-training.

Furthermore, this explicit, modular reasoning structure provides the necessary foundation for more advanced computational paradigms. It opens up a systematically explorable solution space for advanced topological structures like Tree of Thoughts\citep{long2023large, yao2023tree} and Graph of Thoughts\citep{besta2024graph, lei2023boosting}, as well as for search strategies such as Beam Search\citep{xu2024llava} and Monte Carlo Tree Search\citep{dong2024progressive, yao2024mulberry}. It also provides the basis for collaborative, division-of-labor reasoning in agent-assisted frameworks like Insight-V\citep{dong2025insight}. Concurrently, at each step of the reasoning, methods such as KAM-CoT\citep{mondal2024kam} can also precisely inject external knowledge through knowledge augmentation, further enriching and calibrating the reasoning process.

\subsubsection{Process Supervision}
During both training and inference, the CoT paradigm transforms the way models are supervised and optimized. Traditional multimodal models often employ outcome-based supervision, meaning only the final answer generated by the model is evaluated. In contrast, the explicit multi-step reasoning of CoT provides the formal basis for a more robust form of process-based supervision, where the supervisory signal extends throughout the entire process of both training and inference.

In the post-training stage, the focus of supervision shifts from the correctness of the outcome to the rationality of the process. Datasets with Chain-of-Thought annotations, such as Visual CoT\citep{shao2024visual} and LLaVA-CoT-100K\citep{xu2024llava}, provide the model with detailed reasoning paths, enabling it to perform efficient behavioral cloning through supervised fine-tuning or to undergo goal-oriented optimization based on reward signals via reinforcement learning methods.

During the inference stage, the supervision mechanism manifests in a more dynamic and immediate manner. Chain-of-Thought prompting\citep{wei2022chain, kojima2022large} acts as a non-parametric, on-the-fly form of soft supervision, efficiently guiding the model into a structured reasoning mode. More advanced guidance mechanisms, such as the Process Reward Model proposed by VisVM\citep{wang2024scaling}, or the "critic" role in the multi-agent framework employed by Elhenawy et al.\citep{elhenawy2024visual}, constitute external dynamic supervisory signals at inference time. Meanwhile, self-optimization mechanisms, as represented by R\textsuperscript{3}V\citep{cheng2024vision} and EVLM\citep{khalid2024evlm}, embody a higher-order form of internal dynamic supervision. In this mechanism, the model acts as its own critic, forming an effective closed feedback loop by reflecting on, reviewing, and generating corrective instructions for the intermediate steps it has already produced. Whether originating externally or internally, these dynamic supervisory signals collectively provide fine-grained, real-time calibration of the model's reasoning behavior.

\section{Evaluation:How to Assess CoT-MLLMs?}
\label{evaluation}
\begin{figure}[h]
\begin{center}
\includegraphics[width=1.0\textwidth]{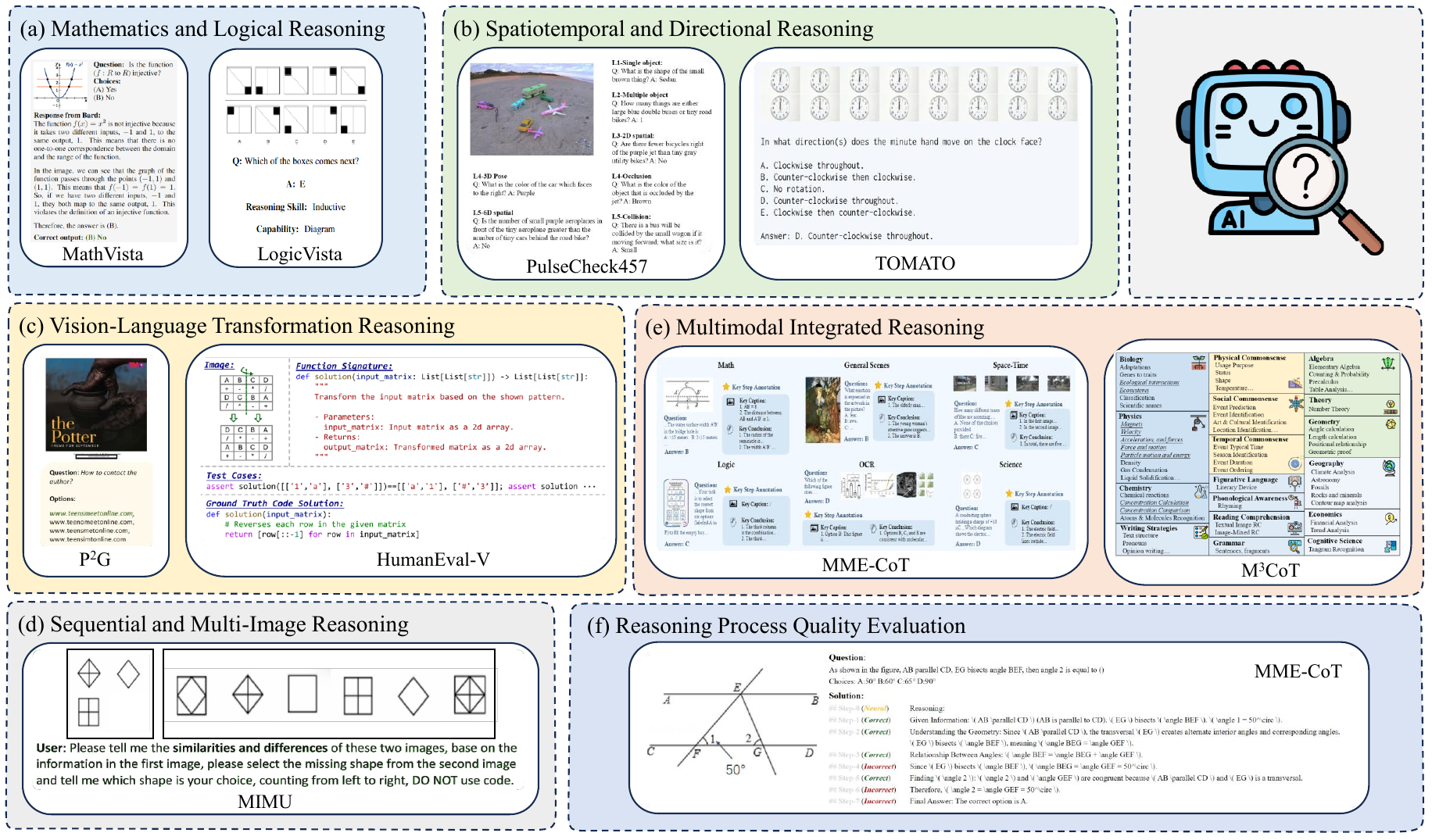}
\end{center}
\caption{Summary of evaluation benchmarks for M-CoT: (a) mathematical and logical reasoning, (b) spatiotemporal and directional reasoning, (c) vision-language transformation reasoning, (d) sequential and multi-image reasoning, (e) multimodal integrated reasoning, (f) reasoning process quality evaluation}
\label{fig_evaluation}
\end{figure}

Due to the distinctive nature of M-CoT reasoning, specialized evaluation benchmarks are required to assess the performance of such models. We conducted a comprehensive survey and summary of existing benchmarks, which are presented in Table~\ref{table_benchmark}. In addition, Figure~\ref{fig_evaluation} provides an intuitive comparative overview of these six categories of evaluation benchmarks, along with representative examples for each category.

\subsection{Mathematical and Logical Reasoning}
As illustrated in subfigure (a) of Figure~\ref{fig_evaluation}, this category of benchmarks focuses on evaluating models' understanding and reasoning capabilities in mathematics and logical relations. These benchmarks typically require models to extract mathematical information from visual inputs, identify logical patterns, apply mathematical principles to solve problems, and perform formal reasoning. They assess not only the accuracy of final answers but also the soundness of the reasoning process and intermediate thinking steps exhibited by the model.

Benchmarks such as MathVista\citep{lu2023mathvista}, MATH-Vision\citep{wang2024measuring}, We-Math\citep{qiao2024we}, and CMM-Math\citep{liu2024cmm} integrate visual information into mathematical reasoning tasks. MathVerse\citep{zhang2024mathverse} aims to evaluate models’ reasoning abilities on math problems involving charts and adopts a chain-of-thought evaluation strategy, leveraging GPT-4(V) for fine-grained assessment of generated answers. ErrorRadar\citep{yan2024errorradar} introduces a new multimodal error detection task in a formalized manner, which includes two subtasks: error step identification and error categorization, thus providing a comprehensive framework for assessing complex mathematical reasoning abilities in multimodal large language models.

Additionally, LogicVista\citep{xiao2024logicvista} targets the evaluation of comprehensive logical reasoning in visual environments, while MME-Reasoning\citep{yuan2025mme} offers a thorough assessment of logical reasoning capabilities, encompassing inductive, deductive, and abductive reasoning. EnigmaEval\citep{wang2025enigmaeval} explores models' abilities to integrate implicit knowledge and perform multi-step deductive reasoning by presenting puzzles of varying complexity. MuCR\citep{li2024multimodal}, a novel multimodal causal reasoning benchmark, leverages synthetic twin image-text pairs to challenge models in identifying causal relationships across different modalities.

\subsection{Spatiotemporal and Directional Reasoning}
This category of benchmarks is designed to evaluate models’ understanding and reasoning capabilities regarding spatial relationships, temporal sequences, and directional concepts. As shown in subfigure (b) of Figure~\ref{fig_evaluation}, these tasks require models to infer topological relationships between objects within visual scenes, comprehend the continuity of events over time, and interpret directional instructions.

PulseCheck457\citep{wang2025pulsecheck457} offers a comprehensive framework for assessing 6D spatial reasoning across varying levels of complexity. It includes seven question types and spans five difficulty levels, ranging from basic single-object recognition to advanced 6D spatial reasoning tasks. GSR-BENCH\citep{rajabi2024gsr} extends the What'sUp dataset and proposes a novel integrated evaluation method for understanding spatial relations. CDR\citep{yin2025multimodal} targets the assessment of directional reasoning in multimodal large language models. It includes images annotated with spatial directions (e.g., up, down, left, right) and compass directions (e.g., north, south, east, west), thereby addressing a previously underexplored area in directional reasoning.For temporal reasoning, TOMATO\citep{shangguan2024tomato} introduces a new benchmark to assess models' ability to reason about temporal dynamics in video understanding. In the context of autonomous driving, DriveLMM-o1\citep{ishaq2025drivelmm} provides a benchmark with over 4,000 visual question-answering instances, covering perception, prediction, and planning tasks. Each instance includes a detailed step-by-step reasoning process.

\subsection{Vision-Language Transformation Reasoning}
As illustrated in subfigure (c) of Figure~\ref{fig_evaluation}, this category of benchmarks is designed to assess a model's ability to convert visual content into structured linguistic forms, such as code or textual descriptions. These tasks evaluate the model's capabilities in cross-modal understanding, information extraction, and representational transformation, requiring it to establish semantic mappings between visual elements and linguistic symbols.

HumanEval-V\citep{zhang2024humaneval} employs a novel code generation task to comprehensively evaluate a model's chart comprehension capabilities. It comprises six task types, each featuring carefully curated charts along with function signatures and test cases. Plot2Code\citep{wu2024plot2code} provides a comprehensive visual-to-code benchmark that includes high-quality matplotlib charts collected from the matplotlib gallery, accompanied by the corresponding source code and descriptive instructions summarized by GPT-4. Similarly, Plot2XML\citep{cui2025draw} offers a benchmark for reconstructing scientific plots into editable XML code. P\textsuperscript{2}GB\citep{chen2024plug} quantitatively evaluates the visual reasoning abilities of multimodal large language models in rich-text or high-resolution image contexts. Current benchmarks for multimodal chain-of-thought (CoT) reasoning predominantly follow the conventional paradigm of multimodal input paired with textual output. To address this limitation, Cheng et al.\citep{cheng2025comt} proposed CoMT, which requires both multimodal input and multimodal reasoning output, aiming to emulate human behavior in integrating visual operations during the reasoning process.

\subsection{Sequential and Multi-Image Reasoning}
This category of benchmarks is designed to evaluate a model’s ability to process temporal data and multi-image inputs. As illustrated in subfigure (d) of Figure~\ref{fig_evaluation}, these tasks require the model to establish associations across multiple visual inputs, track changes, understand event progression within image sequences, and identify relevant patterns.

MIMU\citep{zhang2025benchmark} is developed to assess the performance of multimodal large language models in multi-image input scenarios. It focuses on the model’s ability to perceive fine-grained visual details, with an emphasis on two specific tasks: image-to-image matching and multi-image-to-text matching.

\subsection{Multimodal Integrated Reasoning}
As illustrated in subfigure (e) of Figure~\ref{fig_evaluation}, this category of benchmarks is designed to comprehensively evaluate models’ Chain-of-Thought reasoning capabilities across cross-domain and multimodal tasks. These benchmarks typically integrate various cognitive components, requiring models to handle multimodal inputs, synthesize heterogeneous information sources, and apply reasoning skills in complex scenarios.

MME-CoT\citep{jiang2025mme}, the first comprehensive benchmark for evaluating the CoT reasoning capabilities of multimodal large language models, spans six domains: mathematics, science, optical character recognition (OCR), logic, spatiotemporal reasoning, and general scenarios. M\textsuperscript{3}CoT\citep{chen2024m3cot} takes the first step toward addressing multi-domain, multi-step, and multimodal scenarios in multimodal CoT reasoning. Similarly, EMMA\citep{hao2025can} and VRC-Bench[84] offer multi-step benchmarks that assess CoT reasoning across modalities and domains. In addition, R1-Onevision-Bench\citep{yang2025r1} spans multiple subjects—including mathematics, physics, chemistry, biology, and logical reasoning—covering examination content from middle school to university and beyond. NPHardEval4V\citep{fan2024nphardeval4v} serves as a dynamic benchmark designed to isolate the effects of various factors such as image recognition and instruction following from overall model performance, thereby focusing the evaluation on reasoning capability. Likewise, VisualPuzzles\citep{song2025visualpuzzles} reduces the reliance on domain-specific knowledge while increasing reasoning complexity to better assess a model's multimodal reasoning performance.

\subsection{Reasoning Process Quality Evaluation}
This category of benchmarks focuses on evaluating the quality of the model's reasoning process, such as consistency, interpretability, and accuracy. As illustrated in subfigure (f) of Figure~\ref{fig_evaluation}, these tasks typically require the model to provide a reasoning chain, explain its decision-making process, and maintain logical coherence throughout multi-step reasoning. Such benchmarks are of significant value for assessing the transparency of a model’s cognitive process and the reliability of its reasoning foundations.

MPBench\citep{xu2025mpbench} is a comprehensive multi-task benchmark that systematically evaluates the effectiveness of Process Reward Models (PRMs) through three evaluation paradigms: step correctness identification, answer aggregation, and reasoning path search. MiCEval\citep{zhou2024miceval} assesses the correctness of reasoning chains by evaluating both the quality of model-generated descriptions and each reasoning step. The description evaluation focuses on the accuracy of image descriptions, while the reasoning step evaluation emphasizes the quality of each specific inference. VisualProcessBench\citep{wang2025visualprm}, a benchmark with manually annotated step-by-step correctness labels, is primarily used to assess the ability of PRMs to detect incorrect reasoning steps in multimodal reasoning tasks. In addition, MMIR\citep{yan2025multimodal} and the work by Jia et al.\citep{jia2025exploring} focus on identifying inconsistency issues in multimodal reasoning, offering valuable insights for future improvements of multimodal large language models.

\subsection{Evaluation Metrics}
At present, most existing works\citep{xiang2024atomthink, thawakar2025llamav, li2024multimodal, shangguan2024tomato, wu2024plot2code} assess the reasoning capabilities of LLMs and MLLMs by directly evaluating the accuracy of their final answers, using this as a proxy for chain-of-thought reasoning performance. Alternatively, some studies propose task-specific evaluation metrics. However, these approaches often suffer from limitations such as a lack of comprehensive evaluation and inability to accurately assess complex visual reasoning tasks. To address these issues, MME-CoT\citep{jiang2025mme} proposed the first dedicated evaluation framework for assessing CoT reasoning capabilities in MLLMs, which covers three key aspects: CoT quality, CoT robustness, and CoT efficiency. In addition, MMMR\citep{tie2025mmmr} introduced a modular Reasoning Trace Evaluation Pipeline (RTEP) that provides a holistic assessment of reasoning steps based on three dimensions: relevance to the question, relevance to the answer, and internal consistency across reasoning steps.

\begin{table}[H]
\caption{Summary of evaluation benchmarks for CoT-MLLMs. "MC" and "Open" refer to multiple-choice and open-ended answer formats, while "T", "I", "V", and "PC" represent Text, Image, Video and Point Cloud, respectively.}
\label{table_benchmark}
\centering
\resizebox{\textwidth}{!}{%
\begin{tabular}{ccccccc}
\toprule 
\textbf{Benchmark} & \textbf{Year} & \textbf{Task} & \textbf{Modality} & \textbf{Metric} & \textbf{Format} & \textbf{Size} \\
\midrule 
\multicolumn{7}{c}{\textit{Mathematical and Logical Reasoning}} \\
MathVista\citep{lu2023mathvista} & 2024 & Math QA & T, I & Accuracy & MC, Open & 6,141\\
MATH-Vision\citep{wang2024measuring} & 2024 & Math QA & T, I & Accuracy & MC, Open & 3,040\\
We-Math\citep{qiao2024we} & 2024 & Math QA & T, I & Custom Metric & MC & 6.5K\\
LogicVista\citep{xiao2024logicvista} & 2024 & Logical Reasoning & T, I & Accuracy & MC, Open & 448\\
MathVerse\citep{zhang2024mathverse} & 2024 & Math QA & T, I & Accuracy, CoT-Evaluation & MC, Open & 15,672\\
ErrorRadar\citep{yan2024errorradar} & 2024 & Error Detection & T, I & Accuracy & Open & 2,500\\
CMM-Math\citep{liu2024cmm} & 2024 & Math QA & T, I & Accuracy & MC, Open & 28,069\\
MuCR\citep{li2025multimodal} & 2025 & Causal Reasoning & T, I & Custom Metric & Open & ——\\
EnigmaEval\citep{wang2025enigmaeval} & 2025 & Puzzle Solving & T, I & Accuracy & Open & 1,184\\
MME-Reasoning\citep{yuan2025mme} & 2025 & Logical Reasoning & T, I & Custom Metric & MC, Open, Rule-based & 1,188\\
\midrule 
\multicolumn{7}{c}{\textit{Spatiotemporal and Directional Reasoning}} \\
CDR\citep{yin2025multimodal} & 2024 & Directional Reasoning & T, I & Accuracy & MC & 100K\\
GSR-BENCH\citep{rajabi2024gsr} & 2024 & Spatial Reasoning & T, I & Accuracy & MC & 4,958\\
TOMATO\citep{shangguan2024tomato} & 2024 & Temporal Reasoning & T, V & Custom Metric & MC & 1,484\\
PulseCheck457\citep{wang2025pulsecheck457} & 2025 & Spatial Reasoning & T, I & RPDR & Open & ——\\
DriveLMM-o1\citep{ishaq2025drivelmm} & 2025 & Autonomous Driving & T, I, PC & Custom Metric & MC, Open & ——\\
\midrule
\multicolumn{7}{c}{\textit{Vision-Language Transformation Reasoning}} \\
Plot2Code\citep{wu2024plot2code} & 2024 & Visual Coding & T, I & Custom Metric & Open & 132\\
P\textsuperscript{2}G\citep{chen2024plug} & 2024 & Visual Reasoning & T, I & Accuracy & MC & 2,130\\
HumanEval-V\citep{zhang2024humaneval} & 2025 & Visual Coding & T, I & pass@k & Open & 100K\\
CoMT\citep{cheng2025comt} & 2025 & Multi-task & T, I & Accuracy, Macro-F1 & MC & 3,853\\
Plot2XML\citep{cui2025draw} & 2025 & Visual Coding & T, I & Custom Standard & Open & 247\\
\midrule
\multicolumn{7}{c}{\textit{Sequential and Multi-Image Reasoning}} \\
MIMU\citep{zhang2025benchmark} & 2025 & Multi-task & T, I & Accuracy & Open & ——\\
\midrule
\multicolumn{7}{c}{\textit{Multimodal Integrated Reasoning}} \\
NPHardEval4V\citep{fan2024nphardeval4v} & 2024 & VQA & T, I & Custom Metric & Open & ——\\
M\textsuperscript{3}CoT\citep{chen2024m3cot} & 2024 & VQA & T, I & Accuracy & MC & 11,459\\
EMMA\citep{hao2025can} & 2025 & VQA & T, I & pass@k & MC, Open & 2,788\\
VRC-Bench\citep{thawakar2025llamav} & 2025 & Multi-task & T, I & Custom Multi-metric & MC, Open & ——\\
MME-CoT\citep{jiang2025mme} & 2025 & VQA & T, I & Quality, Robustness, Efficiency & MC, Open & 1,130\\
R1-Onevision-Bench\citep{yang2025r1} & 2025 & VQA & T, I & Accuracy & MC, Open & 942\\
VisualPuzzles\citep{song2025visualpuzzles} & 2025 & VQA & T, I & Accuracy & MC & 1,168\\
\midrule
\multicolumn{7}{c}{\textit{Reasoning Process Quality Evaluation}} \\
MiCEval\citep{zhou2024miceval} & 2025 & VQA & T, I & Custom Metric & Open & 2,130\\
MMIR\citep{yan2025multimodal} & 2025 & Consistency Reasoning & T, I & Accuracy & MC, Open & 534\\
VisualProcessBench\citep{wang2025visualprm} & 2025 & Science QA & T, I & Macro-F1 & MC, Open & 2,866\\
\bottomrule 
\end{tabular}%
}
\end{table}

\section{Application: Where can CoT-MLLMs be applied?}
\label{application}
\begin{figure}[h]
\begin{center}
\includegraphics[width=1.0\textwidth]{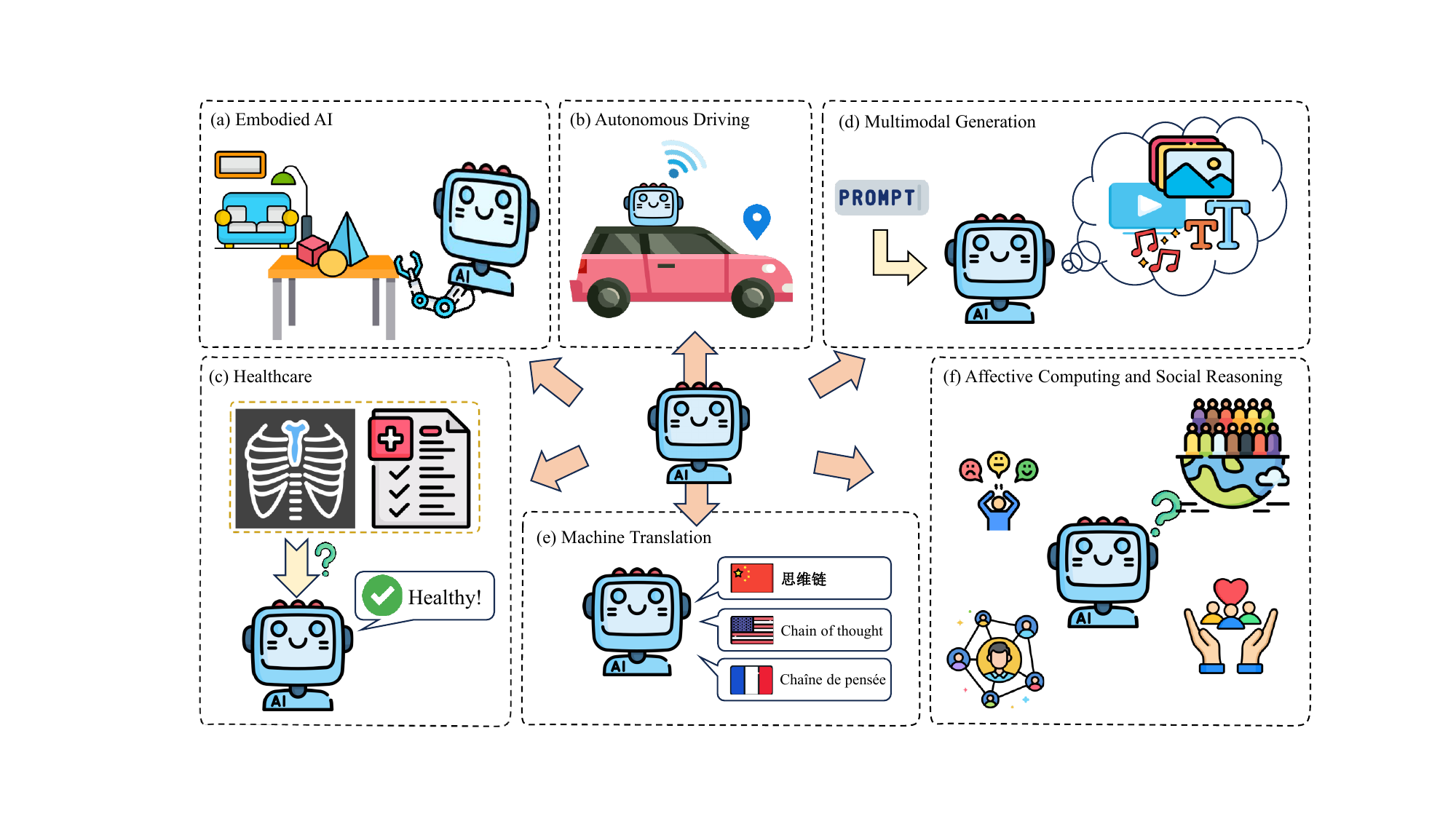}
\end{center}
\caption{Summary of applications of M-CoT: (a) embodied AI, (b) autonomous driving, (c) healthcare, (d) multimodal generation, (e) machine translation, (f) affective computing and social reasoning}
\label{fig_application}
\end{figure}

By introducing a step-by-step reasoning mechanism, M-CoT significantly enhances the reasoning capabilities and interpretability of multimodal systems in complex tasks. Compared with traditional multimodal approaches, CoT enables the staged integration of information from different modalities during the reasoning process, maintaining interpretability and coherence, thereby exhibiting greater robustness and generalization in complex environments. Leveraging these advantages, M-CoT demonstrates broad application potential in fields such as embodied intelligence, autonomous driving, healthcare, multimodal generation, machine translation, and social computing. It has emerged as a promising direction for advancing intelligent systems toward higher-level cognitive reasoning. A summary of these applications is presented in Figure~\ref{fig_application}.

\subsection{Embodied AI}
Embodied Intelligence, as a crucial pathway toward achieving Artificial General Intelligence (AGI), emphasizes embedding artificial intelligence systems within physical carriers such as robots, thereby endowing them with capabilities for environmental perception, autonomous learning, and dynamic interaction. As illustrated in subfigure (a) of Figure~\ref{fig_application}, with the recent introduction of M-CoT methods, embodied intelligence has witnessed significant improvements in core competencies—including task planning, spatial reasoning, navigation, and manipulation—when confronting complex multimodal environments.

In the domain of task decomposition and execution, the works by Hao et al.\citep{hao2025embodied} and Embodied Reasoner\citep{zhang2025embodied} demonstrate the feasibility of automatically dividing operation instructions or task goals into concrete, detailed execution steps. Specifically, Hao et al. enhance humanoid robots’ abilities to perform localized operations in unstructured complex scenarios by incorporating foundation models, while Embodied Reasoner targets interactive search tasks, showcasing the agent’s capacity for autonomous goal recognition, reasoning, and decision-making, thereby strengthening adaptability and proactivity in dynamic tasks.

Regarding task planning and hierarchical coordination, EmbodiedGPT\citep{mu2023embodiedgpt} proposes a unified architecture that integrates high-level strategic planning with low-level control execution, enabling embodied agents to accomplish complex tasks through stepwise planning. The MCoT-Memory framework\citep{liangmemory} introduces a memory retrieval–based, scene graph–driven chain-of-thought mechanism that retains high-confidence experiential knowledge to effectively support long-term and complex task planning in dynamic environments. ECoT\citep{zawalski2024robotic} further expands embodied intelligence’s multimodal reasoning capabilities by incorporating chain-of-thought training mechanisms into Vision-Language-Action (VLA) models, allowing for multi-step task reasoning and subtask decomposition with strong generalization on novel tasks.

In spatial reasoning tasks, EMMA-X\citep{sun2024emma} exhibits foresighted action planning and fact-based chain-of-thought structures, enabling more effective prediction of environmental changes and corresponding action strategy formulation. SpatialCoT\citep{liu2025spatialcot} employs a two-stage training paradigm of “bidirectional spatial coordinate alignment + CoT spatial alignment” to achieve deep synergy between spatial information and chain-of-thought logic. EmbodiedVSR\citep{zhang2025embodiedvsr} explicitly models structured knowledge via dynamic scene graphs, granting zero-shot spatial reasoning abilities and enhancing adaptability and robustness in unseen environments.

For navigation tasks, MCoCoNav\citep{shen2025enhancing} introduces multimodal chain-of-thought for multi-robot collaborative semantic navigation, designing strategies that integrate global semantic maps with direct inter-robot communication to significantly improve navigation efficiency while reducing communication costs. NavCoT\citep{lin2025navcot} achieves self-guided navigation decision-making at low cost through parameter-efficient domain-specific training, simultaneously enhancing cross-domain generalization capabilities.

Notably, COUT\citep{zhang2024smartagent} pioneers preliminary exploration into learning personalized embodied agents by incorporating individual factors such as user personality and preferences into the autonomous agent’s learning process, thereby broadening embodied intelligence’s application prospects in personalized human–machine interaction.

In summary, these studies, by integrating multimodal chain-of-thought with large-scale vision-language models (VLMs) or Vision-Language-Action models (VLAs), not only substantially enhance embodied agents’ multi-step reasoning abilities but also improve their generalization and adaptability across diverse environments and tasks. Concurrently, breakthroughs in system architecture, task execution efficiency, and personalization propel embodied intelligence toward increasingly efficient, cost-effective, and personalized intelligent agent paradigms.

\subsection{Autonomous Driving}
In recent years, multimodal large language models have demonstrated significant potential in the field of Autonomous Driving (AD), offering a new paradigm for constructing next-generation end-to-end autonomous driving systems. As shown in subfigure (b) of Figure~\ref{fig_application}, integrating CoT mechanisms into autonomous driving systems not only improves the system’s perception accuracy and decision robustness in complex traffic environments but also effectively reduces computational resource consumption and engineering costs during model development and deployment through reasoning-time computation methods.

In practical system construction, Dolphins\citep{ma2024dolphins}, a conversational driving assistant system, introduces a M-CoT process based on scene context to enhance its understanding and execution of driving instructions. This system is fine-tuned on datasets specifically designed for intelligent driving, significantly improving its performance in intelligent cockpits and human–machine interaction. Agent-Driver\citep{mao2023language} employs a large language model as the core cognitive agent and transforms traditional autonomous driving pipelines by integrating general-purpose toolkits, achieving end-to-end modeling of the task chain from perception and understanding to decision-making.

Regarding data resources and model validation, several key studies have developed open datasets that support multimodal reasoning and end-to-end driving tasks. DriveCoT\citep{wang2024drivecot} trained the baseline DriveCoT-Agent on its constructed dataset, marking the first successful integration of multimodal CoT methods into an end-to-end autonomous driving system. Reason2Drive \citep{nie2024reason2drive} provides over 600,000 video-text pairs containing explicit reasoning processes, greatly advancing research on interpretable reasoning in autonomous driving. DriveLMM-o1\citep{ishaq2025drivelmm} focuses on stepwise visual reasoning tasks, covering three critical aspects in autonomous driving scenes: perception, prediction, and planning, thus offering datasets and benchmarks for systematic evaluation of multimodal reasoning models.

In terms of reasoning frameworks, PKRD-CoT\citep{luo2024pkrd} proposes a unified prompt design framework for multimodal large language models in autonomous driving, guiding models through a human-like stepwise thinking process to make real-time driving decisions, thereby enhancing system adaptability and responsiveness. OpenEMMA\citep{xing2025openemma} integrates CoT reasoning mechanisms with external expert vision models, demonstrating outstanding performance in vehicle trajectory planning tasks, particularly in 3D structural recognition and precise detection of road targets. Meanwhile, the study by Islam et al.\citep{islam2024application} decomposes the autonomous driving decision-making process into three stages: scene understanding, behavior prediction, and decision generation, introducing CoT as the core pathway linking reasoning across these stages, thereby improving the overall decision-making rationality and coherence.

In summary, the deep integration of multimodal chain-of-thought mechanisms with autonomous driving systems significantly enhances the system’s ability to model and understand complex and dynamic traffic environments, while improving the coherence, interpretability, and robustness of the decision-making process. This advancement propels autonomous driving technology toward a higher level of intelligence and represents a crucial step toward achieving human-like driving.

\subsection{Healthcare}
As illustrated in subfigure (c) of Figure~\ref{fig_application}, in the healthcare domain, the introduction of M-CoT techniques provides a unified reasoning framework for integrating multi-source information such as medical imaging, clinical text, and physiological signals. This mechanism simulates the diagnostic and treatment thinking processes of physicians, offering intelligent assistive systems with greater transparency and interpretability in critical tasks including disease diagnosis, treatment planning, and personalized health management, thereby significantly enhancing the practicality and reliability of artificial intelligence in medical scenarios.

In medical image and video analysis, Chain-of-Look\citep{xi2023chain} models the triplet recognition task in endoscopic videos as a visual prompt generation problem by leveraging vision-language models. It explicitly decomposes the task into a sequence of temporally structured video reasoning steps. This approach not only improves diagnostic accuracy but also strengthens the model’s ability to localize key pathological features within image sequences.

For medical visual question answering (VQA) tasks, MedThink\citep{gai2024medthink} proposes simultaneously generating decision outcomes and corresponding reasoning rationales, intuitively demonstrating the system’s reasoning pathways in answering medical questions, thereby enhancing the interpretability of the decision process. MedCoT\citep{liu2024medcot} introduces a hierarchical expert verification CoT approach that simulates multidisciplinary clinical consultation by integrating multiple “expert models” for reasoning and correction, achieving more reliable and clinically consistent answers. Additionally, it employs a multi-expert mixed review mechanism to effectively mitigate uncertainties caused by ambiguity or incomplete information in medical Q\&A.

In the mental health domain, Dai et al.\citep{dai2024interpretable} apply CoT reasoning methods to stress detection systems, mimicking the cognitive and judgment processes of psychological experts. This enables the model to attain higher interpretability and user trust when assessing highly subjective psychological states.

Regarding medical procedural action recognition, TI-PREGO\citep{plini2024ti} combines in-context learning (ICL) with an automatic chain-of-thought (ACoT) mechanism to perform procedural error detection in egocentric medical videos through action recognition and prediction tasks.

Meanwhile, MedVLM-R1\citep{pan2025medvlm} enhances the reasoning ability of vision-language models in medical image analysis tasks by incorporating reinforcement learning, explicitly strengthening the transparency and traceability of the reasoning paths through natural language reasoning.

In summary, in the healthcare field—where decision interpretability and controllability are critically important—multimodal chain-of-thought techniques exhibit promising application prospects and research potential. By simulating medical experts’ decision logic and achieving unified modeling from the perceptual to cognitive levels, these techniques are expected to lay a solid foundation for future development of trustworthy, high-performance human–machine collaborative medical systems.

\subsection{Multimodal Generation}
As shown in subfigure (d) of Figure~\ref{fig_application}, in multimodal generation tasks, MCoT serves as a key mechanism that guides models to progressively complete cross-modal reasoning and content generation, significantly enhancing the coherence, logical consistency, and modal alignment of the generated outputs. By simulating human cognitive pathways in processing complex multimodal information, MCoT methods effectively mitigate common issues in traditional generative models, such as modal bias and insufficient information fusion when handling heterogeneous modalities.

In terms of system architecture, the MINT model\citep{wang2025mint} introduces a parallel ensemble structure of multiple expert modules—MTXpert (Mixture of Transformer Experts)—which efficiently integrates natural language generation capabilities with image understanding, avoiding generation degradation caused by modal conflicts. MINT adopts a multi-stage CoT paradigm of “thinking—reasoning—reflecting” during generation, substantially improving the semantic consistency and visual quality of generated images. Regarding physical consistency modeling, the Phys-AR framework\citep{lin2025reasoning} combines symbolic reasoning with reinforcement learning mechanisms to construct a multimodal generation system endowed with causal understanding capabilities.

Furthermore, in the context of generation authenticity and traceability, the CoT-Finetuned model \citep{agrahari2025tracing} explores the application of chain-of-thought reasoning mechanisms to AI-generated content detection tasks. This model not only determines whether text is generated by large language models but also identifies the specific model used, thereby effectively enhancing the interpretability and discriminative power of detection systems. This approach holds significant importance for maintaining academic integrity, combating misinformation, and advancing AI ethics standards.

In summary, multimodal chain-of-thought techniques play multiple critical roles in multimodal generation tasks: they optimize cross-modal information fusion pathways, improve the logical and visual consistency of generated content, and promote research on the credibility verification and accountability of generated outputs. Looking forward, with the continuous expansion of model scale and modality types, MCoT-based multimodal generation paradigms are expected to exhibit broader application prospects in scenarios such as virtual reality creation, human–computer interaction systems, and personalized educational and entertainment content.

\subsection{Machine Translation}
Compared to traditional neural network-based machine translation paradigms and the recently emerging large language model-based translation approaches, MCoT introduces a novel research direction for machine translation tasks. As illustrated in subfigure (e) of Figure~\ref{fig_application}, this paradigm emphasizes incorporating external modal information such as visual, speech, and environmental context as key auxiliary inputs during the reasoning process. This enables a more systematic understanding of the source language context, cultural background, and implicit intent, thereby generating target language expressions that are more natural, accurate, and culturally adaptive.

CoT-ST\citep{du2024cot} leverages multimodal chain-of-thought by decomposing speech translation into sequential steps of speech recognition and translation, enhancing the natural synergy and complementarity between these subtasks. This leads to significant improvements in translation performance of speech language models (SLMs). IMAGE\citep{chen2024make} integrates multimodal large models with stable diffusion\citep{rombach2022high}-based networks to explicitly generate images for each sentence to be translated, utilizing visual information to boost machine translation performance. Furthermore, in multimodal translation research targeting low-resource languages, Rajpoot et al.\citep{rajpoot2024multimodal} propose a translation strategy that combines image captioning with chain-of-thought reasoning. This method extracts semantic descriptions from images through multiple stages and embeds them into the reasoning path of the language translation process, achieving promising results in English-to-low-resource Indian language multimodal translation tasks.

In summary, multimodal chain-of-thought introduces a processing approach for machine translation that aligns more closely with human cognitive characteristics. By explicitly connecting deep semantic relationships between language and multimodal information through reasoning processes, it comprehensively enhances translation systems in terms of accuracy, cultural adaptability, and contextual modeling.

\subsection{Affective Computing and Social Reasoning}
As shown in subfigure (f) of Figure~\ref{fig_application}, in fields involving human cognition and behavior modeling such as affective computing and social reasoning, MCoT effectively integrates multi-source information including text, images, and speech by introducing interpretable reasoning paths. This significantly enhances models’ overall performance in emotion recognition, affective understanding, and complex social behavior modeling. The reasoning mechanism not only provides more structured decision support but also strengthens the model’s generalization ability to human contexts and social knowledge.

In affective computing, MM-PEAR-CoT\citep{li2025multimodal} is the first to introduce the chain-of-thought reasoning paradigm into multimodal emotion recognition tasks. By fusing textual, visual, and audio information, it alleviates reasoning biases caused by unimodal errors in traditional chain-of-thought methods, thereby achieving more stable and interpretable performance in emotion label classification tasks. Additionally, the Empatheia system\citep{zhang2025towards} integrates multimodal large language models with empathetic reasoning chains, demonstrating more natural, nuanced, and personalized language generation capabilities in empathetic response generation (e.g., mental health Q\&A, emotional counseling), thus advancing the practical application of intelligent systems in social-affective interaction.

Regarding social reasoning, SOCIAL GENOME\citep{mathur2025social} offers the first benchmark framework to evaluate multimodal models’ social reasoning capabilities, featuring fine-grained reasoning steps supplemented with external knowledge concepts for auxiliary assessment. Yu and Luo\citep{yu2024chain} further explore the application of multimodal large language models enhanced with chain-of-thought prompting methods in demographic inference, significantly reducing prediction bias. Moreover, Hu et al.\citep{hu2024multimodal} focus on child physical and mental health protection by applying multimodal chain-of-thought for age-appropriate rating of mobile applications. CPFEND\citep{xu2024multimodal} and RAGAR\citep{khaliq2024ragar} are devoted to improving models’ reasoning and judgment capabilities in misinformation detection and political fact-checking.

In summary, these studies comprehensively demonstrate the substantial potential of multimodal chain-of-thought reasoning mechanisms to enhance models’ structured cognitive abilities, achieve situational awareness, and model social behaviors.

\section{Challenges and Future Directions: What’s Next?}
\label{challenge}
Although current multimodal chain-of-thought approaches have made some progress and demonstrated promising potential, they still face several key issues and challenges that require further exploration and resolution. In this chapter, we summarize and present several challenges and future directions, with the hope of providing insights for related researchers.

\subsection{Robust Reasoning}
In the context of the increasingly complex development of multimodal reasoning models, robustness has become a critical metric for evaluating the practical applicability of these models. Reasoning robustness not only reflects a model’s tolerance to input perturbations but also encompasses its ability to consistently generate stable and trustworthy conclusions when faced with uncertainty, knowledge gaps, and task complexity. Current multimodal chain-of-thought reasoning models still confront numerous challenges that impair their robustness in handling complex tasks, with hallucination phenomena and reasoning chain length control being the most prominent issues.

The hallucination problem in large language models has long troubled researchers. Traditional large language models, due to the absence of explicit reasoning processes, frequently produce contextually inconsistent or factually inaccurate responses, severely limiting their real-world applications. Although chain-of-thought reasoning alleviates hallucinations to some extent, it also introduces side effects such as “overthinking.” Common hallucination mitigation strategies during the reasoning phase include external knowledge-based methods (e.g., retrieval-augmented generation, expert model-assisted detection\citep{zhang2024meter, zhang2025critic}) and internal mechanism-based methods (e.g., self-reflection, self-consistency reasoning). However, these methods mostly operate at the overall answer level, suffering from error accumulation across intermediate reasoning steps. Future research could explore finer-grained optimization at the step level; for instance, HaluSearch\citep{cheng2025think} models answer generation as an explicit stepwise reasoning process using Monte Carlo tree search and leverages step-level reward mechanisms to explore optimal reasoning paths, thereby enhancing reasoning reliability.

Moreover, in multimodal environments, insufficient alignment between different modalities can also trigger hallucinations. For example, Zheng et al.\citep{zheng2024thinking} pointed out that the commonly used “look-then-think” paradigm in current multimodal chain-of-thought methods—where reasoning chains are generated simultaneously with visual input—can be misled by deceptive visual information, resulting in biased reasoning outcomes. Therefore, cross-modal alignment mechanisms in future multimodal chain-of-thought research will become a key focus. For instance, the VIC framework\citep{zheng2024thinking} constructs the reasoning chain based on textual context prior to introducing visual information, thereby reducing cross-modal bias caused by visual interference and improving reasoning accuracy.

Apart from hallucinations, the length of the reasoning chain also directly impacts reasoning robustness. A chain that is too short may lead to insufficient deliberation, affecting the accuracy of the final conclusion; conversely, an excessively long chain can cause information forgetting and overthinking. Hence, designing effective mechanisms for controlling reasoning chain length is a promising research direction. Some existing works have proposed solutions in this regard. For example, Sun et al.\citep{sun2025mitigating} addressed the issue of gradually diminished attention to visual information during the reasoning process in multimodal large language models by proposing the TVC (Take-along Visual Conditioning) strategy, which maintains image information into critical reasoning stages and employs dynamic pruning to compress redundant visual representations, thereby sustaining the model’s focus on visual content. Meanwhile, ThinkEdit\citep{sun2025thinkedit} tackles performance degradation caused by overly short reasoning chains in mathematical problem solving by introducing a weight editing method to mitigate this issue. Yang et al.\citep{yang2025towards} observed that excessively long chains weaken mathematical reasoning capabilities and thus proposed the TOPS (Thinking-Optimal Scaling) strategy, enabling the model to self-optimize and arrive at correct answers with shorter reasoning chains. Furthermore, Yang et al.\citep{yang2025dynamic} proposed a plug-and-play mechanism that detects the model’s behavior at potential reasoning turning points (e.g., “wait” tokens) and dynamically terminates subsequent chain generation when the model shows high confidence in tentative answers, thus enabling more efficient and rational planning of reasoning paths.

\subsection{Secure Reasoning}
The safety and ethical issues of reasoning models are critical topics that urgently need to be addressed. Due to their characteristics of cross-modal alignment and explicit step-by-step reasoning, multimodal reasoning models exhibit unique vulnerabilities, making them more susceptible to malicious attacks and the generation of harmful information. Zhou et al.\citep{zhou2025hidden} evaluated the robustness of mainstream large reasoning models against adversarial attacks such as jailbreaking and prompt injection, finding that open-source models like DeepSeek-R1 and distilled models have security disadvantages. In multimodal scenarios, Fang et al.\citep{fang2025safemlrm} revealed the "Reasoning Tax" phenomenon, wherein a model's acquisition of complex reasoning capabilities can weaken the safety alignment features of the base model. At the same time, the study found that multimodal reasoning models possess a certain capacity for self-correction when faced with unsafe reasoning chains.

Currently, research on specific attack methods and defense mechanisms for multimodal reasoning models is still in its nascent stages, whereas related research on unimodal (text-only) reasoning models or general-purpose multimodal models is relatively more mature. Wang et al.\citep{wang2025safety} categorized the attack methods for large reasoning models into four types: reasoning length attacks, answer correctness attacks, prompt injection attacks, and jailbreak attacks. Reasoning length attacks induce the model to engage in "overthinking"\citep{kumar2025overthink, zaremba2025trading} or "underthinking"\citep{zaremba2025trading} through prompts, where overthinking attacks are essentially a form of Denial of Service (DoS) attack. Answer correctness attacks tamper with intermediate reasoning steps through backdoor attacks\citep{xiang2024badchain, guo2025darkmind, zhao2025shadowcot} or error injection\citep{cui2025process} to compromise the integrity of the results. Prompt injection attacks disguise malicious instructions as normal user input to bypass the model's preset safety protocols\citep{kumar2024strengthening, liu2023prompt}. Lastly, jailbreak attacks use prompts to evade the model's safety restrictions. Some studies\citep{russinovich2025great, ren2025llms, li2024llm} also utilize the multi-turn interaction of CoT to progressively guide the model to generate harmful content.

For vision-language models, Ye et al. classified attack methods into three categories based on the attacker's knowledge of the model architecture: white-box, gray-box, and black-box\citep{ye2025survey}. White-box attacks\citep{qi2024visual, bailey2023image, wang2024white} occur when the attacker has full knowledge of the model's internal architecture, amplifying the attack's effect by injecting adversarial noise into images and modifying text inputs to exploit modal dependencies. Gray-box attacks\citep{zhao2023evaluating, wang2024break} leverage partial knowledge of the model, using open-source vision encoders like CLIP to generate adversarial samples. Black-box attacks\citep{gong2025figstep, ma2024visual} bypass safety restrictions by designing clever prompts without any knowledge of the model's internal information.

In terms of defense mechanisms, current research primarily focuses on three levels: training-time defense, inference-time defense, and external guardrail models\citep{wang2025safety}. During the model alignment training phase, some work is dedicated to constructing Chain-of-Thought safety datasets\citep{jiang2025safechain, zhang2025realsafe, chen2024dress} and performing safety reasoning alignment through supervised fine-tuning\citep{lou2025think, xia2025msr} or reinforcement learning\citep{guan2024deliberative, zhang2025stair}. To avoid the safety alignment tax—where safety training may sacrifice model performance—researchers are exploring the application of techniques at the inference stage, such as safety prompts\citep{wang2024adashield, zhao2024bluesuffix}, the detection and removal of adversarial noise\citep{sun2024safeguarding}, and safe decoding\citep{jiang2025safechain}, to enhance model security. The guardrail model approach\citep{liu2025guardreasoner, wen2025thinkguard} utilizes an independent auditing model to monitor and filter the inputs and outputs of the main model in real-time, thereby achieving security protection without modifying the original model.

In summary, the unique properties of multimodal reasoning present new challenges to model security. Future research needs to construct a multi-level, coordinated defense system that integrates data, models, and reasoning, with the goal of comprehensively enhancing the safety and robustness of these models without sacrificing their core reasoning performance.

\subsection{Omnimodal Reasoning}
The GPT-4o (mini) model demonstrates significant potential of large models in perceiving and processing multimodal data; however, it lacks chain-of-though reasoning capabilities. Currently, multimodal CoT reasoning primarily focuses on bimodal combinations such as text-image \citep{liu2025visual, li2024streetviewllm}, text-video\citep{fei2024video, himakunthala2023let, zhang2024worldqa}, or text-audio\citep{du2024cot, xie2025audio}, with relatively limited modality coverage, making it insufficient to meet more complex and general reasoning scenarios.

Full-modality chain-of-thought reasoning, also referred to as CoT reasoning Omini, aims to extend the unified processing capability of reasoning models across multiple modalities. It enables collaborative handling of text, image, audio, and video data within a single framework, achieving comprehensive and multi-source information fusion and reasoning support. This direction imposes higher requirements on inter-modality representation alignment, information complementarity mechanisms, and unified modeling of reasoning paths. It is one of the key pathways to advance multimodal reasoning paradigms toward artificial general intelligence (AGI).

In this domain, R1-Omini\citep{zhao2025r1} achieved full-modality expansion of CoT reasoning in emotion recognition tasks and was the first to introduce verifiable reward-based reinforcement learning into full-modality large language models for emotion recognition, where text, visual, and audio modalities all played crucial roles.

\subsection{Efficient Reasoning}
Inspired by human intelligence, some researchers analogize direct reasoning and deep reasoning in large language models to the fast, intuitive System 1 reasoning and the slow, deliberate System 2 reasoning, respectively. Although System 2 reasoning can improve model performance, it incurs significant computational costs; in contrast, System 1 reasoning, being intuitive and rapid, offers higher computational efficiency but often at the expense of some performance. Therefore, balancing model performance and computational resources has become a key challenge in achieving efficient chain-of-thought reasoning.

Regarding reasoning efficiency evaluation, MME-CoT\citep{jiang2025mme} introduces two metrics—relevance and reflection quality—to measure the effectiveness of multimodal chain-of-thought reasoning. To address inefficiencies caused by verbose texts in multimodal reasoning, Shen et al.\citep{shen2025efficient} propose the Heima framework, which aims to enable efficient hidden reasoning. This approach fine-tunes the Heima encoder by encoding each reasoning chain as a single token and incorporates interpretable prompt tokens in the decoder to decode the hidden representations encapsulated in the reasoning tokens. Lu et al.\citep{lu2025prolonged} propose a deterministic adaptive reasoning framework called CAR, which dynamically switches between System 1 and System 2 reasoning based on the model’s perplexity. CAR initially employs System 1 reasoning to generate concise answers and transitions to System 2 for deeper reasoning when the model exhibits high perplexity toward the initial answer.

Wang et al.\citep{wang2025harnessing} and Feng et al.\citep{feng2025efficient} conduct systematic surveys on the reasoning efficiency of large language models and propose a series of efficient reasoning strategies applicable to both post-training and inference stages, some of which have been successfully applied in multimodal large models.

Overall, research on reasoning economy in multimodal chain-of-thought reasoning remains in its early stages, and related evaluation benchmarks and optimization strategies require further systematic and in-depth investigation.

\subsection{Dataset and Benchmark Construction}
Systematic datasets and rigorous evaluation benchmarks are key foundations driving the sustained development of multimodal chain-of-thought research. Due to characteristics such as cross-modal fusion and multi-step reasoning, multimodal chain-of-thought data involve high annotation complexity and substantial construction costs. Currently, there remain deficiencies in modal coverage, reasoning complexity, and dataset scale. Moreover, existing research lacks a unified standardized evaluation framework and sufficiently challenging benchmarks.

Future work should focus on enhancing the construction of high-quality multimodal reasoning datasets and establishing comprehensive, multi-dimensional, and multi-level evaluation systems to more thoroughly assess models’ chain-of-thought reasoning capabilities. Additionally, most current datasets and benchmarks concentrate on domains like mathematics, science, and logical reasoning due to their strong logical structures, which are well suited for chain-of-thought reasoning formats. Therefore, subsequent efforts should expand to more practical application scenarios, building multimodal data resources covering a broader range of fields, thereby promoting the generalization and practical deployment of chain-of-thought reasoning abilities.

\section{Conclusion}
This paper presents a systematic and comprehensive survey of CoT-MLLMs. We begin by introducing the fundamental concepts of M-CoT and the necessity of its study, outlining the limitations faced by current MLLMs in their development and highlighting the unique advantages of the chain-of-thought reasoning paradigm. Subsequently, from three dimensions—CoT paradigm construction, post-training design, and inference methodologies—we explore approaches to endow multimodal large language models with chain-of-thought reasoning capabilities. On the evaluation front, we organize and summarize existing multimodal CoT benchmarks and metric systems. Meanwhile, this paper also reviews research progress of multimodal CoT reasoning in various application domains such as embodied intelligence, healthcare, and machine translation. Finally, we conduct an in-depth analysis of the key challenges currently confronting multimodal CoT reasoning and envision future research directions, aiming to provide a clear development path and theoretical foundation for building more robust, efficient, and fully modal-capable chain-of-thought reasoning systems.

\bibliography{references}
\bibliographystyle{iclr2025_conference}


\end{document}

%% file: math_commands.tex
\usepackage{amsmath,amsfonts,bm}

\def\eqref#1{equation~\ref{#1}}

\def\1{\bm{1}}

\DeclareMathAlphabet{\mathsfit}{\encodingdefault}{\sfdefault}{m}{sl}
\SetMathAlphabet{\mathsfit}{bold}{\encodingdefault}{\sfdefault}{bx}{n}













%% file: CoT-MLLMs_survey.bbl
\begin{thebibliography}{248}
\providecommand{\natexlab}[1]{#1}
\providecommand{\url}[1]{\texttt{#1}}
\expandafter\ifx\csname urlstyle\endcsname\relax
  \providecommand{\doi}[1]{doi: #1}\else
  \providecommand{\doi}{doi: \begingroup \urlstyle{rm}\Url}\fi

\bibitem[Agrahari \& Singh(2025)Agrahari and Singh]{agrahari2025tracing}
Shifali Agrahari and Sanasam~Ranbir Singh.
\newblock Tracing thought: Using chain-of-thought reasoning to identify the llm behind ai-generated text.
\newblock \emph{arXiv preprint arXiv:2504.16913}, 2025.

\bibitem[Alayrac et~al.(2022)Alayrac, Donahue, Luc, Miech, Barr, Hasson, Lenc, Mensch, Millican, Reynolds, et~al.]{alayrac2022flamingo}
Jean-Baptiste Alayrac, Jeff Donahue, Pauline Luc, Antoine Miech, Iain Barr, Yana Hasson, Karel Lenc, Arthur Mensch, Katherine Millican, Malcolm Reynolds, et~al.
\newblock Flamingo: a visual language model for few-shot learning.
\newblock \emph{Advances in neural information processing systems}, 35:\penalty0 23716--23736, 2022.

\bibitem[Amini et~al.(2024)Amini, Vieira, Ash, and Cotterell]{amini2024variational}
Afra Amini, Tim Vieira, Elliott Ash, and Ryan Cotterell.
\newblock Variational best-of-n alignment.
\newblock \emph{arXiv preprint arXiv:2407.06057}, 2024.

\bibitem[Anand et~al.(2024{\natexlab{a}})Anand, Jaiswal, Dharmadhikari, Marathe, Popat, Mital, Prasad, Shah, and Zimmermann]{anand2024improving}
Avinash Anand, Raj Jaiswal, Abhishek Dharmadhikari, Atharva Marathe, Harsh~Parimal Popat, Harshil Mital, Kritarth Prasad, Rajiv~Ratn Shah, and Roger Zimmermann.
\newblock Improving multimodal llms ability in geometry problem solving, reasoning, and multistep scoring.
\newblock \emph{arXiv preprint arXiv:2412.00846}, 2024{\natexlab{a}}.

\bibitem[Anand et~al.(2024{\natexlab{b}})Anand, Kapuriya, Singh, Saraf, Lal, Verma, Gupta, and Shah]{anand2024mm}
Avinash Anand, Janak Kapuriya, Apoorv Singh, Jay Saraf, Naman Lal, Astha Verma, Rushali Gupta, and Rajiv Shah.
\newblock Mm-phyqa: Multimodal physics question-answering with multi-image cot prompting.
\newblock In \emph{Pacific-Asia Conference on Knowledge Discovery and Data Mining}, pp.\  53--64. Springer, 2024{\natexlab{b}}.

\bibitem[Antol et~al.(2015)Antol, Agrawal, Lu, Mitchell, Batra, Zitnick, and Parikh]{antol2015vqa}
Stanislaw Antol, Aishwarya Agrawal, Jiasen Lu, Margaret Mitchell, Dhruv Batra, C~Lawrence Zitnick, and Devi Parikh.
\newblock Vqa: Visual question answering.
\newblock In \emph{Proceedings of the IEEE international conference on computer vision}, pp.\  2425--2433, 2015.

\bibitem[Bai et~al.(2025)Bai, Chen, Liu, Wang, Ge, Song, Dang, Wang, Wang, Tang, et~al.]{bai2025qwen2}
Shuai Bai, Keqin Chen, Xuejing Liu, Jialin Wang, Wenbin Ge, Sibo Song, Kai Dang, Peng Wang, Shijie Wang, Jun Tang, et~al.
\newblock Qwen2. 5-vl technical report.
\newblock \emph{arXiv preprint arXiv:2502.13923}, 2025.

\bibitem[Bai et~al.(2024)Bai, Liang, Wan, Xu, Li, Li, Yang, Li, Wang, Cui, et~al.]{bai2024survey}
Tianyi Bai, Hao Liang, Binwang Wan, Yanran Xu, Xi~Li, Shiyu Li, Ling Yang, Bozhou Li, Yifan Wang, Bin Cui, et~al.
\newblock A survey of multimodal large language model from a data-centric perspective.
\newblock \emph{arXiv preprint arXiv:2405.16640}, 2024.

\bibitem[Bailey et~al.(2023)Bailey, Ong, Russell, and Emmons]{bailey2023image}
Luke Bailey, Euan Ong, Stuart Russell, and Scott Emmons.
\newblock Image hijacks: Adversarial images can control generative models at runtime.
\newblock \emph{arXiv preprint arXiv:2309.00236}, 2023.

\bibitem[Besta et~al.(2024)Besta, Blach, Kubicek, Gerstenberger, Podstawski, Gianinazzi, Gajda, Lehmann, Niewiadomski, Nyczyk, et~al.]{besta2024graph}
Maciej Besta, Nils Blach, Ales Kubicek, Robert Gerstenberger, Michal Podstawski, Lukas Gianinazzi, Joanna Gajda, Tomasz Lehmann, Hubert Niewiadomski, Piotr Nyczyk, et~al.
\newblock Graph of thoughts: Solving elaborate problems with large language models.
\newblock In \emph{Proceedings of the AAAI Conference on Artificial Intelligence}, volume~38, pp.\  17682--17690, 2024.

\bibitem[Bi et~al.(2025)Bi, Liang, Zhou, Liu, Guo, Tang, Song, Huang, Sun, He, et~al.]{bi2025reasoning}
Jing Bi, Susan Liang, Xiaofei Zhou, Pinxin Liu, Junjia Guo, Yunlong Tang, Luchuan Song, Chao Huang, Guangyu Sun, Jinxi He, et~al.
\newblock Why reasoning matters? a survey of advancements in multimodal reasoning (v1).
\newblock \emph{arXiv preprint arXiv:2504.03151}, 2025.

\bibitem[Byun et~al.(2024)Byun, Chun, Kil, and Perrault]{byun2024ares}
Ju-Seung Byun, Jiyun Chun, Jihyung Kil, and Andrew Perrault.
\newblock Ares: Alternating reinforcement learning and supervised fine-tuning for enhanced multi-modal chain-of-thought reasoning through diverse ai feedback.
\newblock In \emph{Proceedings of the 2024 Conference on Empirical Methods in Natural Language Processing}, pp.\  4410--4430, 2024.

\bibitem[Caffagni et~al.(2024)Caffagni, Cocchi, Barsellotti, Moratelli, Sarto, Baraldi, Cornia, and Cucchiara]{caffagni2024revolution}
Davide Caffagni, Federico Cocchi, Luca Barsellotti, Nicholas Moratelli, Sara Sarto, Lorenzo Baraldi, Marcella Cornia, and Rita Cucchiara.
\newblock The revolution of multimodal large language models: a survey.
\newblock \emph{arXiv preprint arXiv:2402.12451}, 2024.

\bibitem[Chen et~al.(2024{\natexlab{a}})Chen, Song, Chen, Yang, Zhao, and Zhang]{chen2024make}
Andong Chen, Yuchen Song, Kehai Chen, Muyun Yang, Tiejun Zhao, and Min Zhang.
\newblock Make imagination clearer! stable diffusion-based visual imagination for multimodal machine translation.
\newblock \emph{arXiv preprint arXiv:2412.12627}, 2024{\natexlab{a}}.

\bibitem[Chen et~al.(2023)Chen, Han, Zhao, Zhang, Shi, Xu, and Xu]{chen2023x}
Feilong Chen, Minglun Han, Haozhi Zhao, Qingyang Zhang, Jing Shi, Shuang Xu, and Bo~Xu.
\newblock X-llm: Bootstrapping advanced large language models by treating multi-modalities as foreign languages.
\newblock \emph{arXiv preprint arXiv:2305.04160}, 2023.

\bibitem[Chen et~al.(2024{\natexlab{b}})Chen, Liu, Li, An, Deng, Feng, Zhao, and Xie]{chen2024plug}
Jiaxing Chen, Yuxuan Liu, Dehu Li, Xiang An, Weimo Deng, Ziyong Feng, Yongle Zhao, and Yin Xie.
\newblock Plug-and-play grounding of reasoning in multimodal large language models.
\newblock \emph{arXiv preprint arXiv:2403.19322}, 2024{\natexlab{b}}.

\bibitem[Chen et~al.(2024{\natexlab{c}})Chen, Qin, Zhang, Chen, Xu, and Che]{chen2024m3cot}
Qiguang Chen, Libo Qin, Jin Zhang, Zhi Chen, Xiao Xu, and Wanxiang Che.
\newblock M3cot: A novel benchmark for multi-domain multi-step multi-modal chain-of-thought.
\newblock In \emph{Proceedings of the 62nd Annual Meeting of the Association for Computational Linguistics (Volume 1: Long Papers)}, pp.\  8199--8221, 2024{\natexlab{c}}.

\bibitem[Chen et~al.(2025{\natexlab{a}})Chen, Qin, Liu, Peng, Guan, Wang, Hu, Zhou, Gao, and Che]{chen2025towards}
Qiguang Chen, Libo Qin, Jinhao Liu, Dengyun Peng, Jiannan Guan, Peng Wang, Mengkang Hu, Yuhang Zhou, Te~Gao, and Wanxiang Che.
\newblock Towards reasoning era: A survey of long chain-of-thought for reasoning large language models.
\newblock \emph{arXiv preprint arXiv:2503.09567}, 2025{\natexlab{a}}.

\bibitem[Chen et~al.(2025{\natexlab{b}})Chen, Zhang, Zhu, Liu, Gao, Xiong, Li, and He]{chen2025bring}
Shiqi Chen, Jinghan Zhang, Tongyao Zhu, Wei Liu, Siyang Gao, Miao Xiong, Manling Li, and Junxian He.
\newblock Bring reason to vision: Understanding perception and reasoning through model merging.
\newblock \emph{arXiv preprint arXiv:2505.05464}, 2025{\natexlab{b}}.

\bibitem[Chen et~al.(2024{\natexlab{d}})Chen, Li, and Niu]{chen2024boosting}
Sijia Chen, Baochun Li, and Di~Niu.
\newblock Boosting of thoughts: Trial-and-error problem solving with large language models.
\newblock \emph{arXiv preprint arXiv:2402.11140}, 2024{\natexlab{d}}.

\bibitem[Chen et~al.(2024{\natexlab{e}})Chen, Sikka, Cogswell, Ji, and Divakaran]{chen2024dress}
Yangyi Chen, Karan Sikka, Michael Cogswell, Heng Ji, and Ajay Divakaran.
\newblock Dress: Instructing large vision-language models to align and interact with humans via natural language feedback.
\newblock In \emph{Proceedings of the IEEE/CVF Conference on Computer Vision and Pattern Recognition}, pp.\  14239--14250, 2024{\natexlab{e}}.

\bibitem[Cheng et~al.(2024)Cheng, Li, Xu, Zhang, Zhou, and Liu]{cheng2024vision}
Kanzhi Cheng, Yantao Li, Fangzhi Xu, Jianbing Zhang, Hao Zhou, and Yang Liu.
\newblock Vision-language models can self-improve reasoning via reflection.
\newblock \emph{arXiv preprint arXiv:2411.00855}, 2024.

\bibitem[Cheng et~al.(2025{\natexlab{a}})Cheng, Li, Zhao, and Wen]{cheng2025think}
Xiaoxue Cheng, Junyi Li, Wayne~Xin Zhao, and Ji-Rong Wen.
\newblock Think more, hallucinate less: Mitigating hallucinations via dual process of fast and slow thinking.
\newblock \emph{arXiv preprint arXiv:2501.01306}, 2025{\natexlab{a}}.

\bibitem[Cheng et~al.(2025{\natexlab{b}})Cheng, Chen, Zhang, Fei, Feng, Che, Li, and Qin]{cheng2025comt}
Zihui Cheng, Qiguang Chen, Jin Zhang, Hao Fei, Xiaocheng Feng, Wanxiang Che, Min Li, and Libo Qin.
\newblock Comt: A novel benchmark for chain of multi-modal thought on large vision-language models.
\newblock In \emph{Proceedings of the AAAI Conference on Artificial Intelligence}, volume~39, pp.\  23678--23686, 2025{\natexlab{b}}.

\bibitem[Chu et~al.(2023)Chu, Chen, Chen, Yu, He, Wang, Peng, Liu, Qin, and Liu]{chu2023navigate}
Zheng Chu, Jingchang Chen, Qianglong Chen, Weijiang Yu, Tao He, Haotian Wang, Weihua Peng, Ming Liu, Bing Qin, and Ting Liu.
\newblock Navigate through enigmatic labyrinth a survey of chain of thought reasoning: Advances, frontiers and future.
\newblock \emph{arXiv preprint arXiv:2309.15402}, 2023.

\bibitem[Chu et~al.(2024)Chu, Chen, Chen, Yu, He, Wang, Peng, Liu, Qin, and Liu]{chu2024navigate}
Zheng Chu, Jingchang Chen, Qianglong Chen, Weijiang Yu, Tao He, Haotian Wang, Weihua Peng, Ming Liu, Bing Qin, and Ting Liu.
\newblock Navigate through enigmatic labyrinth a survey of chain of thought reasoning: Advances, frontiers and future.
\newblock In \emph{Proceedings of the 62nd Annual Meeting of the Association for Computational Linguistics (Volume 1: Long Papers)}, pp.\  1173--1203, 2024.

\bibitem[Cui et~al.(2025{\natexlab{a}})Cui, Hooi, Cai, and Wang]{cui2025process}
Yu~Cui, Bryan Hooi, Yujun Cai, and Yiwei Wang.
\newblock Process or result? manipulated ending tokens can mislead reasoning llms to ignore the correct reasoning steps.
\newblock \emph{arXiv preprint arXiv:2503.19326}, 2025{\natexlab{a}}.

\bibitem[Cui et~al.(2025{\natexlab{b}})Cui, Yuan, Wang, Li, Du, and Ding]{cui2025draw}
Zhiqing Cui, Jiahao Yuan, Hanqing Wang, Yanshu Li, Chenxu Du, and Zhenglong Ding.
\newblock Draw with thought: Unleashing multimodal reasoning for scientific diagram generation.
\newblock \emph{arXiv preprint arXiv:2504.09479}, 2025{\natexlab{b}}.

\bibitem[Dai et~al.(2023)Dai, Li, LI, Tiong, Zhao, Wang, Li, Fung, and Hoi]{dai2023instructblip}
Wenliang Dai, Junnan Li, DONGXU LI, Anthony Tiong, Junqi Zhao, Weisheng Wang, Boyang Li, Pascale~N Fung, and Steven Hoi.
\newblock Instructblip: Towards general-purpose vision-language models with instruction tuning.
\newblock \emph{Advances in Neural Information Processing Systems}, 36:\penalty0 49250--49267, 2023.

\bibitem[Dai(2024)]{dai2024interpretable}
Yi~Dai.
\newblock Interpretable video based stress detection with self-refine chain-of-thought reasoning.
\newblock \emph{arXiv preprint arXiv:2410.09449}, 2024.

\bibitem[Deng et~al.(2024)Deng, Liu, Li, Luo, Wu, Zhang, Lyu, Zhang, Zhang, Ding, et~al.]{deng2024r}
Linger Deng, Yuliang Liu, Bohan Li, Dongliang Luo, Liang Wu, Chengquan Zhang, Pengyuan Lyu, Ziyang Zhang, Gang Zhang, Errui Ding, et~al.
\newblock R-cot: Reverse chain-of-thought problem generation for geometric reasoning in large multimodal models.
\newblock \emph{arXiv preprint arXiv:2410.17885}, 2024.

\bibitem[Dhuliawala et~al.(2023)Dhuliawala, Komeili, Xu, Raileanu, Li, Celikyilmaz, and Weston]{dhuliawala2023chain}
Shehzaad Dhuliawala, Mojtaba Komeili, Jing Xu, Roberta Raileanu, Xian Li, Asli Celikyilmaz, and Jason Weston.
\newblock Chain-of-verification reduces hallucination in large language models.
\newblock \emph{arXiv preprint arXiv:2309.11495}, 2023.

\bibitem[Dong et~al.(2024)Dong, Zhang, Deng, Zhu, Dou, and Wen]{dong2024progressive}
Guanting Dong, Chenghao Zhang, Mengjie Deng, Yutao Zhu, Zhicheng Dou, and Ji-Rong Wen.
\newblock Progressive multimodal reasoning via active retrieval.
\newblock \emph{arXiv preprint arXiv:2412.14835}, 2024.

\bibitem[Dong et~al.(2023)Dong, Han, Peng, Qi, Ge, Yang, Zhao, Sun, Zhou, Wei, et~al.]{dong2023dreamllm}
Runpei Dong, Chunrui Han, Yuang Peng, Zekun Qi, Zheng Ge, Jinrong Yang, Liang Zhao, Jianjian Sun, Hongyu Zhou, Haoran Wei, et~al.
\newblock Dreamllm: Synergistic multimodal comprehension and creation.
\newblock \emph{arXiv preprint arXiv:2309.11499}, 2023.

\bibitem[Dong et~al.(2025)Dong, Liu, Sun, Yang, Hu, Rao, and Liu]{dong2025insight}
Yuhao Dong, Zuyan Liu, Hai-Long Sun, Jingkang Yang, Winston Hu, Yongming Rao, and Ziwei Liu.
\newblock Insight-v: Exploring long-chain visual reasoning with multimodal large language models.
\newblock In \emph{Proceedings of the Computer Vision and Pattern Recognition Conference}, pp.\  9062--9072, 2025.

\bibitem[Du et~al.(2024)Du, Ma, Yang, Deng, Chen, Yang, Xiang, Liu, and Qin]{du2024cot}
Yexing Du, Ziyang Ma, Yifan Yang, Keqi Deng, Xie Chen, Bo~Yang, Yang Xiang, Ming Liu, and Bing Qin.
\newblock Cot-st: Enhancing llm-based speech translation with multimodal chain-of-thought.
\newblock \emph{arXiv preprint arXiv:2409.19510}, 2024.

\bibitem[Du et~al.(2025)Du, Liu, Li, Zhao, Huo, Wang, Chen, Liu, Wang, and Wen]{du2025virgo}
Yifan Du, Zikang Liu, Yifan Li, Wayne~Xin Zhao, Yuqi Huo, Bingning Wang, Weipeng Chen, Zheng Liu, Zhongyuan Wang, and Ji-Rong Wen.
\newblock Virgo: A preliminary exploration on reproducing o1-like mllm.
\newblock \emph{arXiv preprint arXiv:2501.01904}, 2025.

\bibitem[Elhenawy et~al.(2024)Elhenawy, Abutahoun, Alhadidi, Jaber, Ashqar, Jaradat, Abdelhay, Glaser, and Rakotonirainy]{elhenawy2024visual}
Mohammed Elhenawy, Ahmad Abutahoun, Taqwa~I Alhadidi, Ahmed Jaber, Huthaifa~I Ashqar, Shadi Jaradat, Ahmed Abdelhay, Sebastien Glaser, and Andry Rakotonirainy.
\newblock Visual reasoning and multi-agent approach in multimodal large language models (mllms): Solving tsp and mtsp combinatorial challenges.
\newblock \emph{arXiv preprint arXiv:2407.00092}, 2024.

\bibitem[Fan et~al.(2024)Fan, Hua, Li, Zhu, Jin, Li, Ling, Chi, Wang, Ma, et~al.]{fan2024nphardeval4v}
Lizhou Fan, Wenyue Hua, Xiang Li, Kaijie Zhu, Mingyu Jin, Lingyao Li, Haoyang Ling, Jinkui Chi, Jindong Wang, Xin Ma, et~al.
\newblock Nphardeval4v: A dynamic reasoning benchmark of multimodal large language models.
\newblock \emph{arXiv preprint arXiv:2403.01777}, 2024.

\bibitem[Fang et~al.(2025)Fang, Wang, Wang, Yao, Wang, Zhang, Wang, and Chua]{fang2025safemlrm}
Junfeng Fang, Yukai Wang, Ruipeng Wang, Zijun Yao, Kun Wang, An~Zhang, Xiang Wang, and Tat-Seng Chua.
\newblock Safemlrm: Demystifying safety in multi-modal large reasoning models.
\newblock \emph{arXiv preprint arXiv:2504.08813}, 2025.

\bibitem[Fei et~al.(2024)Fei, Wu, Ji, Zhang, Zhang, Lee, and Hsu]{fei2024video}
Hao Fei, Shengqiong Wu, Wei Ji, Hanwang Zhang, Meishan Zhang, Mong~Li Lee, and Wynne Hsu.
\newblock Video-of-thought: step-by-step video reasoning from perception to cognition.
\newblock In \emph{Proceedings of the 41st International Conference on Machine Learning}, pp.\  13109--13125, 2024.

\bibitem[Feng et~al.(2025)Feng, Fang, Ma, and Wang]{feng2025efficient}
Sicheng Feng, Gongfan Fang, Xinyin Ma, and Xinchao Wang.
\newblock Efficient reasoning models: A survey.
\newblock \emph{arXiv preprint arXiv:2504.10903}, 2025.

\bibitem[Gai et~al.(2024)Gai, Zhou, Liu, Feng, Wu, and Liu]{gai2024medthink}
Xiaotang Gai, Chenyi Zhou, Jiaxiang Liu, Yang Feng, Jian Wu, and Zuozhu Liu.
\newblock Medthink: Explaining medical visual question answering via multimodal decision-making rationale.
\newblock \emph{arXiv preprint arXiv:2404.12372}, 2024.

\bibitem[Gao et~al.(2025)Gao, Li, Cao, and Li]{gao2025interleaved}
Jun Gao, Yongqi Li, Ziqiang Cao, and Wenjie Li.
\newblock Interleaved-modal chain-of-thought.
\newblock In \emph{Proceedings of the Computer Vision and Pattern Recognition Conference}, pp.\  19520--19529, 2025.

\bibitem[Gao et~al.(2024)Gao, Chen, Zhang, Fu, Shen, Zhang, Zhang, Zheng, Sun, Cao, et~al.]{gao2024cantor}
Timin Gao, Peixian Chen, Mengdan Zhang, Chaoyou Fu, Yunhang Shen, Yan Zhang, Shengchuan Zhang, Xiawu Zheng, Xing Sun, Liujuan Cao, et~al.
\newblock Cantor: Inspiring multimodal chain-of-thought of mllm.
\newblock In \emph{Proceedings of the 32nd ACM International Conference on Multimedia}, pp.\  9096--9105, 2024.

\bibitem[Ghaffari \& Krishnaswamy(2024)Ghaffari and Krishnaswamy]{ghaffari2024exploring}
Sadaf Ghaffari and Nikhil Krishnaswamy.
\newblock Exploring failure cases in multimodal reasoning about physical dynamics.
\newblock In \emph{Proceedings of the AAAI Symposium Series}, volume~3, pp.\  105--114, 2024.

\bibitem[Ghosal et~al.(2024)Ghosal, Han, Ken, and Poria]{ghosal2024language}
Deepanway Ghosal, Vernon Toh~Yan Han, Chia~Yew Ken, and Soujanya Poria.
\newblock Are language models puzzle prodigies? algorithmic puzzles unveil serious challenges in multimodal reasoning.
\newblock \emph{arXiv preprint arXiv:2403.03864}, 2024.

\bibitem[Gong et~al.(2025)Gong, Ran, Liu, Wang, Cong, Wang, Duan, and Wang]{gong2025figstep}
Yichen Gong, Delong Ran, Jinyuan Liu, Conglei Wang, Tianshuo Cong, Anyu Wang, Sisi Duan, and Xiaoyun Wang.
\newblock Figstep: Jailbreaking large vision-language models via typographic visual prompts.
\newblock In \emph{Proceedings of the AAAI Conference on Artificial Intelligence}, volume~39, pp.\  23951--23959, 2025.

\bibitem[Gou et~al.(2023)Gou, Shao, Gong, Shen, Yang, Duan, and Chen]{gou2023critic}
Zhibin Gou, Zhihong Shao, Yeyun Gong, Yelong Shen, Yujiu Yang, Nan Duan, and Weizhu Chen.
\newblock Critic: Large language models can self-correct with tool-interactive critiquing.
\newblock \emph{arXiv preprint arXiv:2305.11738}, 2023.

\bibitem[Gu et~al.(2024)Gu, Liu, Yin, and Zhang]{gu2024inquire}
Zishan Gu, Fenglin Liu, Changchang Yin, and Ping Zhang.
\newblock Inquire, interact, and integrate: A proactive agent collaborative framework for zero-shot multimodal medical reasoning.
\newblock \emph{arXiv preprint arXiv:2405.11640}, 2024.

\bibitem[Guan et~al.(2024)Guan, Joglekar, Wallace, Jain, Barak, Helyar, Dias, Vallone, Ren, Wei, et~al.]{guan2024deliberative}
Melody~Y Guan, Manas Joglekar, Eric Wallace, Saachi Jain, Boaz Barak, Alec Helyar, Rachel Dias, Andrea Vallone, Hongyu Ren, Jason Wei, et~al.
\newblock Deliberative alignment: Reasoning enables safer language models.
\newblock \emph{arXiv preprint arXiv:2412.16339}, 2024.

\bibitem[Guo et~al.(2025)Guo, Yang, Zhang, Song, Zhang, Xu, Zhu, Ma, Wang, Bi, et~al.]{guo2025deepseek}
Daya Guo, Dejian Yang, Haowei Zhang, Junxiao Song, Ruoyu Zhang, Runxin Xu, Qihao Zhu, Shirong Ma, Peiyi Wang, Xiao Bi, et~al.
\newblock Deepseek-r1: Incentivizing reasoning capability in llms via reinforcement learning.
\newblock \emph{arXiv preprint arXiv:2501.12948}, 2025.

\bibitem[Guo \& Tourani(2025)Guo and Tourani]{guo2025darkmind}
Zhen Guo and Reza Tourani.
\newblock Darkmind: Latent chain-of-thought backdoor in customized llms.
\newblock \emph{arXiv preprint arXiv:2501.18617}, 2025.

\bibitem[Hao et~al.(2025{\natexlab{a}})Hao, Bethala, Pudasaini, Huang, Yuan, Wen, Huang, Nguyen, and Fang]{hao2025embodied}
Yu~Hao, Geeta Chandra~Raju Bethala, Niraj Pudasaini, Hao Huang, Shuaihang Yuan, Congcong Wen, Baoru Huang, Anh Nguyen, and Yi~Fang.
\newblock Embodied chain of action reasoning with multi-modal foundation model for humanoid loco-manipulation.
\newblock \emph{arXiv preprint arXiv:2504.09532}, 2025{\natexlab{a}}.

\bibitem[Hao et~al.(2025{\natexlab{b}})Hao, Gu, Wang, Li, Yang, Wang, and Cheng]{hao2025can}
Yunzhuo Hao, Jiawei Gu, Huichen~Will Wang, Linjie Li, Zhengyuan Yang, Lijuan Wang, and Yu~Cheng.
\newblock Can mllms reason in multimodality? emma: An enhanced multimodal reasoning benchmark.
\newblock \emph{arXiv preprint arXiv:2501.05444}, 2025{\natexlab{b}}.

\bibitem[Himakunthala et~al.(2023)Himakunthala, Ouyang, Rose, He, Mei, Lu, Sonar, Saxon, and Wang]{himakunthala2023let}
Vaishnavi Himakunthala, Andy Ouyang, Daniel Rose, Ryan He, Alex Mei, Yujie Lu, Chinmay Sonar, Michael Saxon, and William~Yang Wang.
\newblock Let's think frame by frame with vip: A video infilling and prediction dataset for evaluating video chain-of-thought.
\newblock \emph{arXiv preprint arXiv:2305.13903}, 2023.

\bibitem[Hu et~al.(2024{\natexlab{a}})Hu, Liu, Yin, Zhou, and Li]{hu2024multimodal}
Chuanbo Hu, Bin Liu, Minglei Yin, Yilu Zhou, and Xin Li.
\newblock Multimodal chain-of-thought reasoning via chatgpt to protect children from age-inappropriate apps.
\newblock \emph{arXiv preprint arXiv:2407.06309}, 2024{\natexlab{a}}.

\bibitem[Hu et~al.(2024{\natexlab{b}})Hu, Shi, Fu, Roth, Ostendorf, Zettlemoyer, Smith, and Krishna]{hu2024visual}
Yushi Hu, Weijia Shi, Xingyu Fu, Dan Roth, Mari Ostendorf, Luke Zettlemoyer, Noah~A Smith, and Ranjay Krishna.
\newblock Visual sketchpad: Sketching as a visual chain of thought for multimodal language models.
\newblock \emph{arXiv preprint arXiv:2406.09403}, 2024{\natexlab{b}}.

\bibitem[Huang \& Chang(2022)Huang and Chang]{huang2022towards}
Jie Huang and Kevin Chen-Chuan Chang.
\newblock Towards reasoning in large language models: A survey.
\newblock \emph{arXiv preprint arXiv:2212.10403}, 2022.

\bibitem[Huang et~al.(2025)Huang, Jia, Zhai, Cao, Ye, Zhao, Xu, Hu, and Lin]{huang2025vision}
Wenxuan Huang, Bohan Jia, Zijie Zhai, Shaosheng Cao, Zheyu Ye, Fei Zhao, Zhe Xu, Yao Hu, and Shaohui Lin.
\newblock Vision-r1: Incentivizing reasoning capability in multimodal large language models.
\newblock \emph{arXiv preprint arXiv:2503.06749}, 2025.

\bibitem[Imam et~al.(2025)Imam, Lyu, and Aji]{imam2025can}
Mohamed~Fazli Imam, Chenyang Lyu, and Alham~Fikri Aji.
\newblock Can multimodal llms do visual temporal understanding and reasoning? the answer is no!
\newblock \emph{arXiv preprint arXiv:2501.10674}, 2025.

\bibitem[Ishaq et~al.(2025)Ishaq, Lahoud, More, Thawakar, Thawkar, Dissanayake, Ahsan, Li, Khan, Cholakkal, et~al.]{ishaq2025drivelmm}
Ayesha Ishaq, Jean Lahoud, Ketan More, Omkar Thawakar, Ritesh Thawkar, Dinura Dissanayake, Noor Ahsan, Yuhao Li, Fahad~Shahbaz Khan, Hisham Cholakkal, et~al.
\newblock Drivelmm-o1: A step-by-step reasoning dataset and large multimodal model for driving scenario understanding.
\newblock \emph{arXiv preprint arXiv:2503.10621}, 2025.

\bibitem[Islam(2024)]{islam2024application}
Md~Robiul Islam.
\newblock Application of multimodal large language models in autonomous driving.
\newblock \emph{arXiv preprint arXiv:2412.16410}, 2024.

\bibitem[Jia et~al.(2025{\natexlab{a}})Jia, Zhang, Zhang, and Wan]{jia2025exploring}
Boyu Jia, Junzhe Zhang, Huixuan Zhang, and Xiaojun Wan.
\newblock Exploring and evaluating multimodal knowledge reasoning consistency of multimodal large language models.
\newblock \emph{arXiv preprint arXiv:2503.04801}, 2025{\natexlab{a}}.

\bibitem[Jia et~al.(2025{\natexlab{b}})Jia, Liu, Li, Liu, and Gao]{jia2025dcot}
Zixi Jia, Jiqiang Liu, Hexiao Li, Qinghua Liu, and Hongbin Gao.
\newblock Dcot: Dual chain-of-thought prompting for large multimodal models.
\newblock In \emph{Asian Conference on Machine Learning}, pp.\  1064--1079. PMLR, 2025{\natexlab{b}}.

\bibitem[Jiang et~al.(2025{\natexlab{a}})Jiang, Zhang, Guo, Li, Qi, Chen, Wang, Jin, Guo, Yan, et~al.]{jiang2025mme}
Dongzhi Jiang, Renrui Zhang, Ziyu Guo, Yanwei Li, Yu~Qi, Xinyan Chen, Liuhui Wang, Jianhan Jin, Claire Guo, Shen Yan, et~al.
\newblock Mme-cot: Benchmarking chain-of-thought in large multimodal models for reasoning quality, robustness, and efficiency.
\newblock \emph{arXiv preprint arXiv:2502.09621}, 2025{\natexlab{a}}.

\bibitem[Jiang et~al.(2025{\natexlab{b}})Jiang, Xu, Li, Niu, Xiang, Li, Lin, and Poovendran]{jiang2025safechain}
Fengqing Jiang, Zhangchen Xu, Yuetai Li, Luyao Niu, Zhen Xiang, Bo~Li, Bill~Yuchen Lin, and Radha Poovendran.
\newblock Safechain: Safety of language models with long chain-of-thought reasoning capabilities.
\newblock \emph{arXiv preprint arXiv:2502.12025}, 2025{\natexlab{b}}.

\bibitem[Jiang et~al.(2024)Jiang, Luo, Yang, Xiong, Chen, Zeng, Ren, and Zhang]{jiang2024chatrex}
Qing Jiang, Gen Luo, Yuqin Yang, Yuda Xiong, Yihao Chen, Zhaoyang Zeng, Tianhe Ren, and Lei Zhang.
\newblock Chatrex: Taming multimodal llm for joint perception and understanding.
\newblock \emph{arXiv preprint arXiv:2411.18363}, 2024.

\bibitem[Khalid et~al.(2024)Khalid, Iqbal, Farooq, Rahnavard, Hua, and Chen]{khalid2024evlm}
Umar Khalid, Hasan Iqbal, Azib Farooq, Nazanin Rahnavard, Jing Hua, and Chen Chen.
\newblock Evlm: Self-reflective multimodal reasoning for cross-dimensional visual editing.
\newblock \emph{arXiv preprint arXiv:2412.10566}, 2024.

\bibitem[Khaliq et~al.(2024)Khaliq, Chang, Ma, Pflugfelder, and Mileti{\'c}]{khaliq2024ragar}
M~Abdul Khaliq, Paul Chang, Mingyang Ma, Bernhard Pflugfelder, and Filip Mileti{\'c}.
\newblock Ragar, your falsehood radar: Rag-augmented reasoning for political fact-checking using multimodal large language models.
\newblock \emph{arXiv preprint arXiv:2404.12065}, 2024.

\bibitem[Kim et~al.(2021)Kim, Son, and Kim]{kim2021vilt}
Wonjae Kim, Bokyung Son, and Ildoo Kim.
\newblock Vilt: Vision-and-language transformer without convolution or region supervision.
\newblock In \emph{International conference on machine learning}, pp.\  5583--5594. PMLR, 2021.

\bibitem[Kojima et~al.(2022)Kojima, Gu, Reid, Matsuo, and Iwasawa]{kojima2022large}
Takeshi Kojima, Shixiang~Shane Gu, Machel Reid, Yutaka Matsuo, and Yusuke Iwasawa.
\newblock Large language models are zero-shot reasoners.
\newblock \emph{Advances in neural information processing systems}, 35:\penalty0 22199--22213, 2022.

\bibitem[Kumar et~al.(2025)Kumar, Roh, Naseh, Karpinska, Iyyer, Houmansadr, and Bagdasarian]{kumar2025overthink}
Abhinav Kumar, Jaechul Roh, Ali Naseh, Marzena Karpinska, Mohit Iyyer, Amir Houmansadr, and Eugene Bagdasarian.
\newblock Overthink: Slowdown attacks on reasoning llms.
\newblock \emph{arXiv preprint arXiv:2502.02542}, 2025.

\bibitem[Kumar et~al.(2024)Kumar, Cummings, and Stimpson]{kumar2024strengthening}
Surender~Suresh Kumar, ML~Cummings, and Alexander Stimpson.
\newblock Strengthening llm trust boundaries: a survey of prompt injection attacks.
\newblock In \emph{2024 IEEE 4th International Conference on Human-Machine Systems (ICHMS)}, pp.\  1--6, 2024.

\bibitem[Le et~al.(2024)Le, Truong-Vinh, Phan, Le, Nguyen, and Bui]{le2024visualcoder}
Cuong~Chi Le, Hoang-Chau Truong-Vinh, Huy~Nhat Phan, Dung~Duy Le, Tien~N Nguyen, and Nghi~DQ Bui.
\newblock Visualcoder: Guiding large language models in code execution with fine-grained multimodal chain-of-thought reasoning.
\newblock \emph{arXiv preprint arXiv:2410.23402}, 2024.

\bibitem[Lee et~al.(2024)Lee, Wang, Li, and Zhang]{lee2024multimodal}
Junlin Lee, Yequan Wang, Jing Li, and Min Zhang.
\newblock Multimodal reasoning with multimodal knowledge graph.
\newblock \emph{arXiv preprint arXiv:2406.02030}, 2024.

\bibitem[Lei et~al.(2023)Lei, Liao, Ding, et~al.]{lei2023boosting}
Bin Lei, Chunhua Liao, Caiwen Ding, et~al.
\newblock Boosting logical reasoning in large language models through a new framework: The graph of thought.
\newblock \emph{arXiv preprint arXiv:2308.08614}, 2023.

\bibitem[Li et~al.(2025{\natexlab{a}})Li, Wu, Zhang, Xia, Mao, Dong, Vuli{\'c}, and Wei]{li2025imagine}
Chengzu Li, Wenshan Wu, Huanyu Zhang, Yan Xia, Shaoguang Mao, Li~Dong, Ivan Vuli{\'c}, and Furu Wei.
\newblock Imagine while reasoning in space: Multimodal visualization-of-thought.
\newblock \emph{arXiv preprint arXiv:2501.07542}, 2025{\natexlab{a}}.

\bibitem[Li et~al.(2021)Li, Selvaraju, Gotmare, Joty, Xiong, and Hoi]{li2021align}
Junnan Li, Ramprasaath Selvaraju, Akhilesh Gotmare, Shafiq Joty, Caiming Xiong, and Steven Chu~Hong Hoi.
\newblock Align before fuse: Vision and language representation learning with momentum distillation.
\newblock \emph{Advances in neural information processing systems}, 34:\penalty0 9694--9705, 2021.

\bibitem[Li et~al.(2022)Li, Li, Xiong, and Hoi]{li2022blip}
Junnan Li, Dongxu Li, Caiming Xiong, and Steven Hoi.
\newblock Blip: Bootstrapping language-image pre-training for unified vision-language understanding and generation.
\newblock In \emph{International conference on machine learning}, pp.\  12888--12900. PMLR, 2022.

\bibitem[Li et~al.(2023{\natexlab{a}})Li, Li, Savarese, and Hoi]{li2023blip}
Junnan Li, Dongxu Li, Silvio Savarese, and Steven Hoi.
\newblock Blip-2: Bootstrapping language-image pre-training with frozen image encoders and large language models.
\newblock In \emph{International conference on machine learning}, pp.\  19730--19742. PMLR, 2023{\natexlab{a}}.

\bibitem[Li et~al.(2024{\natexlab{a}})Li, Han, Steneker, Primack, Goodside, Zhang, Wang, Menghini, and Yue]{li2024llm}
Nathaniel Li, Ziwen Han, Ian Steneker, Willow Primack, Riley Goodside, Hugh Zhang, Zifan Wang, Cristina Menghini, and Summer Yue.
\newblock Llm defenses are not robust to multi-turn human jailbreaks yet.
\newblock \emph{arXiv preprint arXiv:2408.15221}, 2024{\natexlab{a}}.

\bibitem[Li et~al.(2025{\natexlab{b}})Li, Lan, Chen, Lu, and Jiang]{li2025multimodal}
Yan Li, Xiangyuan Lan, Haifeng Chen, Ke~Lu, and Dongmei Jiang.
\newblock Multimodal pear chain-of-thought reasoning for multimodal sentiment analysis.
\newblock \emph{ACM Transactions on Multimedia Computing, Communications and Applications}, 20\penalty0 (9):\penalty0 1--23, 2025{\natexlab{b}}.

\bibitem[Li et~al.(2023{\natexlab{b}})Li, Lin, Zhang, Fu, Chen, Lou, and Chen]{li2023making}
Yifei Li, Zeqi Lin, Shizhuo Zhang, Qiang Fu, Bei Chen, Jian-Guang Lou, and Weizhu Chen.
\newblock Making language models better reasoners with step-aware verifier.
\newblock In \emph{Proceedings of the 61st Annual Meeting of the Association for Computational Linguistics (Volume 1: Long Papers)}, pp.\  5315--5333, 2023{\natexlab{b}}.

\bibitem[Li et~al.(2025{\natexlab{c}})Li, Liu, Li, Zhang, Xu, Chen, Shi, Jiang, Wang, Wang, et~al.]{li2025perception}
Yunxin Li, Zhenyu Liu, Zitao Li, Xuanyu Zhang, Zhenran Xu, Xinyu Chen, Haoyuan Shi, Shenyuan Jiang, Xintong Wang, Jifang Wang, et~al.
\newblock Perception, reason, think, and plan: A survey on large multimodal reasoning models.
\newblock \emph{arXiv preprint arXiv:2505.04921}, 2025{\natexlab{c}}.

\bibitem[Li et~al.(2024{\natexlab{b}})Li, Wang, Liu, Zhang, Ma, Long, and Cai]{li2024multimodal}
Zhiyuan Li, Heng Wang, Dongnan Liu, Chaoyi Zhang, Ao~Ma, Jieting Long, and Weidong Cai.
\newblock Multimodal causal reasoning benchmark: Challenging vision large language models to discern causal links across modalities.
\newblock \emph{arXiv preprint arXiv:2408.08105}, 2024{\natexlab{b}}.

\bibitem[Li et~al.(2024{\natexlab{c}})Li, Xu, Wang, Wu, and Li]{li2024streetviewllm}
Zongrong Li, Junhao Xu, Siqin Wang, Yifan Wu, and Haiyang Li.
\newblock Streetviewllm: Extracting geographic information using a chain-of-thought multimodal large language model.
\newblock \emph{arXiv preprint arXiv:2411.14476}, 2024{\natexlab{c}}.

\bibitem[Liang et~al.(2024)Liang, Lin, Ruan, Liu, Zhuang, and Liang]{liangmemory}
Xiwen Liang, Min Lin, Weiqi Ruan, Yuecheng Liu, Yuzheng Zhuang, and Xiaodan Liang.
\newblock Memory-driven multimodal chain of thought for embodied long-horizon task planning.
\newblock 2024.

\bibitem[Liao et~al.(2025)Liao, Elflein, He, Leal-Taix{\'e}, Choi, Fidler, and Acuna]{liao2025longperceptualthoughts}
Yuan-Hong Liao, Sven Elflein, Liu He, Laura Leal-Taix{\'e}, Yejin Choi, Sanja Fidler, and David Acuna.
\newblock Longperceptualthoughts: Distilling system-2 reasoning for system-1 perception.
\newblock \emph{arXiv preprint arXiv:2504.15362}, 2025.

\bibitem[Lin et~al.(2024)Lin, Ye, Zhu, Cui, Ning, Jin, and Yuan]{lin2024video}
Bin Lin, Yang Ye, Bin Zhu, Jiaxi Cui, Munan Ning, Peng Jin, and Li~Yuan.
\newblock Video-llava: Learning united visual representation by alignment before projection.
\newblock In \emph{Proceedings of the 2024 Conference on Empirical Methods in Natural Language Processing}, pp.\  5971--5984, 2024.

\bibitem[Lin et~al.(2025{\natexlab{a}})Lin, Nie, Wei, Chen, Ma, Han, Xu, Chang, and Liang]{lin2025navcot}
Bingqian Lin, Yunshuang Nie, Ziming Wei, Jiaqi Chen, Shikui Ma, Jianhua Han, Hang Xu, Xiaojun Chang, and Xiaodan Liang.
\newblock Navcot: Boosting llm-based vision-and-language navigation via learning disentangled reasoning.
\newblock \emph{IEEE Transactions on Pattern Analysis and Machine Intelligence}, 2025{\natexlab{a}}.

\bibitem[Lin et~al.(2014)Lin, Maire, Belongie, Hays, Perona, Ramanan, Doll{\'a}r, and Zitnick]{lin2014microsoft}
Tsung-Yi Lin, Michael Maire, Serge Belongie, James Hays, Pietro Perona, Deva Ramanan, Piotr Doll{\'a}r, and C~Lawrence Zitnick.
\newblock Microsoft coco: Common objects in context.
\newblock In \emph{Computer vision--ECCV 2014: 13th European conference, zurich, Switzerland, September 6-12, 2014, proceedings, part v 13}, pp.\  740--755. Springer, 2014.

\bibitem[Lin et~al.(2025{\natexlab{b}})Lin, Jia, Hu, Pan, Yue, Zhao, Chen, Wu, and Zhang]{lin2025reasoning}
Wang Lin, Liyu Jia, Wentao Hu, Kaihang Pan, Zhongqi Yue, Wei Zhao, Jingyuan Chen, Fei Wu, and Hanwang Zhang.
\newblock Reasoning physical video generation with diffusion timestep tokens via reinforcement learning.
\newblock \emph{arXiv preprint arXiv:2504.15932}, 2025{\natexlab{b}}.

\bibitem[Lin et~al.(2025{\natexlab{c}})Lin, Wang, Chen, Liu, Liu, Su, and Xiao]{lin2025investigating}
Yujie Lin, Ante Wang, Moye Chen, Jingyao Liu, Hao Liu, Jinsong Su, and Xinyan Xiao.
\newblock Investigating inference-time scaling for chain of multi-modal thought: A preliminary study.
\newblock \emph{arXiv preprint arXiv:2502.11514}, 2025{\natexlab{c}}.

\bibitem[Lin et~al.(2025{\natexlab{d}})Lin, Gao, Zhao, Yang, and Sang]{lin2025mind}
Zhiyu Lin, Yifei Gao, Xian Zhao, Yunfan Yang, and Jitao Sang.
\newblock Mind with eyes: from language reasoning to multimodal reasoning.
\newblock \emph{arXiv preprint arXiv:2503.18071}, 2025{\natexlab{d}}.

\bibitem[Liu et~al.(2023{\natexlab{a}})Liu, Lyu, Min, Wang, Su, and Wang]{liu2023retrieval}
Bingshuai Liu, Chenyang Lyu, Zijun Min, Zhanyu Wang, Jinsong Su, and Longyue Wang.
\newblock Retrieval-augmented multi-modal chain-of-thoughts reasoning for large language models.
\newblock \emph{arXiv preprint arXiv:2312.01714}, 2023{\natexlab{a}}.

\bibitem[Liu et~al.(2023{\natexlab{b}})Liu, Li, Wu, and Lee]{liu2023visual}
Haotian Liu, Chunyuan Li, Qingyang Wu, and Yong~Jae Lee.
\newblock Visual instruction tuning.
\newblock \emph{Advances in neural information processing systems}, 36:\penalty0 34892--34916, 2023{\natexlab{b}}.

\bibitem[Liu et~al.(2024{\natexlab{a}})Liu, Wang, Du, Zhou, and Liu]{liu2024medcot}
Jiaxiang Liu, Yuan Wang, Jiawei Du, Joey~Tianyi Zhou, and Zuozhu Liu.
\newblock Medcot: Medical chain of thought via hierarchical expert.
\newblock \emph{arXiv preprint arXiv:2412.13736}, 2024{\natexlab{a}}.

\bibitem[Liu et~al.(2024{\natexlab{b}})Liu, Pan, Zhang, Liu, Wu, Zhou, Zhou, Chen, Jiang, and He]{liu2024cmm}
Wentao Liu, Qianjun Pan, Yi~Zhang, Zhuo Liu, Ji~Wu, Jie Zhou, Aimin Zhou, Qin Chen, Bo~Jiang, and Liang He.
\newblock Cmm-math: A chinese multimodal math dataset to evaluate and enhance the mathematics reasoning of large multimodal models.
\newblock \emph{arXiv preprint arXiv:2409.02834}, 2024{\natexlab{b}}.

\bibitem[Liu et~al.(2023{\natexlab{c}})Liu, Deng, Li, Wang, Wang, Wang, Zhang, Liu, Wang, Zheng, et~al.]{liu2023prompt}
Yi~Liu, Gelei Deng, Yuekang Li, Kailong Wang, Zihao Wang, Xiaofeng Wang, Tianwei Zhang, Yepang Liu, Haoyu Wang, Yan Zheng, et~al.
\newblock Prompt injection attack against llm-integrated applications.
\newblock \emph{arXiv preprint arXiv:2306.05499}, 2023{\natexlab{c}}.

\bibitem[Liu et~al.(2023{\natexlab{d}})Liu, Li, Zhang, Huang, Zha, and Huang]{liu2023matcr}
Yiting Liu, Liang Li, Beichen Zhang, Shan Huang, Zheng-Jun Zha, and Qingming Huang.
\newblock Matcr: Modality-aligned thought chain reasoning for multimodal task-oriented dialogue generation.
\newblock In \emph{Proceedings of the 31st ACM International Conference on Multimedia}, pp.\  5776--5785, 2023{\natexlab{d}}.

\bibitem[Liu et~al.(2025{\natexlab{a}})Liu, Gao, Zhai, Xia, Wu, Xue, Chen, Kawaguchi, Zhang, and Hooi]{liu2025guardreasoner}
Yue Liu, Hongcheng Gao, Shengfang Zhai, Jun Xia, Tianyi Wu, Zhiwei Xue, Yulin Chen, Kenji Kawaguchi, Jiaheng Zhang, and Bryan Hooi.
\newblock Guardreasoner: Towards reasoning-based llm safeguards.
\newblock \emph{arXiv preprint arXiv:2501.18492}, 2025{\natexlab{a}}.

\bibitem[Liu et~al.(2025{\natexlab{b}})Liu, Chi, Wu, Zhang, Hu, Zhang, Zhang, Wu, Cao, Huang, et~al.]{liu2025spatialcot}
Yuecheng Liu, Dafeng Chi, Shiguang Wu, Zhanguang Zhang, Yaochen Hu, Lingfeng Zhang, Yingxue Zhang, Shuang Wu, Tongtong Cao, Guowei Huang, et~al.
\newblock Spatialcot: Advancing spatial reasoning through coordinate alignment and chain-of-thought for embodied task planning.
\newblock \emph{arXiv preprint arXiv:2501.10074}, 2025{\natexlab{b}}.

\bibitem[Liu et~al.(2025{\natexlab{c}})Liu, Li, Wei, Xie, Hu, Xu, Zhang, Han, Yang, and Wu]{liu2025infiguiagent}
Yuhang Liu, Pengxiang Li, Zishu Wei, Congkai Xie, Xueyu Hu, Xinchen Xu, Shengyu Zhang, Xiaotian Han, Hongxia Yang, and Fei Wu.
\newblock Infiguiagent: A multimodal generalist gui agent with native reasoning and reflection.
\newblock \emph{arXiv preprint arXiv:2501.04575}, 2025{\natexlab{c}}.

\bibitem[Liu et~al.(2025{\natexlab{d}})Liu, Peng, Zhong, Yue, Lu, Yu, and Jia]{liu2025seg}
Yuqi Liu, Bohao Peng, Zhisheng Zhong, Zihao Yue, Fanbin Lu, Bei Yu, and Jiaya Jia.
\newblock Seg-zero: Reasoning-chain guided segmentation via cognitive reinforcement.
\newblock \emph{arXiv preprint arXiv:2503.06520}, 2025{\natexlab{d}}.

\bibitem[Liu et~al.(2025{\natexlab{e}})Liu, Sun, Zang, Dong, Cao, Duan, Lin, and Wang]{liu2025visual}
Ziyu Liu, Zeyi Sun, Yuhang Zang, Xiaoyi Dong, Yuhang Cao, Haodong Duan, Dahua Lin, and Jiaqi Wang.
\newblock Visual-rft: Visual reinforcement fine-tuning.
\newblock \emph{arXiv preprint arXiv:2503.01785}, 2025{\natexlab{e}}.

\bibitem[Liu et~al.(2024{\natexlab{c}})Liu, Dong, Rao, Zhou, and Lu]{liu2024chain}
Zuyan Liu, Yuhao Dong, Yongming Rao, Jie Zhou, and Jiwen Lu.
\newblock Chain-of-spot: Interactive reasoning improves large vision-language models.
\newblock \emph{arXiv preprint arXiv:2403.12966}, 2024{\natexlab{c}}.

\bibitem[Long(2023)]{long2023large}
Jieyi Long.
\newblock Large language model guided tree-of-thought.
\newblock \emph{arXiv preprint arXiv:2305.08291}, 2023.

\bibitem[Lou et~al.(2025)Lou, Li, Xu, Shi, Chen, and Huang]{lou2025think}
Xinyue Lou, You Li, Jinan Xu, Xiangyu Shi, Chi Chen, and Kaiyu Huang.
\newblock Think in safety: Unveiling and mitigating safety alignment collapse in multimodal large reasoning model.
\newblock \emph{arXiv preprint arXiv:2505.06538}, 2025.

\bibitem[Lu et~al.(2024)Lu, Liu, Zhang, Wang, Dong, Liu, Sun, Ren, Li, Yang, et~al.]{lu2024deepseek}
Haoyu Lu, Wen Liu, Bo~Zhang, Bingxuan Wang, Kai Dong, Bo~Liu, Jingxiang Sun, Tongzheng Ren, Zhuoshu Li, Hao Yang, et~al.
\newblock Deepseek-vl: towards real-world vision-language understanding.
\newblock \emph{arXiv preprint arXiv:2403.05525}, 2024.

\bibitem[Lu et~al.(2025)Lu, Yu, Xu, Ran, Tang, Wang, Shan, Fu, Feng, Tang, et~al.]{lu2025prolonged}
Jinghui Lu, Haiyang Yu, Siliang Xu, Shiwei Ran, Guozhi Tang, Siqi Wang, Bin Shan, Teng Fu, Hao Feng, Jingqun Tang, et~al.
\newblock Prolonged reasoning is not all you need: Certainty-based adaptive routing for efficient llm/mllm reasoning.
\newblock \emph{arXiv preprint arXiv:2505.15154}, 2025.

\bibitem[Lu et~al.(2022)Lu, Mishra, Xia, Qiu, Chang, Zhu, Tafjord, Clark, and Kalyan]{lu2022learn}
Pan Lu, Swaroop Mishra, Tanglin Xia, Liang Qiu, Kai-Wei Chang, Song-Chun Zhu, Oyvind Tafjord, Peter Clark, and Ashwin Kalyan.
\newblock Learn to explain: Multimodal reasoning via thought chains for science question answering.
\newblock \emph{Advances in Neural Information Processing Systems}, 35:\penalty0 2507--2521, 2022.

\bibitem[Lu et~al.(2023)Lu, Bansal, Xia, Liu, Li, Hajishirzi, Cheng, Chang, Galley, and Gao]{lu2023mathvista}
Pan Lu, Hritik Bansal, Tony Xia, Jiacheng Liu, Chunyuan Li, Hannaneh Hajishirzi, Hao Cheng, Kai-Wei Chang, Michel Galley, and Jianfeng Gao.
\newblock Mathvista: Evaluating mathematical reasoning of foundation models in visual contexts.
\newblock \emph{arXiv preprint arXiv:2310.02255}, 2023.

\bibitem[Luo et~al.(2025)Luo, Zheng, Wang, Ni, Lin, Jiang, Yu, Shi, Chu, Zeng, et~al.]{luo2025ursa}
Ruilin Luo, Zhuofan Zheng, Yifan Wang, Xinzhe Ni, Zicheng Lin, Songtao Jiang, Yiyao Yu, Chufan Shi, Ruihang Chu, Jin Zeng, et~al.
\newblock Ursa: Understanding and verifying chain-of-thought reasoning in multimodal mathematics.
\newblock \emph{arXiv preprint arXiv:2501.04686}, 2025.

\bibitem[Luo et~al.(2024)Luo, Ding, Song, Zhang, and Loo]{luo2024pkrd}
Xuewen Luo, Fan Ding, Yinsheng Song, Xiaofeng Zhang, and Junnyong Loo.
\newblock Pkrd-cot: A unified chain-of-thought prompting for multi-modal large language models in autonomous driving.
\newblock \emph{arXiv preprint arXiv:2412.02025}, 2024.

\bibitem[Ma et~al.(2024{\natexlab{a}})Ma, Luo, Wang, and Liu]{ma2024visual}
Siyuan Ma, Weidi Luo, Yu~Wang, and Xiaogeng Liu.
\newblock Visual-roleplay: Universal jailbreak attack on multimodal large language models via role-playing image character.
\newblock \emph{arXiv preprint arXiv:2405.20773}, 2024{\natexlab{a}}.

\bibitem[Ma et~al.(2024{\natexlab{b}})Ma, Cao, Sun, Pavone, and Xiao]{ma2024dolphins}
Yingzi Ma, Yulong Cao, Jiachen Sun, Marco Pavone, and Chaowei Xiao.
\newblock Dolphins: Multimodal language model for driving.
\newblock In \emph{European Conference on Computer Vision}, pp.\  403--420. Springer, 2024{\natexlab{b}}.

\bibitem[Madaan et~al.(2023)Madaan, Tandon, Gupta, Hallinan, Gao, Wiegreffe, Alon, Dziri, Prabhumoye, Yang, et~al.]{madaan2023self}
Aman Madaan, Niket Tandon, Prakhar Gupta, Skyler Hallinan, Luyu Gao, Sarah Wiegreffe, Uri Alon, Nouha Dziri, Shrimai Prabhumoye, Yiming Yang, et~al.
\newblock Self-refine: Iterative refinement with self-feedback.
\newblock \emph{Advances in Neural Information Processing Systems}, 36:\penalty0 46534--46594, 2023.

\bibitem[Ma{\l}ki{\'n}ski et~al.(2024)Ma{\l}ki{\'n}ski, Pawlonka, and Ma{\'n}dziuk]{malkinski2024reasoning}
Miko{\l}aj Ma{\l}ki{\'n}ski, Szymon Pawlonka, and Jacek Ma{\'n}dziuk.
\newblock Reasoning limitations of multimodal large language models. a case study of bongard problems.
\newblock \emph{arXiv preprint arXiv:2411.01173}, 2024.

\bibitem[Mao et~al.(2023)Mao, Ye, Qian, Pavone, and Wang]{mao2023language}
Jiageng Mao, Junjie Ye, Yuxi Qian, Marco Pavone, and Yue Wang.
\newblock A language agent for autonomous driving.
\newblock \emph{arXiv preprint arXiv:2311.10813}, 2023.

\bibitem[Mathur et~al.(2025)Mathur, Qian, Liang, and Morency]{mathur2025social}
Leena Mathur, Marian Qian, Paul~Pu Liang, and Louis-Philippe Morency.
\newblock Social genome: Grounded social reasoning abilities of multimodal models.
\newblock \emph{arXiv preprint arXiv:2502.15109}, 2025.

\bibitem[Mitra et~al.(2024)Mitra, Huang, Darrell, and Herzig]{mitra2024compositional}
Chancharik Mitra, Brandon Huang, Trevor Darrell, and Roei Herzig.
\newblock Compositional chain-of-thought prompting for large multimodal models.
\newblock In \emph{Proceedings of the IEEE/CVF Conference on Computer Vision and Pattern Recognition}, pp.\  14420--14431, 2024.

\bibitem[Mondal et~al.(2024)Mondal, Modi, Panda, Singh, and Rao]{mondal2024kam}
Debjyoti Mondal, Suraj Modi, Subhadarshi Panda, Rituraj Singh, and Godawari~Sudhakar Rao.
\newblock Kam-cot: Knowledge augmented multimodal chain-of-thoughts reasoning.
\newblock In \emph{Proceedings of the AAAI conference on artificial intelligence}, volume~38, pp.\  18798--18806, 2024.

\bibitem[Mu et~al.(2023)Mu, Zhang, Hu, Wang, Ding, Jin, Wang, Dai, Qiao, and Luo]{mu2023embodiedgpt}
Yao Mu, Qinglong Zhang, Mengkang Hu, Wenhai Wang, Mingyu Ding, Jun Jin, Bin Wang, Jifeng Dai, Yu~Qiao, and Ping Luo.
\newblock Embodiedgpt: Vision-language pre-training via embodied chain of thought.
\newblock \emph{Advances in Neural Information Processing Systems}, 36:\penalty0 25081--25094, 2023.

\bibitem[Ni et~al.(2024)Ni, Hao, Wu, Kou, Liu, Zheng, Wang, and Zhuang]{ni2024generate}
Fei Ni, Jianye Hao, Shiguang Wu, Longxin Kou, Jiashun Liu, Yan Zheng, Bin Wang, and Yuzheng Zhuang.
\newblock Generate subgoal images before act: Unlocking the chain-of-thought reasoning in diffusion model for robot manipulation with multimodal prompts.
\newblock In \emph{Proceedings of the IEEE/CVF Conference on Computer Vision and Pattern Recognition}, pp.\  13991--14000, 2024.

\bibitem[Nie et~al.(2024)Nie, Peng, Wang, Cai, Han, Xu, and Zhang]{nie2024reason2drive}
Ming Nie, Renyuan Peng, Chunwei Wang, Xinyue Cai, Jianhua Han, Hang Xu, and Li~Zhang.
\newblock Reason2drive: Towards interpretable and chain-based reasoning for autonomous driving.
\newblock In \emph{European Conference on Computer Vision}, pp.\  292--308. Springer, 2024.

\bibitem[Pan et~al.(2025)Pan, Liu, Wu, Liu, Zhu, Li, Chen, Ouyang, and Rueckert]{pan2025medvlm}
Jiazhen Pan, Che Liu, Junde Wu, Fenglin Liu, Jiayuan Zhu, Hongwei~Bran Li, Chen Chen, Cheng Ouyang, and Daniel Rueckert.
\newblock Medvlm-r1: Incentivizing medical reasoning capability of vision-language models (vlms) via reinforcement learning.
\newblock \emph{arXiv preprint arXiv:2502.19634}, 2025.

\bibitem[Paul et~al.(2023)Paul, Ismayilzada, Peyrard, Borges, Bosselut, West, and Faltings]{paul2023refiner}
Debjit Paul, Mete Ismayilzada, Maxime Peyrard, Beatriz Borges, Antoine Bosselut, Robert West, and Boi Faltings.
\newblock Refiner: Reasoning feedback on intermediate representations.
\newblock \emph{arXiv preprint arXiv:2304.01904}, 2023.

\bibitem[Plini et~al.(2024)Plini, Scofano, De~Matteis, di~Melendugno, Flaborea, Sanchietti, Farinella, Galasso, and Furnari]{plini2024ti}
Leonardo Plini, Luca Scofano, Edoardo De~Matteis, Guido Maria~D'Amely di~Melendugno, Alessandro Flaborea, Andrea Sanchietti, Giovanni~Maria Farinella, Fabio Galasso, and Antonino Furnari.
\newblock Ti-prego: Chain of thought and in-context learning for online mistake detection in procedural egocentric videos.
\newblock \emph{arXiv preprint arXiv:2411.02570}, 2024.

\bibitem[Qi et~al.(2024)Qi, Huang, Panda, Henderson, Wang, and Mittal]{qi2024visual}
Xiangyu Qi, Kaixuan Huang, Ashwinee Panda, Peter Henderson, Mengdi Wang, and Prateek Mittal.
\newblock Visual adversarial examples jailbreak aligned large language models.
\newblock In \emph{Proceedings of the AAAI conference on artificial intelligence}, volume~38, pp.\  21527--21536, 2024.

\bibitem[Qiao et~al.(2024)Qiao, Tan, Dong, Wu, Sun, Song, GongQue, Lei, Wei, Zhang, et~al.]{qiao2024we}
Runqi Qiao, Qiuna Tan, Guanting Dong, Minhui Wu, Chong Sun, Xiaoshuai Song, Zhuoma GongQue, Shanglin Lei, Zhe Wei, Miaoxuan Zhang, et~al.
\newblock We-math: Does your large multimodal model achieve human-like mathematical reasoning?
\newblock \emph{arXiv preprint arXiv:2407.01284}, 2024.

\bibitem[Qiu et~al.(2024)Qiu, Lu, Zeng, Guo, Geng, Wang, Huang, Wu, and Wang]{qiu2024treebon}
Jiahao Qiu, Yifu Lu, Yifan Zeng, Jiacheng Guo, Jiayi Geng, Huazheng Wang, Kaixuan Huang, Yue Wu, and Mengdi Wang.
\newblock Treebon: Enhancing inference-time alignment with speculative tree-search and best-of-n sampling.
\newblock \emph{arXiv preprint arXiv:2410.16033}, 2024.

\bibitem[Rafailov et~al.(2023)Rafailov, Sharma, Mitchell, Manning, Ermon, and Finn]{rafailov2023direct}
Rafael Rafailov, Archit Sharma, Eric Mitchell, Christopher~D Manning, Stefano Ermon, and Chelsea Finn.
\newblock Direct preference optimization: Your language model is secretly a reward model.
\newblock \emph{Advances in Neural Information Processing Systems}, 36:\penalty0 53728--53741, 2023.

\bibitem[Rajabi \& Kosecka(2024)Rajabi and Kosecka]{rajabi2024gsr}
Navid Rajabi and Jana Kosecka.
\newblock Gsr-bench: A benchmark for grounded spatial reasoning evaluation via multimodal llms.
\newblock \emph{arXiv preprint arXiv:2406.13246}, 2024.

\bibitem[Rajpoot et~al.(2024)Rajpoot, Bhat, and Shrivastava]{rajpoot2024multimodal}
Pawan Rajpoot, Nagaraj Bhat, and Ashish Shrivastava.
\newblock Multimodal machine translation for low-resource indic languages: A chain-of-thought approach using large language models.
\newblock In \emph{Proceedings of the Ninth Conference on Machine Translation}, pp.\  833--838, 2024.

\bibitem[Ren et~al.(2025)Ren, Li, Liu, Xie, Lu, Qiao, Sha, Yan, Ma, and Shao]{ren2025llms}
Qibing Ren, Hao Li, Dongrui Liu, Zhanxu Xie, Xiaoya Lu, Yu~Qiao, Lei Sha, Junchi Yan, Lizhuang Ma, and Jing Shao.
\newblock Llms know their vulnerabilities: Uncover safety gaps through natural distribution shifts.
\newblock In \emph{Proceedings of the 63rd Annual Meeting of the Association for Computational Linguistics (Volume 1: Long Papers)}, pp.\  24763--24785, 2025.

\bibitem[Rombach et~al.(2022)Rombach, Blattmann, Lorenz, Esser, and Ommer]{rombach2022high}
Robin Rombach, Andreas Blattmann, Dominik Lorenz, Patrick Esser, and Bj{\"o}rn Ommer.
\newblock High-resolution image synthesis with latent diffusion models.
\newblock In \emph{Proceedings of the IEEE/CVF conference on computer vision and pattern recognition}, pp.\  10684--10695, 2022.

\bibitem[Rose et~al.(2023)Rose, Himakunthala, Ouyang, He, Mei, Lu, Saxon, Sonar, Mirza, and Wang]{rose2023visual}
Daniel Rose, Vaishnavi Himakunthala, Andy Ouyang, Ryan He, Alex Mei, Yujie Lu, Michael Saxon, Chinmay Sonar, Diba Mirza, and William~Yang Wang.
\newblock Visual chain of thought: bridging logical gaps with multimodal infillings.
\newblock \emph{arXiv preprint arXiv:2305.02317}, 2023.

\bibitem[Russinovich et~al.(2025)Russinovich, Salem, and Eldan]{russinovich2025great}
Mark Russinovich, Ahmed Salem, and Ronen Eldan.
\newblock Great, now write an article about that: The crescendo $\{$Multi-Turn$\}$$\{$LLM$\}$ jailbreak attack.
\newblock In \emph{34th USENIX Security Symposium (USENIX Security 25)}, pp.\  2421--2440, 2025.

\bibitem[Shangguan et~al.(2024)Shangguan, Li, Ding, Zheng, Zhao, Fitzgerald, and Cohan]{shangguan2024tomato}
Ziyao Shangguan, Chuhan Li, Yuxuan Ding, Yanan Zheng, Yilun Zhao, Tesca Fitzgerald, and Arman Cohan.
\newblock Tomato: Assessing visual temporal reasoning capabilities in multimodal foundation models.
\newblock \emph{arXiv preprint arXiv:2410.23266}, 2024.

\bibitem[Shao et~al.(2024{\natexlab{a}})Shao, Qian, Xiao, Song, Zong, Wang, Liu, and Li]{shao2024visual}
Hao Shao, Shengju Qian, Han Xiao, Guanglu Song, Zhuofan Zong, Letian Wang, Yu~Liu, and Hongsheng Li.
\newblock Visual cot: Advancing multi-modal language models with a comprehensive dataset and benchmark for chain-of-thought reasoning.
\newblock \emph{Advances in Neural Information Processing Systems}, 37:\penalty0 8612--8642, 2024{\natexlab{a}}.

\bibitem[Shao et~al.(2024{\natexlab{b}})Shao, Wang, Zhu, Xu, Song, Bi, Zhang, Zhang, Li, Wu, et~al.]{shao2024deepseekmath}
Zhihong Shao, Peiyi Wang, Qihao Zhu, Runxin Xu, Junxiao Song, Xiao Bi, Haowei Zhang, Mingchuan Zhang, YK~Li, Y~Wu, et~al.
\newblock Deepseekmath: Pushing the limits of mathematical reasoning in open language models.
\newblock \emph{arXiv preprint arXiv:2402.03300}, 2024{\natexlab{b}}.

\bibitem[Shen et~al.(2025{\natexlab{a}})Shen, Wang, Shi, Wang, Zhao, and Gu]{shen2025efficient}
Xuan Shen, Yizhou Wang, Xiangxi Shi, Yanzhi Wang, Pu~Zhao, and Jiuxiang Gu.
\newblock Efficient reasoning with hidden thinking.
\newblock \emph{arXiv preprint arXiv:2501.19201}, 2025{\natexlab{a}}.

\bibitem[Shen et~al.(2025{\natexlab{b}})Shen, Luo, Chen, Lv, and Li]{shen2025enhancing}
Zhixuan Shen, Haonan Luo, Kexun Chen, Fengmao Lv, and Tianrui Li.
\newblock Enhancing multi-robot semantic navigation through multimodal chain-of-thought score collaboration.
\newblock In \emph{Proceedings of the AAAI Conference on Artificial Intelligence}, volume~39, pp.\  14664--14672, 2025{\natexlab{b}}.

\bibitem[Shi et~al.(2024)Shi, Hu, Bin, Liu, Yang, Ng, Bing, and Lee]{shi2024math}
Wenhao Shi, Zhiqiang Hu, Yi~Bin, Junhua Liu, Yang Yang, See-Kiong Ng, Lidong Bing, and Roy Ka-Wei Lee.
\newblock Math-llava: Bootstrapping mathematical reasoning for multimodal large language models.
\newblock \emph{arXiv preprint arXiv:2406.17294}, 2024.

\bibitem[Shinn et~al.(2023)Shinn, Cassano, Gopinath, Narasimhan, and Yao]{shinn2023reflexion}
Noah Shinn, Federico Cassano, Ashwin Gopinath, Karthik Narasimhan, and Shunyu Yao.
\newblock Reflexion: Language agents with verbal reinforcement learning.
\newblock \emph{Advances in Neural Information Processing Systems}, 36:\penalty0 8634--8652, 2023.

\bibitem[Shiri et~al.(2024)Shiri, Guo, Far, Yu, Haf, and Li]{shiri2024empirical}
Fatemeh Shiri, Xiao-Yu Guo, Mona Far, Xin Yu, Reza Haf, and Yuan-Fang Li.
\newblock An empirical analysis on spatial reasoning capabilities of large multimodal models.
\newblock In \emph{Proceedings of the 2024 Conference on Empirical Methods in Natural Language Processing}, pp.\  21440--21455, 2024.

\bibitem[Song et~al.(2025)Song, Ou, Kong, Li, Neubig, and Yue]{song2025visualpuzzles}
Yueqi Song, Tianyue Ou, Yibo Kong, Zecheng Li, Graham Neubig, and Xiang Yue.
\newblock Visualpuzzles: Decoupling multimodal reasoning evaluation from domain knowledge.
\newblock \emph{arXiv preprint arXiv:2504.10342}, 2025.

\bibitem[Sun et~al.(2025{\natexlab{a}})Sun, Yan, and Weng]{sun2025thinkedit}
Chung-En Sun, Ge~Yan, and Tsui-Wei Weng.
\newblock Thinkedit: Interpretable weight editing to mitigate overly short thinking in reasoning models.
\newblock \emph{arXiv preprint arXiv:2503.22048}, 2025{\natexlab{a}}.

\bibitem[Sun et~al.(2025{\natexlab{b}})Sun, Sun, Peng, and Ye]{sun2025mitigating}
Hai-Long Sun, Zhun Sun, Houwen Peng, and Han-Jia Ye.
\newblock Mitigating visual forgetting via take-along visual conditioning for multi-modal long cot reasoning.
\newblock \emph{arXiv preprint arXiv:2503.13360}, 2025{\natexlab{b}}.

\bibitem[Sun et~al.(2024{\natexlab{a}})Sun, Haider, Zhang, Yang, Qiu, Yin, Wang, Bartlett, and Zanette]{sun2024fast}
Hanshi Sun, Momin Haider, Ruiqi Zhang, Huitao Yang, Jiahao Qiu, Ming Yin, Mengdi Wang, Peter Bartlett, and Andrea Zanette.
\newblock Fast best-of-n decoding via speculative rejection.
\newblock \emph{arXiv preprint arXiv:2410.20290}, 2024{\natexlab{a}}.

\bibitem[Sun et~al.(2024{\natexlab{b}})Sun, Wang, Wang, Zhang, and Xiao]{sun2024safeguarding}
Jiachen Sun, Changsheng Wang, Jiongxiao Wang, Yiwei Zhang, and Chaowei Xiao.
\newblock Safeguarding vision-language models against patched visual prompt injectors.
\newblock \emph{arXiv preprint arXiv:2405.10529}, 2024{\natexlab{b}}.

\bibitem[Sun et~al.(2025{\natexlab{c}})Sun, Liang, Wei, Yu, Li, Yang, Zhou, and Zhang]{sun2025mm}
Linzhuang Sun, Hao Liang, Jingxuan Wei, Bihui Yu, Tianpeng Li, Fan Yang, Zenan Zhou, and Wentao Zhang.
\newblock Mm-verify: Enhancing multimodal reasoning with chain-of-thought verification.
\newblock \emph{arXiv preprint arXiv:2502.13383}, 2025{\natexlab{c}}.

\bibitem[Sun et~al.(2024{\natexlab{c}})Sun, Hong, Pala, Toh, Tan, Ghosal, Poria, et~al.]{sun2024emma}
Qi~Sun, Pengfei Hong, Tej~Deep Pala, Vernon Toh, U~Tan, Deepanway Ghosal, Soujanya Poria, et~al.
\newblock Emma-x: An embodied multimodal action model with grounded chain of thought and look-ahead spatial reasoning.
\newblock \emph{arXiv preprint arXiv:2412.11974}, 2024{\natexlab{c}}.

\bibitem[Tan et~al.(2024{\natexlab{a}})Tan, Wei, Gao, Sun, Li, Guo, Yu, and Li]{tan2024boosting}
Cheng Tan, Jingxuan Wei, Zhangyang Gao, Linzhuang Sun, Siyuan Li, Ruifeng Guo, Bihui Yu, and Stan~Z Li.
\newblock Boosting the power of small multimodal reasoning models to match larger models with self-consistency training.
\newblock In \emph{European Conference on Computer Vision}, pp.\  305--322. Springer, 2024{\natexlab{a}}.

\bibitem[Tan et~al.(2024{\natexlab{b}})Tan, Wei, Sun, Gao, Li, Yu, Guo, and Li]{tan2024retrieval}
Cheng Tan, Jingxuan Wei, Linzhuang Sun, Zhangyang Gao, Siyuan Li, Bihui Yu, Ruifeng Guo, and Stan~Z Li.
\newblock Retrieval meets reasoning: Even high-school textbook knowledge benefits multimodal reasoning.
\newblock \emph{arXiv preprint arXiv:2405.20834}, 2024{\natexlab{b}}.

\bibitem[Thawakar et~al.(2025)Thawakar, Dissanayake, More, Thawkar, Heakl, Ahsan, Li, Zumri, Lahoud, Anwer, et~al.]{thawakar2025llamav}
Omkar Thawakar, Dinura Dissanayake, Ketan More, Ritesh Thawkar, Ahmed Heakl, Noor Ahsan, Yuhao Li, Mohammed Zumri, Jean Lahoud, Rao~Muhammad Anwer, et~al.
\newblock Llamav-o1: Rethinking step-by-step visual reasoning in llms.
\newblock \emph{arXiv preprint arXiv:2501.06186}, 2025.

\bibitem[Tie et~al.(2025)Tie, Zhou, Gu, Zhang, Hu, Zhang, Sun, Zhang, Zhou, and Sun]{tie2025mmmr}
Guiyao Tie, Xueyang Zhou, Tianhe Gu, Ruihang Zhang, Chaoran Hu, Sizhe Zhang, Mengqu Sun, Yan Zhang, Pan Zhou, and Lichao Sun.
\newblock Mmmr: Benchmarking massive multi-modal reasoning tasks.
\newblock \emph{arXiv preprint arXiv:2505.16459}, 2025.

\bibitem[Toh et~al.(2025)Toh, Chia, Ghosal, and Poria]{toh2025jumping}
Vernon~YH Toh, Yew~Ken Chia, Deepanway Ghosal, and Soujanya Poria.
\newblock The jumping reasoning curve? tracking the evolution of reasoning performance in gpt-[n] and o-[n] models on multimodal puzzles.
\newblock \emph{arXiv preprint arXiv:2502.01081}, 2025.

\bibitem[Venugopalan et~al.(2015)Venugopalan, Rohrbach, Donahue, Mooney, Darrell, and Saenko]{venugopalan2015sequence}
Subhashini Venugopalan, Marcus Rohrbach, Jeffrey Donahue, Raymond Mooney, Trevor Darrell, and Kate Saenko.
\newblock Sequence to sequence-video to text.
\newblock In \emph{Proceedings of the IEEE international conference on computer vision}, pp.\  4534--4542, 2015.

\bibitem[Wang et~al.(2025{\natexlab{a}})Wang, Liu, Bi, Zhang, Li, Ma, He, Yu, Li, Fang, et~al.]{wang2025safety}
Cheng Wang, Yue Liu, Baolong Bi, Duzhen Zhang, Zhong-Zhi Li, Yingwei Ma, Yufei He, Shengju Yu, Xinfeng Li, Junfeng Fang, et~al.
\newblock Safety in large reasoning models: A survey.
\newblock \emph{arXiv preprint arXiv:2504.17704}, 2025{\natexlab{a}}.

\bibitem[Wang et~al.(2025{\natexlab{b}})Wang, Lee, Menghini, Mols, Doughty, Khoja, Lynch, Hendryx, Yue, and Hendrycks]{wang2025enigmaeval}
Clinton~J Wang, Dean Lee, Cristina Menghini, Johannes Mols, Jack Doughty, Adam Khoja, Jayson Lynch, Sean Hendryx, Summer Yue, and Dan Hendrycks.
\newblock Enigmaeval: A benchmark of long multimodal reasoning challenges.
\newblock \emph{arXiv preprint arXiv:2502.08859}, 2025{\natexlab{b}}.

\bibitem[Wang et~al.(2024{\natexlab{a}})Wang, Wei, Liu, Zhang, Zhang, Zhang, Chong, and Zhang]{wang2024mr}
Guanqun Wang, Xinyu Wei, Jiaming Liu, Ray Zhang, Yichi Zhang, Kevin Zhang, Maurice Chong, and Shanghang Zhang.
\newblock Mr-mllm: Mutual reinforcement of multimodal comprehension and vision perception.
\newblock \emph{arXiv preprint arXiv:2406.15768}, 2024{\natexlab{a}}.

\bibitem[Wang et~al.(2024{\natexlab{b}})Wang, Pan, Shi, Lu, Ren, Zhou, Zhan, and Li]{wang2024measuring}
Ke~Wang, Junting Pan, Weikang Shi, Zimu Lu, Houxing Ren, Aojun Zhou, Mingjie Zhan, and Hongsheng Li.
\newblock Measuring multimodal mathematical reasoning with math-vision dataset.
\newblock \emph{Advances in Neural Information Processing Systems}, 37:\penalty0 95095--95169, 2024{\natexlab{b}}.

\bibitem[Wang et~al.(2025{\natexlab{c}})Wang, Wang, Xue, Pang, Liu, Chen, Qiu, Wong, Ji, and Wong]{wang2025harnessing}
Rui Wang, Hongru Wang, Boyang Xue, Jianhui Pang, Shudong Liu, Yi~Chen, Jiahao Qiu, Derek~Fai Wong, Heng Ji, and Kam-Fai Wong.
\newblock Harnessing the reasoning economy: A survey of efficient reasoning for large language models.
\newblock \emph{arXiv preprint arXiv:2503.24377}, 2025{\natexlab{c}}.

\bibitem[Wang et~al.(2024{\natexlab{c}})Wang, Ma, Zhou, Ji, Ye, and Jiang]{wang2024white}
Ruofan Wang, Xingjun Ma, Hanxu Zhou, Chuanjun Ji, Guangnan Ye, and Yu-Gang Jiang.
\newblock White-box multimodal jailbreaks against large vision-language models.
\newblock In \emph{Proceedings of the 32nd ACM International Conference on Multimedia}, pp.\  6920--6928, 2024{\natexlab{c}}.

\bibitem[Wang et~al.(2023{\natexlab{a}})Wang, Zhang, Fei, Zheng, Tang, Li, Gao, and Zhao]{wang2023caption}
Teng Wang, Jinrui Zhang, Junjie Fei, Hao Zheng, Yunlong Tang, Zhe Li, Mingqi Gao, and Shanshan Zhao.
\newblock Caption anything: Interactive image description with diverse multimodal controls.
\newblock \emph{arXiv preprint arXiv:2305.02677}, 2023{\natexlab{a}}.

\bibitem[Wang et~al.(2024{\natexlab{d}})Wang, Xie, Chu, Li, and Luo]{wang2024drivecot}
Tianqi Wang, Enze Xie, Ruihang Chu, Zhenguo Li, and Ping Luo.
\newblock Drivecot: Integrating chain-of-thought reasoning with end-to-end driving.
\newblock \emph{arXiv preprint arXiv:2403.16996}, 2024{\natexlab{d}}.

\bibitem[Wang et~al.(2024{\natexlab{e}})Wang, Chen, Wang, Cao, Liu, Gao, Zhu, Zhu, Lu, Qiao, et~al.]{wang2024enhancing}
Weiyun Wang, Zhe Chen, Wenhai Wang, Yue Cao, Yangzhou Liu, Zhangwei Gao, Jinguo Zhu, Xizhou Zhu, Lewei Lu, Yu~Qiao, et~al.
\newblock Enhancing the reasoning ability of multimodal large language models via mixed preference optimization.
\newblock \emph{arXiv preprint arXiv:2411.10442}, 2024{\natexlab{e}}.

\bibitem[Wang et~al.(2025{\natexlab{d}})Wang, Gao, Chen, Chen, Zhu, Zhao, Liu, Cao, Ye, Zhu, et~al.]{wang2025visualprm}
Weiyun Wang, Zhangwei Gao, Lianjie Chen, Zhe Chen, Jinguo Zhu, Xiangyu Zhao, Yangzhou Liu, Yue Cao, Shenglong Ye, Xizhou Zhu, et~al.
\newblock Visualprm: An effective process reward model for multimodal reasoning.
\newblock \emph{arXiv preprint arXiv:2503.10291}, 2025{\natexlab{d}}.

\bibitem[Wang et~al.(2023{\natexlab{b}})Wang, Bao, Dong, Bjorck, Peng, Liu, Aggarwal, Mohammed, Singhal, Som, et~al.]{wang2023image}
Wenhui Wang, Hangbo Bao, Li~Dong, Johan Bjorck, Zhiliang Peng, Qiang Liu, Kriti Aggarwal, Owais~Khan Mohammed, Saksham Singhal, Subhojit Som, et~al.
\newblock Image as a foreign language: Beit pretraining for vision and vision-language tasks.
\newblock In \emph{Proceedings of the IEEE/CVF Conference on Computer Vision and Pattern Recognition}, pp.\  19175--19186, 2023{\natexlab{b}}.

\bibitem[Wang et~al.(2024{\natexlab{f}})Wang, Li, and Zhang]{wang2024improved}
Xiaofei Wang, Jinhua Li, and Yifan Zhang.
\newblock Improved value alignment in large language models using variational best-of-n techniques.
\newblock 2024{\natexlab{f}}.

\bibitem[Wang et~al.(2025{\natexlab{e}})Wang, Ma, Zhang, de~Melo, Chen, and Yuille]{wang2025pulsecheck457}
Xingrui Wang, Wufei Ma, Tiezheng Zhang, Celso~M de~Melo, Jieneng Chen, and Alan Yuille.
\newblock Pulsecheck457: A diagnostic benchmark for 6d spatial reasoning of large multimodal models.
\newblock \emph{arXiv e-prints}, pp.\  arXiv--2502, 2025{\natexlab{e}}.

\bibitem[Wang et~al.(2024{\natexlab{g}})Wang, Yang, Li, Lu, Xu, Lin, Lin, Huang, and Wang]{wang2024scaling}
Xiyao Wang, Zhengyuan Yang, Linjie Li, Hongjin Lu, Yuancheng Xu, Chung-Ching Lin, Kevin Lin, Furong Huang, and Lijuan Wang.
\newblock Scaling inference-time search with vision value model for improved visual comprehension.
\newblock \emph{arXiv preprint arXiv:2412.03704}, 2024{\natexlab{g}}.

\bibitem[Wang et~al.(2022)Wang, Wei, Schuurmans, Le, Chi, Narang, Chowdhery, and Zhou]{wang2022self}
Xuezhi Wang, Jason Wei, Dale Schuurmans, Quoc Le, Ed~Chi, Sharan Narang, Aakanksha Chowdhery, and Denny Zhou.
\newblock Self-consistency improves chain of thought reasoning in language models.
\newblock \emph{arXiv preprint arXiv:2203.11171}, 2022.

\bibitem[Wang et~al.(2025{\natexlab{f}})Wang, Wu, Zhang, Yan, Liu, Luo, and Fei]{wang2025multimodal}
Yaoting Wang, Shengqiong Wu, Yuecheng Zhang, Shuicheng Yan, Ziwei Liu, Jiebo Luo, and Hao Fei.
\newblock Multimodal chain-of-thought reasoning: A comprehensive survey.
\newblock \emph{arXiv preprint arXiv:2503.12605}, 2025{\natexlab{f}}.

\bibitem[Wang et~al.(2025{\natexlab{g}})Wang, Liu, He, Zhang, Huang, Zhang, Shu, Tao, She, Yu, et~al.]{wang2025mint}
Yi~Wang, Mushui Liu, Wanggui He, Longxiang Zhang, Ziwei Huang, Guanghao Zhang, Fangxun Shu, Zhong Tao, Dong She, Zhelun Yu, et~al.
\newblock Mint: Multi-modal chain of thought in unified generative models for enhanced image generation.
\newblock \emph{arXiv preprint arXiv:2503.01298}, 2025{\natexlab{g}}.

\bibitem[Wang et~al.(2025{\natexlab{h}})Wang, Wang, Cheng, Fei, Ding, Guo, Tao, and Qiu]{wang2025visuothink}
Yikun Wang, Siyin Wang, Qinyuan Cheng, Zhaoye Fei, Liang Ding, Qipeng Guo, Dacheng Tao, and Xipeng Qiu.
\newblock Visuothink: Empowering lvlm reasoning with multimodal tree search.
\newblock \emph{arXiv preprint arXiv:2504.09130}, 2025{\natexlab{h}}.

\bibitem[Wang et~al.(2024{\natexlab{h}})Wang, Liu, Li, Chen, and Xiao]{wang2024adashield}
Yu~Wang, Xiaogeng Liu, Yu~Li, Muhao Chen, and Chaowei Xiao.
\newblock Adashield: Safeguarding multimodal large language models from structure-based attack via adaptive shield prompting.
\newblock In \emph{European Conference on Computer Vision}, pp.\  77--94. Springer, 2024{\natexlab{h}}.

\bibitem[Wang et~al.(2024{\natexlab{i}})Wang, Liu, Qu, Cao, Jiang, and Xu]{wang2024break}
Yubo Wang, Chaohu Liu, Yanqiu Qu, Haoyu Cao, Deqiang Jiang, and Linli Xu.
\newblock Break the visual perception: Adversarial attacks targeting encoded visual tokens of large vision-language models.
\newblock In \emph{Proceedings of the 32nd ACM International Conference on Multimedia}, pp.\  1072--1081, 2024{\natexlab{i}}.

\bibitem[Wang et~al.(2024{\natexlab{j}})Wang, Bingham, Yu, Le, Luong, and Ghiasi]{wang2024haloquest}
Zhecan Wang, Garrett Bingham, Adams~Wei Yu, Quoc~V Le, Thang Luong, and Golnaz Ghiasi.
\newblock Haloquest: A visual hallucination dataset for advancing multimodal reasoning.
\newblock In \emph{European Conference on Computer Vision}, pp.\  288--304. Springer, 2024{\natexlab{j}}.

\bibitem[Wei et~al.(2022)Wei, Wang, Schuurmans, Bosma, Xia, Chi, Le, Zhou, et~al.]{wei2022chain}
Jason Wei, Xuezhi Wang, Dale Schuurmans, Maarten Bosma, Fei Xia, Ed~Chi, Quoc~V Le, Denny Zhou, et~al.
\newblock Chain-of-thought prompting elicits reasoning in large language models.
\newblock \emph{Advances in neural information processing systems}, 35:\penalty0 24824--24837, 2022.

\bibitem[Wei et~al.(2024)Wei, Tan, Gao, Sun, Li, Yu, Guo, and Li]{wei2024enhancing}
Jingxuan Wei, Cheng Tan, Zhangyang Gao, Linzhuang Sun, Siyuan Li, Bihui Yu, Ruifeng Guo, and Stan~Z Li.
\newblock Enhancing human-like multimodal reasoning: a new challenging dataset and comprehensive framework.
\newblock \emph{Neural Computing and Applications}, 36\penalty0 (33):\penalty0 20849--20861, 2024.

\bibitem[Wen et~al.(2025)Wen, Zhou, Mo, and Chen]{wen2025thinkguard}
Xiaofei Wen, Wenxuan Zhou, Wenjie~Jacky Mo, and Muhao Chen.
\newblock Thinkguard: Deliberative slow thinking leads to cautious guardrails.
\newblock \emph{arXiv preprint arXiv:2502.13458}, 2025.

\bibitem[Wu et~al.(2024)Wu, Ge, Guo, Wang, Liang, Lu, Shan, and Luo]{wu2024plot2code}
Chengyue Wu, Yixiao Ge, Qiushan Guo, Jiahao Wang, Zhixuan Liang, Zeyu Lu, Ying Shan, and Ping Luo.
\newblock Plot2code: A comprehensive benchmark for evaluating multi-modal large language models in code generation from scientific plots.
\newblock \emph{arXiv preprint arXiv:2405.07990}, 2024.

\bibitem[Wu et~al.(2025)Wu, Feng, Zhang, Jin, Che, Wen, and Tao]{wu2025boosting}
Jinyang Wu, Mingkuan Feng, Shuai Zhang, Ruihan Jin, Feihu Che, Zengqi Wen, and Jianhua Tao.
\newblock Boosting multimodal reasoning with mcts-automated structured thinking.
\newblock \emph{arXiv preprint arXiv:2502.02339}, 2025.

\bibitem[Xi et~al.(2023)Xi, Meng, and Yuan]{xi2023chain}
Nan Xi, Jingjing Meng, and Junsong Yuan.
\newblock Chain-of-look prompting for verb-centric surgical triplet recognition in endoscopic videos.
\newblock In \emph{Proceedings of the 31st ACM International Conference on Multimedia}, pp.\  5007--5016, 2023.

\bibitem[Xia et~al.(2025{\natexlab{a}})Xia, Jiang, Tan, Zhu, Yue, and Zheng]{xia2025msr}
Yinan Xia, Yilei Jiang, Yingshui Tan, Xiaoyong Zhu, Xiangyu Yue, and Bo~Zheng.
\newblock Msr-align: Policy-grounded multimodal alignment for safety-aware reasoning in vision-language models.
\newblock \emph{arXiv preprint arXiv:2506.19257}, 2025{\natexlab{a}}.

\bibitem[Xia et~al.(2025{\natexlab{b}})Xia, Wang, Liu, Li, Yu, Chen, McAuley, and Li]{xia2025beyond}
Yu~Xia, Rui Wang, Xu~Liu, Mingyan Li, Tong Yu, Xiang Chen, Julian McAuley, and Shuai Li.
\newblock Beyond chain-of-thought: A survey of chain-of-x paradigms for llms.
\newblock In \emph{Proceedings of the 31st International Conference on Computational Linguistics}, pp.\  10795--10809, 2025{\natexlab{b}}.

\bibitem[Xiang et~al.(2024{\natexlab{a}})Xiang, Liu, Jiang, Nie, Huang, Fan, Li, Huang, Zeng, Han, et~al.]{xiang2024atomthink}
Kun Xiang, Zhili Liu, Zihao Jiang, Yunshuang Nie, Runhui Huang, Haoxiang Fan, Hanhui Li, Weiran Huang, Yihan Zeng, Jianhua Han, et~al.
\newblock Atomthink: A slow thinking framework for multimodal mathematical reasoning.
\newblock \emph{arXiv preprint arXiv:2411.11930}, 2024{\natexlab{a}}.

\bibitem[Xiang et~al.(2025)Xiang, Liu, Jiang, Nie, Cai, Yin, Huang, Fan, Li, Huang, et~al.]{xiang2025can}
Kun Xiang, Zhili Liu, Zihao Jiang, Yunshuang Nie, Kaixin Cai, Yiyang Yin, Runhui Huang, Haoxiang Fan, Hanhui Li, Weiran Huang, et~al.
\newblock Can atomic step decomposition enhance the self-structured reasoning of multimodal large models?
\newblock \emph{arXiv preprint arXiv:2503.06252}, 2025.

\bibitem[Xiang et~al.(2024{\natexlab{b}})Xiang, Jiang, Xiong, Ramasubramanian, Poovendran, and Li]{xiang2024badchain}
Zhen Xiang, Fengqing Jiang, Zidi Xiong, Bhaskar Ramasubramanian, Radha Poovendran, and Bo~Li.
\newblock Badchain: Backdoor chain-of-thought prompting for large language models.
\newblock \emph{arXiv preprint arXiv:2401.12242}, 2024{\natexlab{b}}.

\bibitem[Xiao et~al.(2024{\natexlab{a}})Xiao, Lin, and Li]{xiao2024mlrqa}
Jing Xiao, Guijin Lin, and Ping Li.
\newblock Mlrqa: A dataset with multimodal logical reasoning challenges.
\newblock In \emph{Pacific Rim International Conference on Artificial Intelligence}, pp.\  3--14. Springer, 2024{\natexlab{a}}.

\bibitem[Xiao et~al.(2024{\natexlab{b}})Xiao, Sun, Liu, and Wang]{xiao2024logicvista}
Yijia Xiao, Edward Sun, Tianyu Liu, and Wei Wang.
\newblock Logicvista: Multimodal llm logical reasoning benchmark in visual contexts.
\newblock \emph{arXiv preprint arXiv:2407.04973}, 2024{\natexlab{b}}.

\bibitem[Xie et~al.(2025)Xie, Lin, Liu, Wu, Yan, and Miao]{xie2025audio}
Zhifei Xie, Mingbao Lin, Zihang Liu, Pengcheng Wu, Shuicheng Yan, and Chunyan Miao.
\newblock Audio-reasoner: Improving reasoning capability in large audio language models.
\newblock \emph{arXiv preprint arXiv:2503.02318}, 2025.

\bibitem[Xing et~al.(2025)Xing, Qian, Wang, Hua, Tian, Zhou, and Tu]{xing2025openemma}
Shuo Xing, Chengyuan Qian, Yuping Wang, Hongyuan Hua, Kexin Tian, Yang Zhou, and Zhengzhong Tu.
\newblock Openemma: Open-source multimodal model for end-to-end autonomous driving.
\newblock In \emph{Proceedings of the Winter Conference on Applications of Computer Vision}, pp.\  1001--1009, 2025.

\bibitem[Xu et~al.(2024{\natexlab{a}})Xu, Jin, Hao, Song, Sun, and Yuan]{xu2024llava}
Guowei Xu, Peng Jin, Li~Hao, Yibing Song, Lichao Sun, and Li~Yuan.
\newblock Llava-o1: Let vision language models reason step-by-step.
\newblock \emph{arXiv preprint arXiv:2411.10440}, 2024{\natexlab{a}}.

\bibitem[Xu et~al.(2024{\natexlab{b}})Xu, Ge, Lyu, Li, and Li]{xu2024multimodal}
Yingrui Xu, Jingguo Ge, Guangxu Lyu, Guoyi Li, and Hui Li.
\newblock Multimodal fake news detection based on chain-of-thought prompting large language models.
\newblock In \emph{2024 IEEE International Conference on Systems, Man, and Cybernetics (SMC)}, pp.\  559--566. IEEE, 2024{\natexlab{b}}.

\bibitem[Xu et~al.(2025)Xu, Zhou, Ai, Zhao, Wang, Peng, Shao, Yao, and Zhang]{xu2025mpbench}
Zhaopan Xu, Pengfei Zhou, Jiaxin Ai, Wangbo Zhao, Kai Wang, Xiaojiang Peng, Wenqi Shao, Hongxun Yao, and Kaipeng Zhang.
\newblock Mpbench: A comprehensive multimodal reasoning benchmark for process errors identification.
\newblock \emph{arXiv preprint arXiv:2503.12505}, 2025.

\bibitem[Yan et~al.(2025{\natexlab{a}})Yan, Zheng, Wang, Yin, Liu, Cao, Su, Chen, Wu, Liao, et~al.]{yan2025visuriddles}
Hao Yan, Handong Zheng, Hao Wang, Liang Yin, Xingchen Liu, Zhenbiao Cao, Xinxing Su, Zihao Chen, Jihao Wu, Minghui Liao, et~al.
\newblock Visuriddles: Fine-grained perception is a primary bottleneck for multimodal large language models in abstract visual reasoning.
\newblock \emph{arXiv preprint arXiv:2506.02537}, 2025{\natexlab{a}}.

\bibitem[Yan et~al.(2025{\natexlab{b}})Yan, Fan, Li, Jiang, Zhao, Guan, Kuo, and Wang]{yan2025multimodal}
Qianqi Yan, Yue Fan, Hongquan Li, Shan Jiang, Yang Zhao, Xinze Guan, Ching-Chen Kuo, and Xin~Eric Wang.
\newblock Multimodal inconsistency reasoning (mmir): A new benchmark for multimodal reasoning models.
\newblock \emph{arXiv preprint arXiv:2502.16033}, 2025{\natexlab{b}}.

\bibitem[Yan et~al.(2024)Yan, Wang, Huo, Li, Li, Su, Gao, Zhang, Xu, Chu, et~al.]{yan2024errorradar}
Yibo Yan, Shen Wang, Jiahao Huo, Hang Li, Boyan Li, Jiamin Su, Xiong Gao, Yi-Fan Zhang, Tianlong Xu, Zhendong Chu, et~al.
\newblock Errorradar: Benchmarking complex mathematical reasoning of multimodal large language models via error detection.
\newblock \emph{arXiv preprint arXiv:2410.04509}, 2024.

\bibitem[Yang et~al.(2025{\natexlab{a}})Yang, Si, Duan, Zhu, Zhu, Li, Lin, Cao, and Wang]{yang2025dynamic}
Chenxu Yang, Qingyi Si, Yongjie Duan, Zheliang Zhu, Chenyu Zhu, Qiaowei Li, Zheng Lin, Li~Cao, and Weiping Wang.
\newblock Dynamic early exit in reasoning models.
\newblock \emph{arXiv preprint arXiv:2504.15895}, 2025{\natexlab{a}}.

\bibitem[Yang et~al.(2025{\natexlab{b}})Yang, Ma, Lin, and Wei]{yang2025towards}
Wenkai Yang, Shuming Ma, Yankai Lin, and Furu Wei.
\newblock Towards thinking-optimal scaling of test-time compute for llm reasoning.
\newblock \emph{arXiv preprint arXiv:2502.18080}, 2025{\natexlab{b}}.

\bibitem[Yang et~al.(2025{\natexlab{c}})Yang, He, Pan, Jiang, Deng, Yang, Lu, Yin, Rao, Zhu, et~al.]{yang2025r1}
Yi~Yang, Xiaoxuan He, Hongkun Pan, Xiyan Jiang, Yan Deng, Xingtao Yang, Haoyu Lu, Dacheng Yin, Fengyun Rao, Minfeng Zhu, et~al.
\newblock R1-onevision: Advancing generalized multimodal reasoning through cross-modal formalization.
\newblock \emph{arXiv preprint arXiv:2503.10615}, 2025{\natexlab{c}}.

\bibitem[Yao et~al.(2023{\natexlab{a}})Yao, Tian, Liu, Zhang, Liu, Jin, Li, Li, and Sun]{yao2023thinking}
Fanglong Yao, Changyuan Tian, Jintao Liu, Zequn Zhang, Qing Liu, Li~Jin, Shuchao Li, Xiaoyu Li, and Xian Sun.
\newblock Thinking like an expert: Multimodal hypergraph-of-thought (hot) reasoning to boost foundation modals.
\newblock \emph{arXiv preprint arXiv:2308.06207}, 2023{\natexlab{a}}.

\bibitem[Yao et~al.(2024)Yao, Huang, Wu, Zhang, Wang, Liu, Wang, Song, Feng, Shen, et~al.]{yao2024mulberry}
Huanjin Yao, Jiaxing Huang, Wenhao Wu, Jingyi Zhang, Yibo Wang, Shunyu Liu, Yingjie Wang, Yuxin Song, Haocheng Feng, Li~Shen, et~al.
\newblock Mulberry: Empowering mllm with o1-like reasoning and reflection via collective monte carlo tree search.
\newblock \emph{arXiv preprint arXiv:2412.18319}, 2024.

\bibitem[Yao et~al.(2023{\natexlab{b}})Yao, Yu, Zhao, Shafran, Griffiths, Cao, and Narasimhan]{yao2023tree}
Shunyu Yao, Dian Yu, Jeffrey Zhao, Izhak Shafran, Tom Griffiths, Yuan Cao, and Karthik Narasimhan.
\newblock Tree of thoughts: Deliberate problem solving with large language models.
\newblock \emph{Advances in neural information processing systems}, 36:\penalty0 11809--11822, 2023{\natexlab{b}}.

\bibitem[Ye et~al.(2025)Ye, Rong, Huang, Du, Yu, and Tao]{ye2025survey}
Mang Ye, Xuankun Rong, Wenke Huang, Bo~Du, Nenghai Yu, and Dacheng Tao.
\newblock A survey of safety on large vision-language models: Attacks, defenses and evaluations.
\newblock \emph{arXiv preprint arXiv:2502.14881}, 2025.

\bibitem[Yin et~al.(2025)Yin, Lin, Liu, Sun, and Li]{yin2025multimodal}
Hang Yin, Zhifeng Lin, Xin Liu, Bin Sun, and Kan Li.
\newblock Do multimodal language models really understand direction? a benchmark for compass direction reasoning.
\newblock In \emph{ICASSP 2025-2025 IEEE International Conference on Acoustics, Speech and Signal Processing (ICASSP)}, pp.\  1--5. IEEE, 2025.

\bibitem[Yin et~al.(2023)Yin, Fu, Zhao, Li, Sun, Xu, and Chen]{yin2023survey}
Shukang Yin, Chaoyou Fu, Sirui Zhao, Ke~Li, Xing Sun, Tong Xu, and Enhong Chen.
\newblock A survey on multimodal large language models.
\newblock \emph{arXiv preprint arXiv:2306.13549}, 2023.

\bibitem[Yu et~al.(2022)Yu, Wang, Vasudevan, Yeung, Seyedhosseini, and Wu]{yu2022coca}
Jiahui Yu, Zirui Wang, Vijay Vasudevan, Legg Yeung, Mojtaba Seyedhosseini, and Yonghui Wu.
\newblock Coca: Contrastive captioners are image-text foundation models.
\newblock \emph{arXiv preprint arXiv:2205.01917}, 2022.

\bibitem[Yu et~al.(2025)Yu, Zhang, Zhang, Liang, Zhang, Zhang, Khademi, Awadalla, Wang, Yang, et~al.]{yu2025chain}
Yiyao Yu, Yuxiang Zhang, Dongdong Zhang, Xiao Liang, Hengyuan Zhang, Xingxing Zhang, Mahmoud Khademi, Hany Awadalla, Junjie Wang, Yujiu Yang, et~al.
\newblock Chain-of-reasoning: Towards unified mathematical reasoning in large language models via a multi-paradigm perspective.
\newblock \emph{arXiv preprint arXiv:2501.11110}, 2025.

\bibitem[Yu \& Luo(2024)Yu and Luo]{yu2024chain}
Yongsheng Yu and Jiebo Luo.
\newblock Chain-of-thought prompting for demographic inference with large multimodal models.
\newblock In \emph{2024 IEEE International Conference on Multimedia and Expo (ICME)}, pp.\  1--7. IEEE, 2024.

\bibitem[Yuan et~al.(2025)Yuan, Peng, Jiang, Lu, Zhang, Feng, Fu, Chen, Bai, Zhang, et~al.]{yuan2025mme}
Jiakang Yuan, Tianshuo Peng, Yilei Jiang, Yiting Lu, Renrui Zhang, Kaituo Feng, Chaoyou Fu, Tao Chen, Lei Bai, Bo~Zhang, et~al.
\newblock Mme-reasoning: A comprehensive benchmark for logical reasoning in mllms.
\newblock \emph{arXiv preprint arXiv:2505.21327}, 2025.

\bibitem[Zaremba et~al.(2025)Zaremba, Nitishinskaya, Barak, Lin, Toyer, Yu, Dias, Wallace, Xiao, Heidecke, et~al.]{zaremba2025trading}
Wojciech Zaremba, Evgenia Nitishinskaya, Boaz Barak, Stephanie Lin, Sam Toyer, Yaodong Yu, Rachel Dias, Eric Wallace, Kai Xiao, Johannes Heidecke, et~al.
\newblock Trading inference-time compute for adversarial robustness.
\newblock \emph{arXiv preprint arXiv:2501.18841}, 2025.

\bibitem[Zawalski et~al.(2024)Zawalski, Chen, Pertsch, Mees, Finn, and Levine]{zawalski2024robotic}
Micha{\l} Zawalski, William Chen, Karl Pertsch, Oier Mees, Chelsea Finn, and Sergey Levine.
\newblock Robotic control via embodied chain-of-thought reasoning.
\newblock \emph{arXiv preprint arXiv:2407.08693}, 2024.

\bibitem[Zhai et~al.(2024)Zhai, Bai, Lin, Pan, Tong, Zhou, Suhr, Xie, LeCun, Ma, et~al.]{zhai2024fine}
Simon Zhai, Hao Bai, Zipeng Lin, Jiayi Pan, Peter Tong, Yifei Zhou, Alane Suhr, Saining Xie, Yann LeCun, Yi~Ma, et~al.
\newblock Fine-tuning large vision-language models as decision-making agents via reinforcement learning.
\newblock \emph{Advances in neural information processing systems}, 37:\penalty0 110935--110971, 2024.

\bibitem[Zhang et~al.(2025{\natexlab{a}})Zhang, Liu, Dong, Zang, Zhang, Duan, Cao, Lin, and Wang]{zhang2025booststep}
Beichen Zhang, Yuhong Liu, Xiaoyi Dong, Yuhang Zang, Pan Zhang, Haodong Duan, Yuhang Cao, Dahua Lin, and Jiaqi Wang.
\newblock Booststep: Boosting mathematical capability of large language models via improved single-step reasoning.
\newblock \emph{arXiv preprint arXiv:2501.03226}, 2025{\natexlab{a}}.

\bibitem[Zhang et~al.(2024{\natexlab{a}})Zhang, Yang, Lyu, Jin, Yao, Chen, and Luo]{zhang2024cocot}
Daoan Zhang, Junming Yang, Hanjia Lyu, Zijian Jin, Yuan Yao, Mingkai Chen, and Jiebo Luo.
\newblock Cocot: Contrastive chain-of-thought prompting for large multimodal models with multiple image inputs.
\newblock \emph{arXiv preprint arXiv:2401.02582}, 2024{\natexlab{a}}.

\bibitem[Zhang et~al.(2025{\natexlab{b}})Zhang, Yang, Lyu, Jin, Yao, Chen, and Luo]{zhang2025benchmark}
Daoan Zhang, Junming Yang, Hanjia Lyu, Zijian Jin, Yuan Yao, Mingkai Chen, and Jiebo Luo.
\newblock A benchmark and chain-of-thought prompting strategy for large multimodal models with multiple image inputs.
\newblock In \emph{International Conference on Pattern Recognition}, pp.\  226--241. Springer, 2025{\natexlab{b}}.

\bibitem[Zhang et~al.(2025{\natexlab{c}})Zhang, Lei, Li, Wang, Liu, Yang, Li, Wang, Yang, Wu, et~al.]{zhang2025critic}
Di~Zhang, Jingdi Lei, Junxian Li, Xunzhi Wang, Yujie Liu, Zonglin Yang, Jiatong Li, Weida Wang, Suorong Yang, Jianbo Wu, et~al.
\newblock Critic-v: Vlm critics help catch vlm errors in multimodal reasoning.
\newblock In \emph{Proceedings of the Computer Vision and Pattern Recognition Conference}, pp.\  9050--9061, 2025{\natexlab{c}}.

\bibitem[Zhang et~al.(2024{\natexlab{b}})Zhang, Yu, Dong, Li, Su, Chu, and Yu]{zhang2024mm}
Duzhen Zhang, Yahan Yu, Jiahua Dong, Chenxing Li, Dan Su, Chenhui Chu, and Dong Yu.
\newblock Mm-llms: Recent advances in multimodal large language models.
\newblock \emph{arXiv preprint arXiv:2401.13601}, 2024{\natexlab{b}}.

\bibitem[Zhang et~al.(2024{\natexlab{c}})Zhang, Wu, Bai, Lin, Li, Yu, Wang, Chen, and Keung]{zhang2024humaneval}
Fengji Zhang, Linquan Wu, Huiyu Bai, Guancheng Lin, Xiao Li, Xiao Yu, Yue Wang, Bei Chen, and Jacky Keung.
\newblock Humaneval-v: Evaluating visual understanding and reasoning abilities of large multimodal models through coding tasks.
\newblock \emph{arXiv preprint arXiv:2410.12381}, 2024{\natexlab{c}}.

\bibitem[Zhang et~al.(2025{\natexlab{d}})Zhang, Meng, Luo, Han, Liao, Cambria, and Fei]{zhang2025towards}
Han Zhang, Zixiang Meng, Meng Luo, Hong Han, Lizi Liao, Erik Cambria, and Hao Fei.
\newblock Towards multimodal empathetic response generation: A rich text-speech-vision avatar-based benchmark.
\newblock In \emph{Proceedings of the ACM on Web Conference 2025}, pp.\  2872--2881, 2025{\natexlab{d}}.

\bibitem[Zhang et~al.(2023{\natexlab{a}})Zhang, Li, and Bing]{zhang2023video}
Hang Zhang, Xin Li, and Lidong Bing.
\newblock Video-llama: An instruction-tuned audio-visual language model for video understanding.
\newblock In \emph{Proceedings of the 2023 Conference on Empirical Methods in Natural Language Processing: System Demonstrations}, pp.\  543--553, 2023{\natexlab{a}}.

\bibitem[Zhang et~al.(2024{\natexlab{d}})Zhang, Gao, Zhang, Li, and Yin]{zhang2024smartagent}
Jiaqi Zhang, Chen Gao, Liyuan Zhang, Yong Li, and Hongzhi Yin.
\newblock Smartagent: Chain-of-user-thought for embodied personalized agent in cyber world.
\newblock \emph{arXiv preprint arXiv:2412.07472}, 2024{\natexlab{d}}.

\bibitem[Zhang et~al.(2024{\natexlab{e}})Zhang, Jiang, Zhang, Lin, Guo, Qiu, Zhou, Lu, Chang, Qiao, et~al.]{zhang2024mathverse}
Renrui Zhang, Dongzhi Jiang, Yichi Zhang, Haokun Lin, Ziyu Guo, Pengshuo Qiu, Aojun Zhou, Pan Lu, Kai-Wei Chang, Yu~Qiao, et~al.
\newblock Mathverse: Does your multi-modal llm truly see the diagrams in visual math problems?
\newblock In \emph{European Conference on Computer Vision}, pp.\  169--186. Springer, 2024{\natexlab{e}}.

\bibitem[Zhang et~al.(2024{\natexlab{f}})Zhang, Chen, Dai, Jiang, Hu, Liu, and Cao]{zhang2024meter}
Ruwen Zhang, Jinglu Chen, Mingjie Dai, Xinyi Jiang, Yuxin Hu, Bo~Liu, and Jiuxin Cao.
\newblock Meter: Multimodal hallucination detection with mixture of experts via tools ensembling and reasoning.
\newblock In \emph{CCF International Conference on Natural Language Processing and Chinese Computing}, pp.\  274--286. Springer, 2024{\natexlab{f}}.

\bibitem[Zhang et~al.(2025{\natexlab{e}})Zhang, Wang, Liu, Huixin, Jiang, Shen, Hou, Zheng, Zhang, Li, et~al.]{zhang2025embodied}
Wenqi Zhang, Mengna Wang, Gangao Liu, Xu~Huixin, Yiwei Jiang, Yongliang Shen, Guiyang Hou, Zhe Zheng, Hang Zhang, Xin Li, et~al.
\newblock Embodied-reasoner: Synergizing visual search, reasoning, and action for embodied interactive tasks.
\newblock \emph{arXiv preprint arXiv:2503.21696}, 2025{\natexlab{e}}.

\bibitem[Zhang et~al.(2023{\natexlab{b}})Zhang, Wu, Zhao, Lin, Zhang, Wang, and Xie]{zhang2023pmc}
Xiaoman Zhang, Chaoyi Wu, Ziheng Zhao, Weixiong Lin, Ya~Zhang, Yanfeng Wang, and Weidi Xie.
\newblock Pmc-vqa: Visual instruction tuning for medical visual question answering.
\newblock \emph{arXiv preprint arXiv:2305.10415}, 2023{\natexlab{b}}.

\bibitem[Zhang et~al.(2025{\natexlab{f}})Zhang, Zhang, Ju, Liu, Mao, Sun, Wu, Gao, Cai, Qin, et~al.]{zhang2025embodiedvsr}
Yi~Zhang, Qiang Zhang, Xiaozhu Ju, Zhaoyang Liu, Jilei Mao, Jingkai Sun, Jintao Wu, Shixiong Gao, Shihan Cai, Zhiyuan Qin, et~al.
\newblock Embodiedvsr: Dynamic scene graph-guided chain-of-thought reasoning for visual spatial tasks.
\newblock \emph{arXiv preprint arXiv:2503.11089}, 2025{\natexlab{f}}.

\bibitem[Zhang et~al.(2025{\natexlab{g}})Zhang, Zeng, Li, Huang, Deng, and Dong]{zhang2025realsafe}
Yichi Zhang, Zihao Zeng, Dongbai Li, Yao Huang, Zhijie Deng, and Yinpeng Dong.
\newblock Realsafe-r1: Safety-aligned deepseek-r1 without compromising reasoning capability.
\newblock \emph{arXiv preprint arXiv:2504.10081}, 2025{\natexlab{g}}.

\bibitem[Zhang et~al.(2025{\natexlab{h}})Zhang, Zhang, Huang, Xia, Fang, Yang, Duan, Yan, Dong, and Zhu]{zhang2025stair}
Yichi Zhang, Siyuan Zhang, Yao Huang, Zeyu Xia, Zhengwei Fang, Xiao Yang, Ranjie Duan, Dong Yan, Yinpeng Dong, and Jun Zhu.
\newblock Stair: Improving safety alignment with introspective reasoning.
\newblock \emph{arXiv preprint arXiv:2502.02384}, 2025{\natexlab{h}}.

\bibitem[Zhang et~al.(2024{\natexlab{g}})Zhang, Zhang, Li, Pu, Setiadharma, Yang, and Liu]{zhang2024worldqa}
Yuanhan Zhang, Kaichen Zhang, Bo~Li, Fanyi Pu, Christopher~Arif Setiadharma, Jingkang Yang, and Ziwei Liu.
\newblock Worldqa: Multimodal world knowledge in videos through long-chain reasoning.
\newblock \emph{arXiv preprint arXiv:2405.03272}, 2024{\natexlab{g}}.

\bibitem[Zhang et~al.(2023{\natexlab{c}})Zhang, Zhang, Li, Zhao, Karypis, and Smola]{zhang2023multimodal}
Zhuosheng Zhang, Aston Zhang, Mu~Li, Hai Zhao, George Karypis, and Alex Smola.
\newblock Multimodal chain-of-thought reasoning in language models.
\newblock \emph{arXiv preprint arXiv:2302.00923}, 2023{\natexlab{c}}.

\bibitem[Zhao et~al.(2025{\natexlab{a}})Zhao, Wu, Zhang, and Vasilakos]{zhao2025shadowcot}
Gejian Zhao, Hanzhou Wu, Xinpeng Zhang, and Athanasios~V Vasilakos.
\newblock Shadowcot: Cognitive hijacking for stealthy reasoning backdoors in llms.
\newblock \emph{arXiv preprint arXiv:2504.05605}, 2025{\natexlab{a}}.

\bibitem[Zhao et~al.(2025{\natexlab{b}})Zhao, Wei, and Bo]{zhao2025r1}
Jiaxing Zhao, Xihan Wei, and Liefeng Bo.
\newblock R1-omni: Explainable omni-multimodal emotion recognition with reinforcement learning.
\newblock \emph{arXiv preprint arXiv:2503.05379}, 2025{\natexlab{b}}.

\bibitem[Zhao et~al.(2024{\natexlab{a}})Zhao, Huang, Fu, Li, Gong, Liu, Bi, and Kong]{zhao2024bba}
Xueliang Zhao, Xinting Huang, Tingchen Fu, Qintong Li, Shansan Gong, Lemao Liu, Wei Bi, and Lingpeng Kong.
\newblock Bba: Bi-modal behavioral alignment for reasoning with large vision-language models.
\newblock \emph{arXiv preprint arXiv:2402.13577}, 2024{\natexlab{a}}.

\bibitem[Zhao et~al.(2024{\natexlab{b}})Zhao, Zheng, Luo, Li, Ma, and Jiang]{zhao2024bluesuffix}
Yunhan Zhao, Xiang Zheng, Lin Luo, Yige Li, Xingjun Ma, and Yu-Gang Jiang.
\newblock Bluesuffix: Reinforced blue teaming for vision-language models against jailbreak attacks.
\newblock \emph{arXiv preprint arXiv:2410.20971}, 2024{\natexlab{b}}.

\bibitem[Zhao et~al.(2023)Zhao, Pang, Du, Yang, Li, Cheung, and Lin]{zhao2023evaluating}
Yunqing Zhao, Tianyu Pang, Chao Du, Xiao Yang, Chongxuan Li, Ngai-Man~Man Cheung, and Min Lin.
\newblock On evaluating adversarial robustness of large vision-language models.
\newblock \emph{Advances in Neural Information Processing Systems}, 36:\penalty0 54111--54138, 2023.

\bibitem[Zheng et~al.(2024{\natexlab{a}})Zheng, Liang, Zhang, Wei, Chua, and Li]{zheng2024picture}
Changmeng Zheng, Dayong Liang, Wengyu Zhang, Xiao-Yong Wei, Tat-Seng Chua, and Qing Li.
\newblock A picture is worth a graph: A blueprint debate paradigm for multimodal reasoning.
\newblock In \emph{Proceedings of the 32nd ACM International Conference on Multimedia}, pp.\  419--428, 2024{\natexlab{a}}.

\bibitem[Zheng et~al.(2023)Zheng, Yang, Tang, Zhou, and Yang]{zheng2023ddcot}
Ge~Zheng, Bin Yang, Jiajin Tang, Hong-Yu Zhou, and Sibei Yang.
\newblock Ddcot: Duty-distinct chain-of-thought prompting for multimodal reasoning in language models.
\newblock \emph{Advances in Neural Information Processing Systems}, 36:\penalty0 5168--5191, 2023.

\bibitem[Zheng et~al.(2024{\natexlab{b}})Zheng, Xu, Sun, Pu, Chen, and Sun]{zheng2024thinking}
Haojie Zheng, Tianyang Xu, Hanchi Sun, Shu Pu, Ruoxi Chen, and Lichao Sun.
\newblock Thinking before looking: Improving multimodal llm reasoning via mitigating visual hallucination.
\newblock \emph{arXiv preprint arXiv:2411.12591}, 2024{\natexlab{b}}.

\bibitem[Zhou et~al.(2025)Zhou, Liu, Zhao, Jangam, Srinivasa, Liu, Song, and Wang]{zhou2025hidden}
Kaiwen Zhou, Chengzhi Liu, Xuandong Zhao, Shreedhar Jangam, Jayanth Srinivasa, Gaowen Liu, Dawn Song, and Xin~Eric Wang.
\newblock The hidden risks of large reasoning models: A safety assessment of r1.
\newblock \emph{arXiv preprint arXiv:2502.12659}, 2025.

\bibitem[Zhou et~al.(2024{\natexlab{a}})Zhou, Zhou, Hu, Lu, Gao, and Zhang]{zhou2024image}
Qiji Zhou, Ruochen Zhou, Zike Hu, Panzhong Lu, Siyang Gao, and Yue Zhang.
\newblock Image-of-thought prompting for visual reasoning refinement in multimodal large language models.
\newblock \emph{arXiv preprint arXiv:2405.13872}, 2024{\natexlab{a}}.

\bibitem[Zhou et~al.(2024{\natexlab{b}})Zhou, He, Chen, Li, Chen, Guti{\'e}rrez-Basulto, Pan, and Chen]{zhou2024miceval}
Xiongtao Zhou, Jie He, Lanyu Chen, Jingyu Li, Haojing Chen, V{\'\i}ctor Guti{\'e}rrez-Basulto, Jeff~Z Pan, and Hanjie Chen.
\newblock Miceval: Unveiling multimodal chain of thought's quality via image description and reasoning steps.
\newblock \emph{arXiv preprint arXiv:2410.14668}, 2024{\natexlab{b}}.

\bibitem[Zhu et~al.(2023)Zhu, Chen, Shen, Li, and Elhoseiny]{zhu2023minigpt}
Deyao Zhu, Jun Chen, Xiaoqian Shen, Xiang Li, and Mohamed Elhoseiny.
\newblock Minigpt-4: Enhancing vision-language understanding with advanced large language models.
\newblock \emph{arXiv preprint arXiv:2304.10592}, 2023.

\end{thebibliography}
